\documentclass{article}

\pdfoutput=1

\usepackage{arxiv}
\usepackage{graphicx}
\usepackage{wrapfig}
\usepackage{subcaption}
\usepackage{amsmath}
\usepackage{amsfonts}
\usepackage{array}
\usepackage{multirow}
\usepackage{hyperref}
\usepackage[utf8]{inputenc}
\usepackage[T1]{fontenc}
\usepackage{url}
\usepackage{booktabs}
\usepackage{nicefrac}
\usepackage{microtype}
\usepackage{xcolor}
\usepackage{authblk}

\DeclareMathOperator*{\argmin}{arg\,min}
\graphicspath{ {./images/} }

\title{
    Conditioning Generative Latent Optimization for Sparse-View CT Image Reconstruction
}

\author[1]{Thomas Braure}
\author[2]{Delphine Lazaro}
\author[1]{David Hateau}
\author[1]{Vincent Brandon}
\author[1]{K\'evin Ginsburger}

\affil[1]{\footnotesize CEA DIF, 91297 Arpajon Cedex, France}
\affil[2]{\footnotesize CEA LIST, 91191 Gif-sur-Yvette Cedex, France}

\begin{document}
\maketitle

\begin{abstract}

  Computed Tomography (CT) is a prominent example of Imaging Inverse Problem (IIP),
  highlighting the unrivaled performances of data-driven methods in degraded measurements
  setups like sparse X-ray projections. Although a significant proportion of deep learning
  approaches benefit from large supervised datasets to directly map experimental measurements
  to medical scans, they cannot generalize to new experimental setups, e.g.\ changes concerning
  source/detector specifications and/or the acquisition geometry. In contrast, fully
  unsupervised techniques, most notably using conditioned Score-based Generative Models (cSGM)
  with Manifold Constrained Gradients (MCG), have recently demonstrated similar or better
  performances compared to supervised approaches to solve IIPs while being flexible at test
  time regarding the experimental setup. However, their use cases are limited as they need
  considerable amounts of training data to have good generalization properties. Another
  unsupervised approach taking advantage of the implicit natural bias of deep convolutional
  networks, Deep Image Prior (DIP), has recently been adapted to solve low-dose and sparse CT
  by reparameterizing the reconstruction optimization problem. Although this methodology does
  not require any training dataset, it enforces a weaker prior on the reconstructions when
  compared to data-driven methods. To fill the gap between these two strategies, we propose an
  unsupervised conditional approach to the Generative Latent Optimization framework (cGLO), in
  which a decoder is used for reconstruction purposes, by performing Generative Latent
  Optimization with a loss function directly comparing simulated measurements from proposed
  reconstructions to experimental measurements. Similarly to DIP, without any training dataset,
  cGLO benefits from the structural bias of a decoder network.  However, the prior is further
  reinforced as the effect of a likelihood objective shared between multiple slices being
  reconstructed simultaneously through the same decoder network. In addition, the parameters of
  the decoder may be initialized on an unsupervised, and eventually very small, training
  dataset to enhance the reconstruction. The resulting approach is tested on full-dose
  sparse-view CT using multiple training dataset sizes and varying numbers of viewing angles.
  Our method demonstrates better reconstruction Structural SIMilarity (SSIM) compared to the
  state-of-the-art MCG strategy, between $+0.041$ and $+0.134$. cGLO also shows an increasing
  performance advantage over MCG for smaller training datasets, reaching a $+2.58$ dB Peak
  Signal to Noise Ratio (PSNR) gain. When no training dataset is available, our method also
  outperforms the DIP approach with at least $+1.52$ dB PSNR advantage, and especially with
  fewer viewing angles reaching $+3.15$ dB. Furthermore, cGLO does not require any backward
  operator and could expand use cases, like recent developments of score-based models, to
  non-linear IIPs.

\end{abstract}

\section{Introduction}

  Imaging Inverse Problems (IIPs) are a large and swiftly growing area of research. IIPs cover
  a large variety of applications, ranging from generic image processing topics such as
  denoising, super-resolution and deconvolution to more specific domains usually involving the
  characterization of a system of interest from indirect observations. In this latter case, the
  problem of reconstructing scans from medical imaging devices such as Computed Tomography
  (CT), Magnetic Resonance Imaging (MRI) and Positron Emission Tomography (PET) offers major
  challenges yet to be solved. Among those, CT reconstruction is a prime example of medical
  IIPs. The issue of delivered doses to patients has led to sparser sets of X-ray projection
  angles and/or a decrease in X-ray intensities resulting in noisier, less contrasted and
  detailed projection images. When using such degraded measurement setups, reconstructions
  obtained by conventional numerical methods, e.g.\ Filtered Back-Projection (FBP)
  \cite{deans_radon_1983}, are severely worsened.

  Recent progress have been made using deep learning approaches for CT reconstruction.
  However, a large part of these novel data-driven methods employ a supervised training setup
  \cite{wei_sparse_2020, adler_deep_2018, adler_learned_2018,
  lee_deep-neural-network-based_2019, zhu_image_2018, tonolini_variational_2020,
  adler_solving_2017}. In supervised training pairs, ground truths consist of 3D volumetric
  scans reconstructed from conventional techniques such as the well-known FBP algorithm, using
  high-dose and densely sampled X-ray measurements. A degraded set of measurements is
  associated to each ground truth. These are usually obtained by taking a subset of the original
  X-ray projections (full-dose sparse view CT or sparse CT) and/or by adding noise to simulate
  dose reduction (low-dose CT). Building such supervised datasets thus implies that a fixed
  physical model has been employed for all training pairs, which must be identical to the one
  used for inference. In other words, supervised reconstruction strategies need new training
  whenever the acquisition process changes. This can be problematic when the viewing angles,
  X-ray spectrum or beam geometry vary. To circumvent this drawback, several unsupervised
  strategies have recently been introduced.

  Current approaches dealing with ill-posed IIPs in an unsupervised way are mostly based on the
  use of generative models. As such, the Generative Adversarial Networks (GANs)
  \cite{goodfellow_generative_2014} are largely employed for unsupervised reconstruction tasks
  \cite{xia_gan_2023}. As serious contenders to GANs, Diffusion Models (DMs)
  \cite{song_generative_2019, ho_denoising_2020} have gained a lot of attention in the past few
  years, as recent improvements have led to the generation of high-quality image samples
  without requiring a complex adversarial optimization \cite{dhariwal_diffusion_2021,
  kim_clip_2021, rombach_latent_2021, song_consistency_2023}. The basic principle of DMs is to
  add noise of increasing intensity to data during the training process and learn to reverse
  this process, i.e.\ starting from random noise and producing a sample from the data
  distribution through successive denoisings. In the continuous approach
  \cite{song_score-based_2021}, the discrete sequence of noise levels is replaced by a
  continuum of distributions progressively diffusing data points into random noise following a
  Stochastic Differential Equation (SDE) describing a markovian diffusion process. The
  reverse-time SDE is approximated with a time-dependent neural network estimating the gradient
  of the prior likelihood, also called the score function. Using standard iterative sampling
  techniques then leads to generative capability. Following this general strategy, two types of
  SDE formulations can be distinguished: the variance preserving Denoising Diffusion
  Probabilistic Models (DDPMs) \cite{ho_denoising_2020, nichol_improved_2021} and the variance
  exploding Score-matching Generative Models (SGMs) \cite{song_generative_2019,
  song_score-based_2021, song_improved_2020}. In particular, recent works involving DMs have
  proven strong performances in solving IIPs \cite{chung_come_2021, lugmayr_repaint_2022,
  chung_parallel_2022, chung_posterior_2022, liu_dolce_2022} and more specifically, sparse CT
  reconstruction \cite{chung_posterior_2022, song_solving_2022, chung_manifold_2022}. Both GANs
  and DMs are very efficient to learn unconditional prior distributions on slices of 3D
  reconstructed volumes. However, difficulties arise when a conditional sampling is needed to
  reconstruct slices from experimental X-ray projections.  Various strategies have been
  proposed to deal with this conditional sampling issue.

  GAN inversion approaches \cite{xia_gan_2023} start from a \textit{trained and fixed} decoder,
  and aim at estimating the latent code corresponding to a given observation. The inversion
  process can either be learning-based \cite{richardson_encoding_2021,
  perarnau_invertible_2016}, e.g.\ learning an encoder to inverse the decoder, or
  optimization-based where the optimization problem is solved by finding the latent code
  minimizing the given objective function \cite{abdal_image2stylegan_2019,
  huh_transforming_2020}, or both \cite{zhu_generative_2016, zhu_-domain_2020,
  bau_seeing_2019}. Other approaches using a parameterization of the latent space have been
  proposed, where the decoder and the encoder are jointly learned using an adversarial process
  \cite{donahue_adversarial_2017, dumoulin_adversarially_2017, zhu_disentangled_2022,
  daras_intermediate_2021, ulyanov_it_2018, kelkar_prior_2021, menon_pulse_2020}. Similarly to
  GANs, conditional inference using DMs is an open field of research. It is generally performed
  using a biased sampling of the posterior distribution to generate approximate samples from a
  stochastic process conditioned to experimental measurements. For instance, in the context of
  sparse CT reconstruction, conditioning an SGM (cSGM) to X-ray projections, as in
  \cite{song_solving_2022}, involves computing the FBP on experimental measurements completed
  with simulated measurements from the candidate reconstruction, i.e.\ a proximal projection
  step throwing the sample path off the data manifold according to \cite{chung_manifold_2022}.
  In the latter, a Manifold Constraint on the Gradient (MCG) is enforced during the sampling
  process as an additional correction term resulting in more robust reconstructions. While both
  these methods are restricted to linear IIPs, a recent work from Chung et al.\
  \cite{chung_posterior_2022}, known as Deep Posterior Sampling (DPS), has extended the use
  cases of DMs to noisy non-linear problems. As interestingly noted in
  \cite{chung_posterior_2022}, DPS is equivalent to MCG in \textit{noiseless} configurations,
  e.g.\ full-dose sparse view CT which is the application of interest in this paper.

  Unsupervised generative models require large amounts of training data, which can be difficult
  to collect. Furthermore, while these approaches have good generalization given sufficient
  training data, in practice it is not clear how much is necessary. To address this issue,
  Ulyanov et al.\ proposed a method specific to IIP solving, Deep Image Prior (DIP)
  \cite{ulyanov_deep_2018}. They showed that the structure of a deep convolutional
  network inherently captures a significant amount of low-level image statistics without any
  learning. That is to say, it induces an implicit structural bias, i.e.\ a prior. DIP uses a
  randomly initialized U-Net architecture \cite{ronneberger_u-net_2015}, and a fixed input
  noise. The network weights can then be optimized to solve any ill-posed IIP with known
  forward model, such as inpainting or super-resolution. It has been shown that this IIP
  reparameterization, when optimized with gradient descent, hierarchically reconstructs from
  low to high frequencies, i.e.\ converging faster towards "natural images" than noise
  \cite{heckel_denoising_2020, chakrabarty_spectral_2019}. The method has recently been adapted
  in \cite{baguer_computed_2020} to solve low-dose and sparse CT reconstruction. While this
  strategy showed good results on use cases with missing or small datasets, it imposes a quite
  rough prior which cannot compete with the reconstruction performances of its data-driven
  counterparts in degraded measurement setups.

  This paper bridges the gap between, on one side, data-hungry generative models offering
  strong priors, and data-free methods providing weak priors on the other side, by proposing a
  reconstruction method readily usable \textit{with} or \textit{without} training data. Our
  method is a conditional version of the Generative Latent Optimization (GLO)
  \cite{bojanowski_optimizing_2018} framework. GLO was designed as a generative model
  attempting to produce competitive results when compared to GANs without using an adversarial
  training protocol. Models building on GLO are focused on improving its generative
  characteristics \cite{azuri_generative_2021, hoshen_non-adversarial_2019,
  xiao_optimizing_2019}, and to the best of our knowledge, it has never been conditioned to
  solve IIPs. Indeed, exploring or sampling the latent space of GLO is not trivial
  \cite{hoshen_non-adversarial_2019, xiao_optimizing_2019}, thus making it difficult to apply
  standard GAN inversion technique to directly solve IIPs. Instead, in this work, GLO is used
  to reparameterize the IIP. Similarly to DIP, our conditioned version of GLO (cGLO) exploits
  the inherent structural prior induced by a convolutional decoder, allowing reconstructions
  without any training dataset. However, contrary to DIP, cGLO can benefit from a shared
  objective by reconstructing multiple slices at the same time and an unsupervised training
  dataset of any size to initialize its decoder weights, resulting in greatly improved
  reconstruction results. The higher the quantity of training data is used for initialization,
  the stronger the induced prior is.

  The paper is organized as follows. Section 2 details our cGLO method, built upon the
  reparameterization strategy of DIP and the framework of GLO. Consequently, both DIP and GLO
  are introduced beforehand. Section 3 describes the experimental setups used to benchmark our
  method on the full-dose sparse view CT reconstruction task. These setups correspond to
  situations with sparse or abundant training data. The section ends with implementation
  details regarding network architecture and hyperparameter values. Our method cGLO is then
  compared, in Section 4, through quantitative and qualitative results for two distinct
  scenarios. It is first compared to DIP and FBP \textit{without} any training data available.
  Then it is compared to cSGM and MCG \textit{with} prior unsupervised training on datasets of
  varying sizes. The paper concludes with a summary and suggestions for future works in
  Section 5.

\section{Method}

  The method we propose here builds on the framework described by GLO and on the DIP approach
  which is a method dedicated to solve IIPs through reparameterization of the objective. This
  reformulation of DIP is detailed for a denoising task and for its adaptation to sparse CT
  reconstruction. Then, GLO and the concept of latent optimization are presented in the context
  of \textit{unconditioned} generation, i.e.\ not directly applicable to IIP solving. Finally,
  our method cGLO is described with an objective function corresponding to the experiments
  conducted in this paper, i.e.\ full-dose sparse CT. In addition, a formalism for the Radon
  transform/CT and for score-based methods (SGM, cSGM and MCG) is given in Appendix A and B.

  \subsection{Deep Image Prior (DIP)}

    Ulyanov et al.\ \cite{ulyanov_deep_2018} introduced DIP to solve classic IIPs such as
    denoising and super-resolution. The core idea of DIP is to regularize IIPs by taking
    advantage of the structural bias of a U-Net \cite{ronneberger_u-net_2015} $f_\theta$
    parameterized with a set of weights $\theta$. It operates through two mechanisms:
    reparameterization and early stopping ($ES$). For instance, in the case of denoising, the
    optimization problem is reparameterized as:

    \begin{equation}
    \label{unconditioned_dip}
        x^* = \lbrace
            f_{\theta^*}(z) \bigm|
            \theta^* = \argmin_{\theta} \lVert
                f_{\theta}(z) - (x_0 + \eta)
            \rVert^2_2
        \rbrace
    \end{equation}

    where $x_0$ is the initial image perturbed with unknown white noise $\eta$ and $z$ is a
    fixed white noise with the same dimensions as $x_0$.  Experiments in
    \cite{ulyanov_deep_2018} showed that given sufficient capacity and time/iterations, via
    gradient descent, the randomly initialized and over-parameterized U-Net can fit the output
    signal almost perfectly, including the noise $\eta$. However, Ulyanov et al.\
    \cite{ulyanov_deep_2018} showed that the weights descent sequence $\theta_1, ...,
    \theta_N$, with $f_{\theta_N} \approx x_0+ \eta$, contains an early stopping point
    $\theta_{ES}$, such that $f_{\theta_{ES}} \approx x_0$. This phenomenon has been proven to
    be a consequence of the structure of the generative network, more specifically the
    convolutional and upsampling layers \cite{heckel_denoising_2020,
    chakrabarty_spectral_2019}. They induce a spectral bias in which the decoder learns to
    construct the image from low to high frequencies, meaning that with an appropriate choice
    of $ES$, one can prevent the decoder from fitting the high frequency perturbations.

    This methodology has recently been tailored to reconstruct 3D CT volumes by Baguer et al.\
    \cite{baguer_computed_2020}, using an adaptation of Eq.\ \eqref{unconditioned_dip} further
    regularized with Total Variation (TV):

    \begin{equation}
    \label{conditioned_dip}
        x^* = \lbrace
            f_{\theta^*}(z) \bigm|
            \theta^* = \argmin_{\theta} \lVert
                \mathcal{T}_{\phi_e} \circ f_{\theta}(z) - y_{\phi_e}
            \rVert^2_2 + \alpha \mathrm{TV} \circ f_{\theta}(z)
        \rbrace
    \end{equation}

    where $\mathcal{T}_{\phi_e}$ is the Radon transform sampled on experimental viewing angles
    $\phi_e$, $y_{\phi_e}$ are the experimental measurements and $\alpha$ is an hyper-parameter
    balancing the regularization. In this formulation, the issue of finding an appropriate
    early stopping for denoising is replaced with the necessity to find an optimal value for
    $\alpha$, depending on the ill-posedness of the IIP at hand.

    Further experiments in \cite{baguer_computed_2020} showed improvements by initializing the
    decoder weights using the result $x_s$ of a supervised method, e.g.\ Learned Primal-Dual
    reconstruction \cite{adler_learned_2018}:

    \begin{equation}
    \label{IDIP}
        \theta_0 = \argmin_{\theta} \lVert f_{\theta}(z) - x_s \rVert^2_2
    \end{equation}

    However, replacing random weights of the decoder with such a supervised initialization
    restricts the applicability of the approach to the supervised use case, i.e.\ fixed
    acquisition parameters. To avoid this, Baguer et al.\ use for this initialization a
    reconstruction computed with standard data-free techniques instead, e.g.\ FBP.

  \subsection{Generative Latent Optimization (GLO)}

    The GLO technique uses a decoder network and a set of \textit{learnable} unit noise
    vectors, i.e.\ latent codes, where each code is associated to a single individual of the
    unsupervised training dataset. During training, the gradient is back-propagated both on the
    weights of the decoder and on the latent codes.

    Formally it consists in mapping one freely \textit{learned} unit latent vector $z_i$ to
    each image $x_i$ of a given dataset $X = \lbrace x_1, ..., x_N \rbrace$, through a decoder
    network $f_\theta$ with parameters $\theta$, e.g.\ a DCGAN
    \cite{radford_unsupervised_2016}. With gradient descent, it comes to jointly learn decoder
    network parameters and latent codes $Z = \lbrace z_1, ..., z_N \rbrace$, both randomly
    initialized, with the objective function:

    \begin{equation}
    \label{glo_objective}
        Z^*, \theta^* = \argmin_{Z, \theta}
        \frac{1}{N} \sum_{i=1}^N \lVert f_\theta(z_i) - x_i \rVert^2_2
        \quad s.t. \quad \lVert z_i \rVert_2 = 1
    \end{equation}

    It is mentioned in \cite{bojanowski_optimizing_2018} that, as is conventionally done for
    training GANs, vectors sampled from an n-dimensional normal distribution are close to the
    surface of an n-sphere with radius $\sqrt{n}$. Following this idea, Bojanowski et al.\
    propose to constrain the latent vectors to the unit n-sphere. It facilitates the exploration
    of the latent space and allows data transformation via geodesic interpolation. This
    constraint does not deteriorates the optimization and is numerically simple as it consists
    in projecting the latent vectors onto the unit n-sphere, i.e.\ normalizing, after each
    back-propagation.

    Experiments in \cite{bojanowski_optimizing_2018} and \cite{azuri_generative_2021} have
    shown that the GLO approach does not suffer from mode collapse and significantly outperform
    GANs when trained on small datasets. However, like Variational Auto-Encoders (VAE)
    \cite{kingma_auto-encoding_2014}, the quality of generated samples deteriorates when the
    data variability overcomes the network capacity \cite{theis_note_2016}.

  \subsection{Conditioning GLO (cGLO)}

    Following GAN inversion literature \cite{xia_gan_2023}, one could try to search the latent
    space to find the best candidate latent vector so that the output of the decoder is close
    to experimental observations.  However, attempting to invert GLO, i.e.\ sampling or
    exploring the manifold entailed by latent vectors learned through Eq.\
    \eqref{glo_objective}, is not straightforward \cite{hoshen_non-adversarial_2019,
    xiao_optimizing_2019}. The latent vectors are not uniformly distributed on the surface of
    the n-sphere. Furthermore, the decoder generation quality rapidly worsens when evaluated on
    vectors outside geodesic interpolation lines between pairs of latent vectors. In other
    words, the manifold on which the trained decoder can produce plausible images can be
    described as a complete graph, heterogeneously distributed on the surface of the unit
    n-sphere.

    Instead of inverting GLO, our method cGLO uses a reparameterization of the IIP, like DIP
    \cite{ulyanov_deep_2018}, to benefit from the structural bias of the convolutional decoder.
    However, the objective function described in Eq.\ \eqref{conditioned_dip}, is cast into the
    framework of GLO \cite{bojanowski_optimizing_2018}, detailed in Eq.\ \eqref{glo_objective}.

    At examination time, without prior training, the $K$ slices $\Bar{X} = \lbrace \Bar{x}_1,
    ..., \Bar{x}_K \rbrace$ from one or several patients are reconstructed together such that
    the parameters $\theta$, and the latent unit vectors $\Bar{Z} = \lbrace \Bar{z}_1, \, ...,
    \Bar{z}_K \rbrace$ are jointly learned through:

    \begin{equation}
    \label{cglo_objective}
    \Bar{Z}^*, \theta^* = \argmin_{\Bar{Z}, \theta}
    \frac{1}{K} \; \sum_{i=1}^K \lVert
        \mathcal{T}_{\phi_e} \circ f_\theta(\Bar{z}_i) - y_{i, \: \phi_e}
    \rVert^2_2
    \quad s.t. \quad \lVert \Bar{z}_i \rVert_2 = 1
    \end{equation}

    \begin{equation}
    \label{cglo_reconstruction}
    \Bar{X}^* = f_{\theta^*}(\Bar{Z}^*)
    \end{equation}

    where $\mathcal{T}_{\phi_e}$ is the Radon transform sampled on experimental viewing angles
    $\phi_e$ and $y_{i, \: \phi_e}$ are the experimental profiles obtained from slice
    $\Bar{x}_i$. This method comes with two improvements: (i) it avoids the overfitting issue
    of DIP because of the shared decoder between the entire set of slices $\Bar{X}^*$, given
    sufficient slices to reconstruct, and (ii) leads to better reconstructions as adding
    supplementary experimental profiles to the objective tends to stabilize the optimization
    process detailed in Eq.\ \eqref{cglo_objective}. Furthermore, the set of profiles can
    easily be augmented by linear interpolation along the vertical axis, i.e.\ the axis
    orthogonal to the slices plane. In this case, reconstructions are conducted using the
    entire augmented set, then reconstructions corresponding to interpolated measurements are
    discarded. The impact of the number of profiles, $K$, on performances is detailed in the
    ablation study in Appendix C.

    Given availability of a training set of full-dose CT scans, i.e.\ a set of slices $X =
    \lbrace x_1, ..., x_N \rbrace$, additional prior knowledge may be enforced by learning an
    appropriate initial set of parameters $\theta_0$, for the decoder $f_\theta$, in an
    unsupervised manner following Eq.\ \eqref{IDIP} and Eq.\ \eqref{glo_objective}:

    \begin{equation}
    \label{cglo_init}
        \theta_0 = \argmin_{Z, \theta}
        \frac{1}{N} \sum_{i=1}^N \lVert f_\theta(z_i) - x_i \rVert^2_2
        \quad s.t. \quad \lVert z_i \rVert_2 = 1
    \end{equation}

    It is expected for the set of parameters $\theta_0$ to be closer to $\theta^*$ than random
    weights, thus easing the optimization of Eq.\ \eqref{cglo_objective}, especially when
    $\phi_e$ is very sparse. The latent vectors from Eq.\ \eqref{cglo_init} are not reused so
    that the set $\Bar{Z}$ of Eq.\ \eqref{cglo_objective} is always randomly initialized.

\section{Experiments}

  Experiments have been conducted on two commonly used datasets which are publicly available on
  The Cancer Imaging Archive (TCIA) platform \cite{clark_cancer_2013}. The methods involved in
  these experiments are: (a) cGLO, FBP \cite{deans_radon_1983} and DIP
  \cite{baguer_computed_2020} \textit{without} training data, and (b) cGLO, cSGM
  \cite{song_solving_2022} and MCG \cite{chung_manifold_2022} \textit{with} training
  sub-datasets of varying sizes.  Sub-datasets are limited to 35\% portions of the initial
  datasets because the tested methods did not achieve significant performance gain above this
  threshold. All the methods are compared with the same numbers of experimental viewing angles:
  9, 23 and 50. The DPS method \cite{chung_posterior_2022} is omitted as it is equivalent to
  MCG for experiments conducted in this work, i.e.\ full-dose sparse CT. Implementation details
  concerning our decoder network architecture and hyperparameter values for each method are
  provided at the end of the section.

  \subsection{Datasets}

    The Lung Image Database Consortium (LIDC) \cite{armato_iii_lung_2011} image collection was
    collected using lung cancer screenings from 1018 patients. It consists of thoracic CT scans
    completed with radiologists annotations for nodule segmentation. The Low-dose CT image and
    projection dataset (LDCT) \cite{moen_low-dose_2021} is comprised of 299 CT scans of patient
    heads, chests and abdomens and their respective (full) clinical doses. In addition, the
    LDCT dataset also includes simulated reduced doses, by Poisson noise insertion, and the
    location and diagnosis for positive findings. In the presented experiments, only CT scan
    slices from both datasets are used. Depending on the optimization process, slices are
    either directly compared using a pixel-wise metric, or indirectly via simulated
    measurements obtained from the application of the same forward operator, i.e.\ a full dose
    parallel-beam geometry with viewing angles equally distributed across 180 degrees.
    Similarly to Song et al.\ \cite{song_solving_2022} and Chung et al.\
    \cite{chung_manifold_2022}, our approach treats every slice independently. Therefore,
    though 3D scan geometric parameters such as pixel spacing and slice thickness may vary
    across patients, slices are not resampled, such that 3D volumes can have different number
    of slices and spatial scales. The native resolution of slices is 512$\times$512 pixels for
    both datasets, but the LIDC slices are downsized to a resolution of 320$\times$320 pixels.
    For experiments to be representative of various realistic situations with sparse or
    abundant data, training sub-datasets consisting of portions of the LIDC and LDCT
    collections are prepared. Details of such datasets are given in table \ref{tab:data}. The
    LIDC and the LDCT test sets consist of five scans respectively totaling 791 and 1089
    slices. Regarding experiments involving training sub-datasets, Peak Signal to Noise Ratio
    (PSNR) and Structural SIMilarity (SSIM) metrics are estimated on the total test set where
    each 3D volume is reconstructed independently, meaning that five independent
    reconstructions are computed. However, experiments conducted with cGLO and DIP, without any
    training datasets, are tested on a restricted set of 100 slices from 1 patient due to the
    computational time of the DIP approach. For these latter experiments, reconstructions
    computed with FBP and associated quantitative metrics are added as a baseline.

    \begin{table}[h]
        \begin{subtable}[h]{0.45\textwidth}
            \centering
            \begin{tabular}{l | l | l}
            Portion & \# of scans & \# of slices \\
            \hline \hline
            35\% & 350 & 65 033 \\
            10\% & 100 & 17 898 \\
            2\% & 20 & 2 906
           \end{tabular}
           \caption{LIDC sub-datasets}
           \label{tab:lidc}
        \end{subtable}
        \hfill
        \begin{subtable}[h]{0.45\textwidth}
            \centering
            \begin{tabular}{l | l | l}
            Portion & \# of scans & \# of slices \\
            \hline \hline
            35\% & 103 & 16 924 \\
            10\% & 29 & 4 989 \\
            2\% & 6 & 930
           \end{tabular}
           \caption{LDCT sub-datasets}
           \label{tab:ldct}
        \end{subtable}
        \caption{
            Sub-datasets used for training with their corresponding number
            of 3D scans and slices. 3D scans are always fully included.
        }
        \label{tab:data}
    \end{table}

  \subsection{Implementation details}

\begin{figure}[h]
    \centering
    \includegraphics[width=\textwidth]{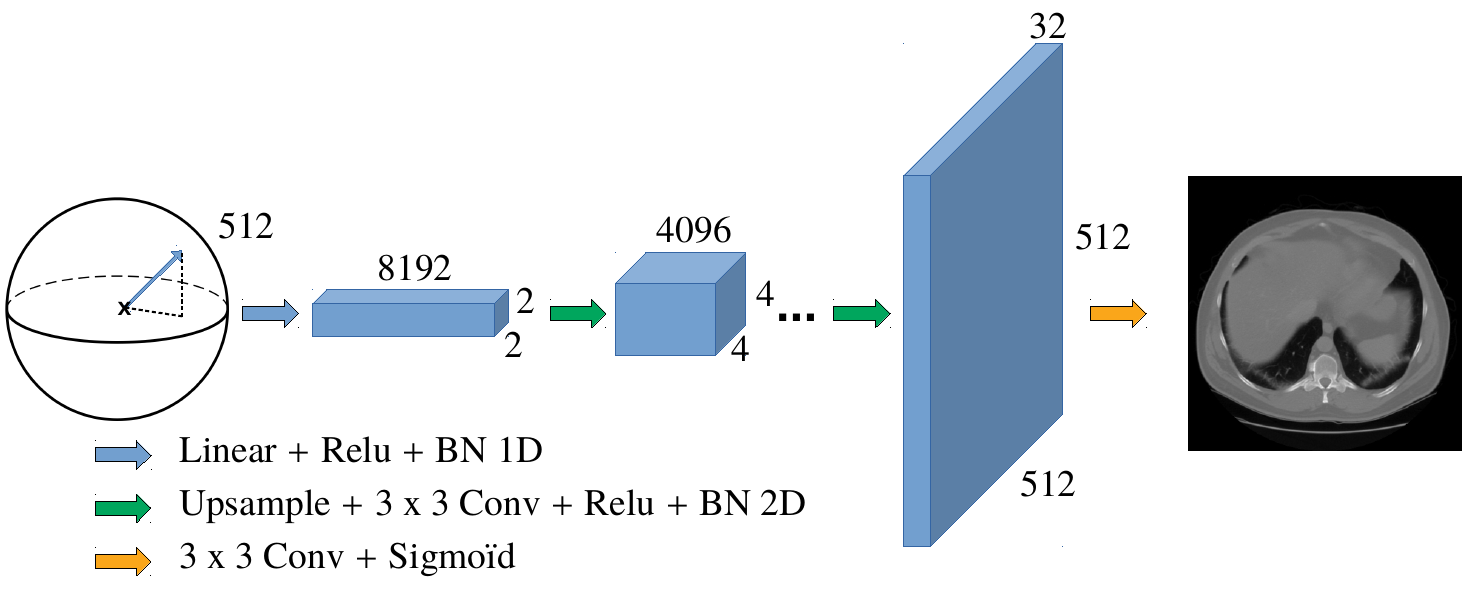}
    \caption{
        Generator DCGAN-like architecture, latent dimension (512) and final
        resolution (512$\times$512) corresponding to the LDCT sub-datasets experiments.
    }
    \label{fig:decoder}
    \end{figure}

    In the presented experiments, the geometry of the decoder's latent space is set to the
    unit n-sphere, similarly to state-of-the art representation learning approaches
    \cite{bojanowski_optimizing_2018}.  The latent space dimension is set to 320 for
    experiments on the LIDC dataset and 512 for the LDCT dataset. Being a reference in the
    literature of generative models, the Deep Convolutional Generative Adversarial Network
    (DCGAN) \cite{radford_unsupervised_2016} architecture is used for the decoder. As
    illustrated in figure \ref{fig:decoder}, the latent code firstly goes through a linear
    block and is mapped to a feature map of shape $2 \, \times \, 2 \, \times \,
    \mathbf{C}$, with $\mathbf{C}$ the desired number of channels for the first
    convolutional layer. From layer to layer the input is upscaled by a factor of 2 (last
    layer upscale factor may vary to fit the desired output image resolution) and the
    number of channels is cut by half. Hence, the image resolution, the latent dimension
    and $\mathbf{C}$ completely define the structure of the decoder. The number of input
    channels $\mathbf{C}$ is set to 8192 for both experiments on the LIDC and the LDCT
    sub-datasets.

    cGLO optimizations are conducted with the Adam algorithm \cite{kingma_adam_2015}. The
    learning rates for the latent codes and the decoder weights are both set to $10^{-3}$ while
    training on CT reconstruction, i.e.\ the step described in Eq.\ \eqref{cglo_init}, and
    respectively set to $10^{-2}$ and $10^{-4}$ when reconstructing from experimental
    measurements, i.e.\ the step detailed in Eq.\ \eqref{cglo_objective}. Although learning
    rates are kept constant, batch sizes are increased along specified schedules. It leads to
    faster training and is statistically equivalent to decreasing learning rates
    \cite{smith_dont_2018}.  Measurements are augmented by a factor of 8 for reconstructions
    involving parameters initialization with training set and by factors (8, 16 and 16) for
    angles (9, 23 and 50) when no training data is used.

    DIP, cSGM and MCG hyper-parameters, model architecture, and other implementation details
    are set to what is respectively provided by Baguer et al.\ \cite{baguer_computed_2020},
    Song et al.\ \cite{song_solving_2022} and Chung et al.\ \cite{chung_manifold_2022} for
    their sparse-view CT experiments. The DIP network is initialized as described in Eq.\
    \eqref{IDIP} on a reconstruction computed with FBP.

\section{Results and discussion}

  \begin{figure}[h!]
    \centering
    \begin{subfigure}[b]{0.245\textwidth}
      \includegraphics[width=\textwidth]{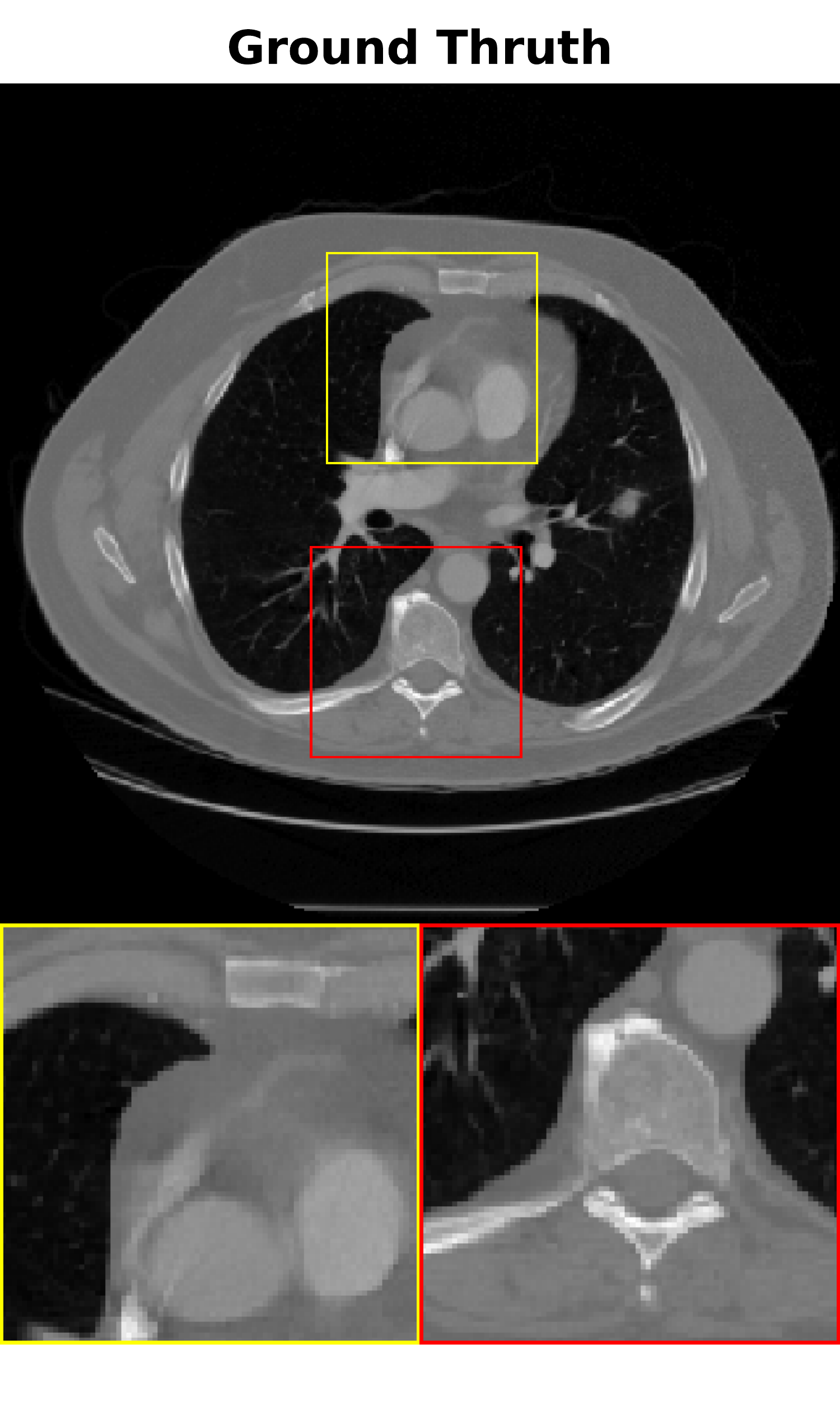}
    \end{subfigure}
    \hfill
    \begin{subfigure}[b]{0.245\textwidth}
      \includegraphics[width=\textwidth]{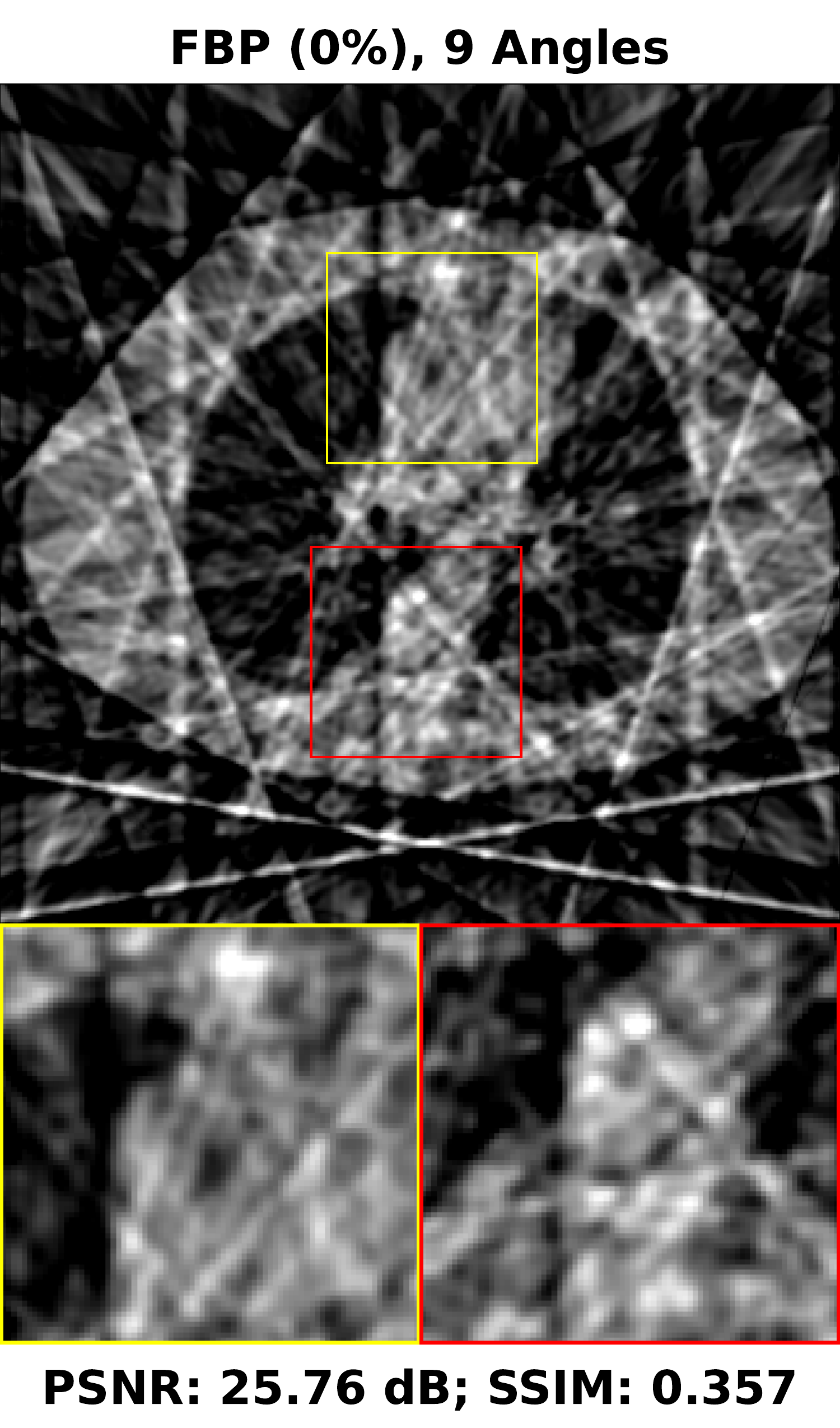}
    \end{subfigure}
    \hfill
    \begin{subfigure}[b]{0.245\textwidth}
      \includegraphics[width=\textwidth]{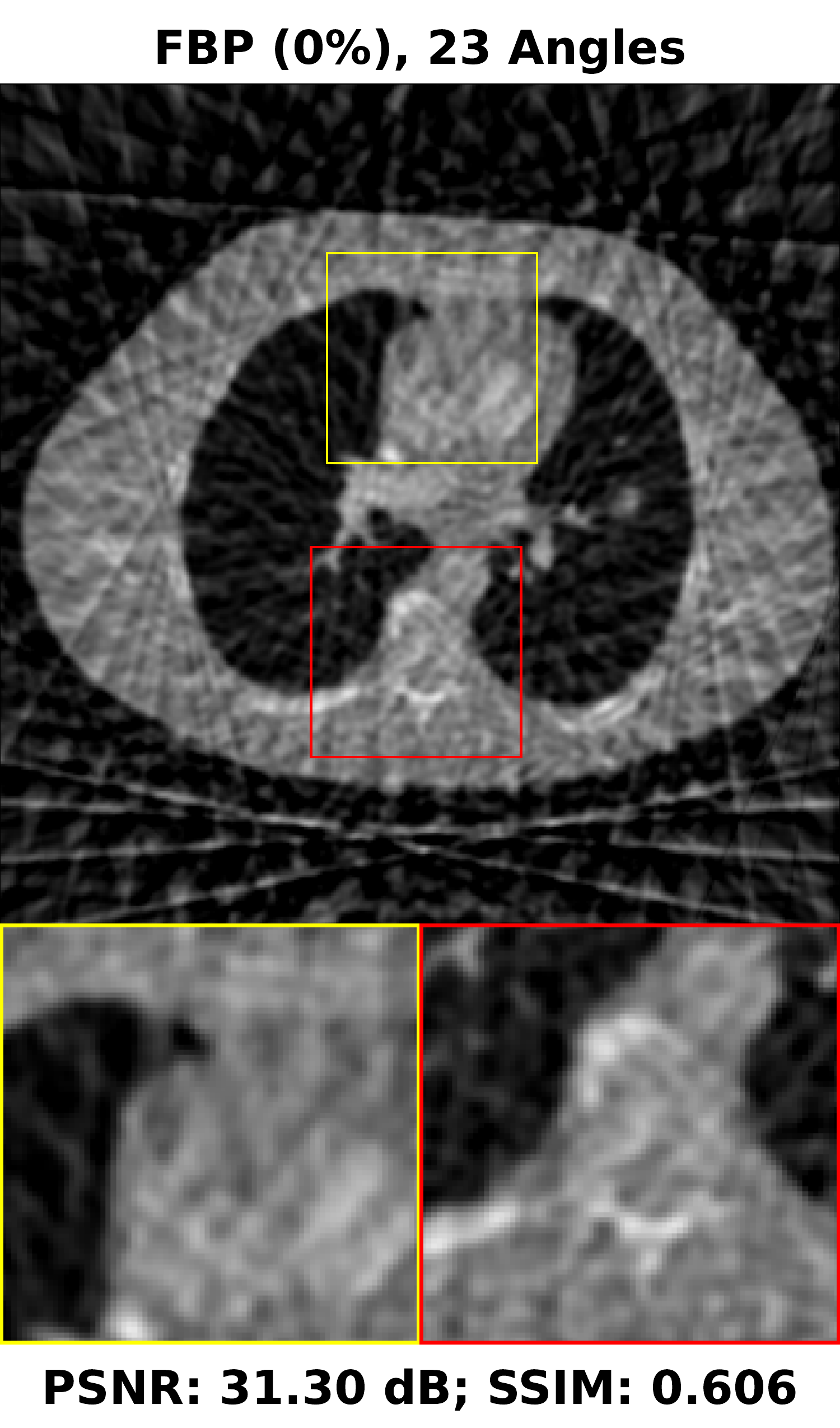}
    \end{subfigure}
    \hfill
    \begin{subfigure}[b]{0.245\textwidth}
      \includegraphics[width=\textwidth]{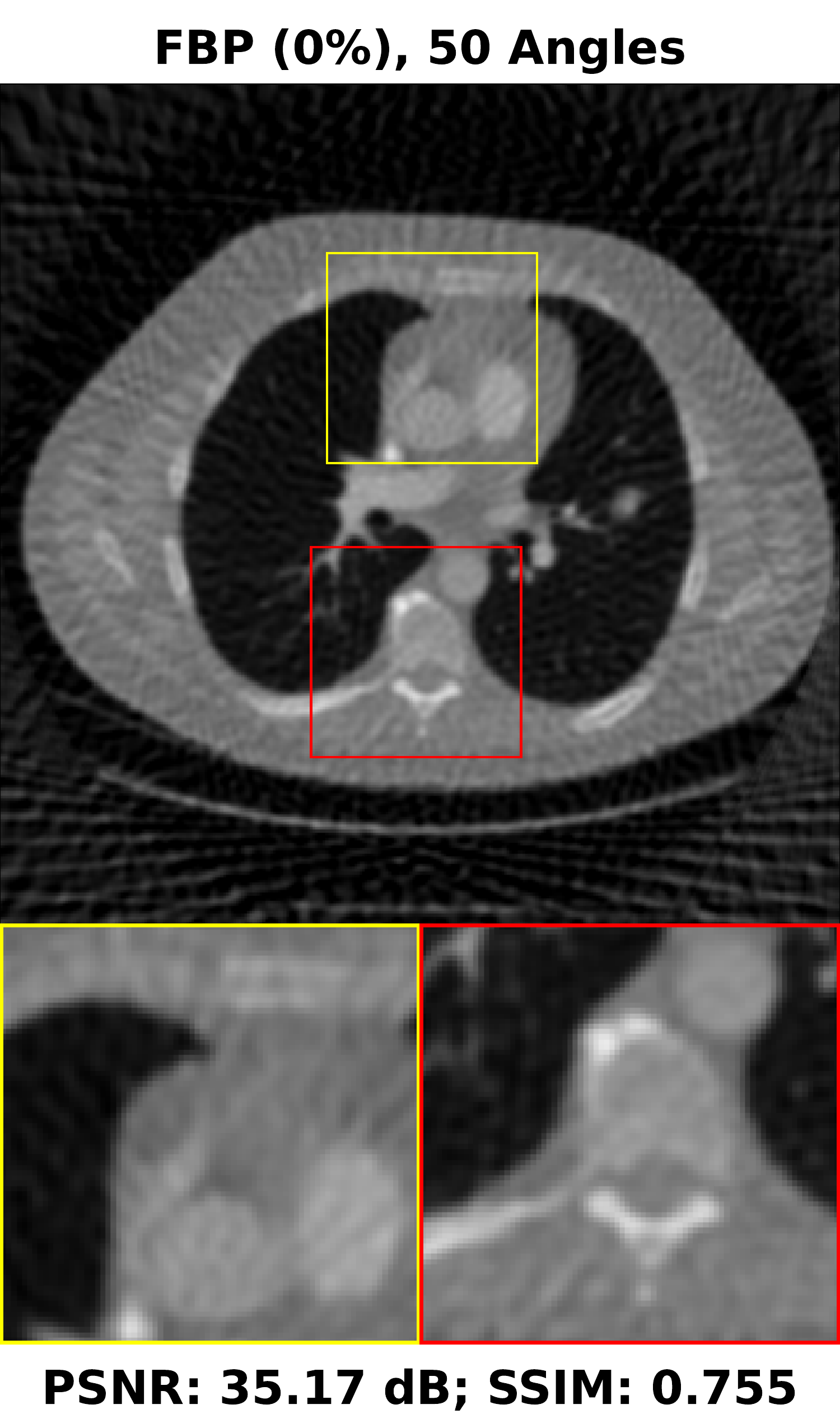}
    \end{subfigure}
    \hfill
    \begin{subfigure}[b]{0.245\textwidth}
      \includegraphics[width=\textwidth]{images/GTX.pdf}
    \end{subfigure}
    \hfill
    \begin{subfigure}[b]{0.245\textwidth}
      \includegraphics[width=\textwidth]{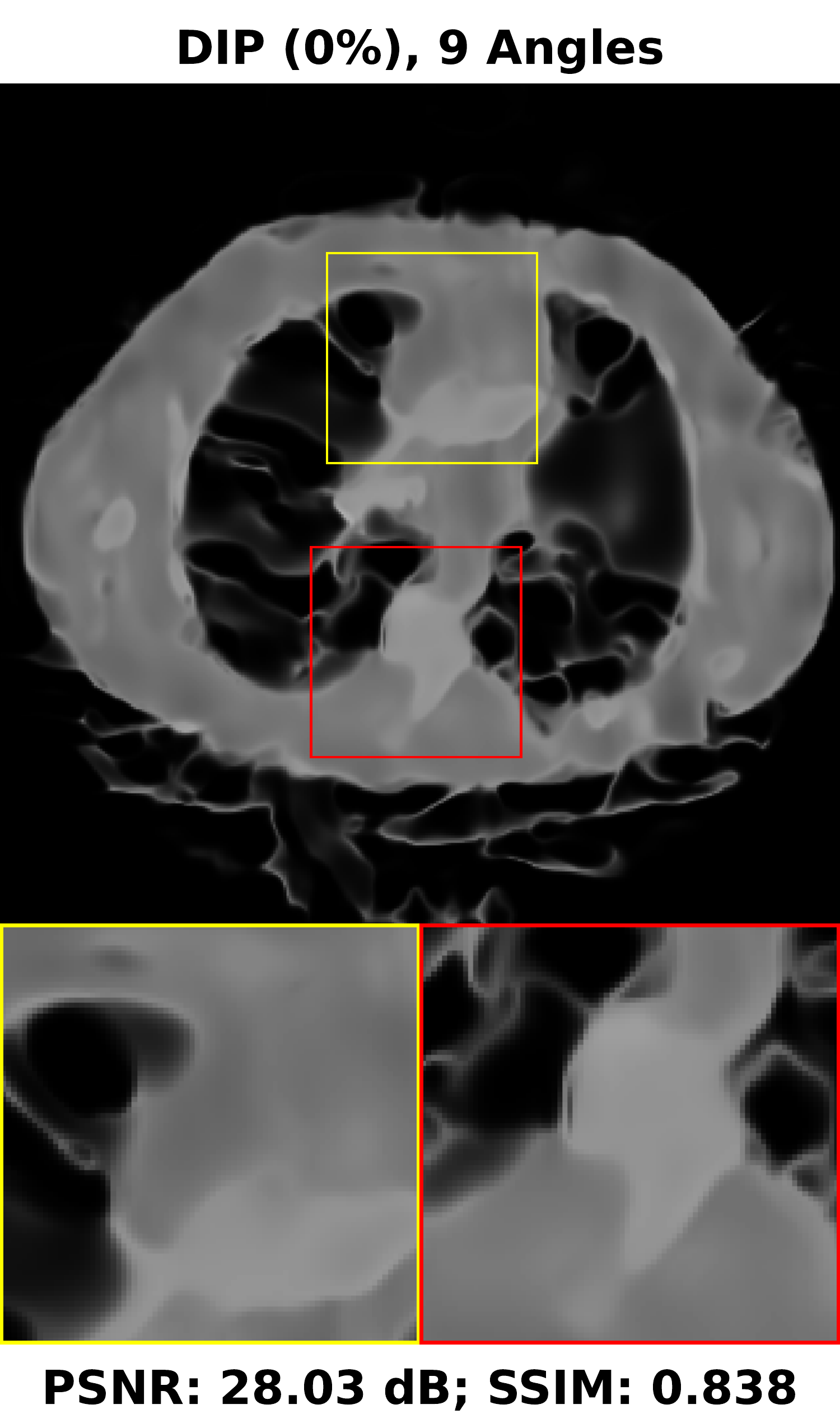}
    \end{subfigure}
    \hfill
    \begin{subfigure}[b]{0.245\textwidth}
      \includegraphics[width=\textwidth]{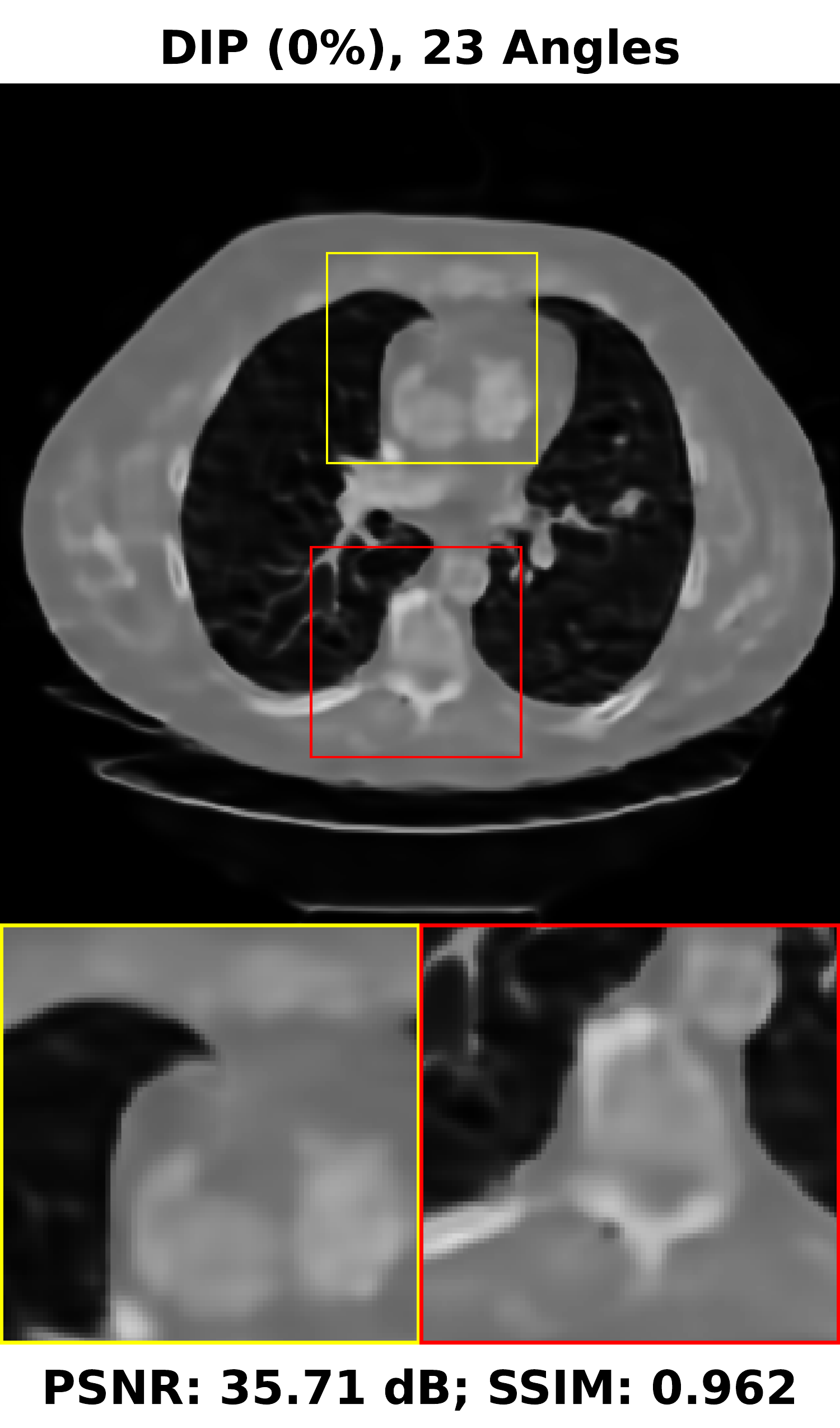}
    \end{subfigure}
    \hfill
    \begin{subfigure}[b]{0.245\textwidth}
      \includegraphics[width=\textwidth]{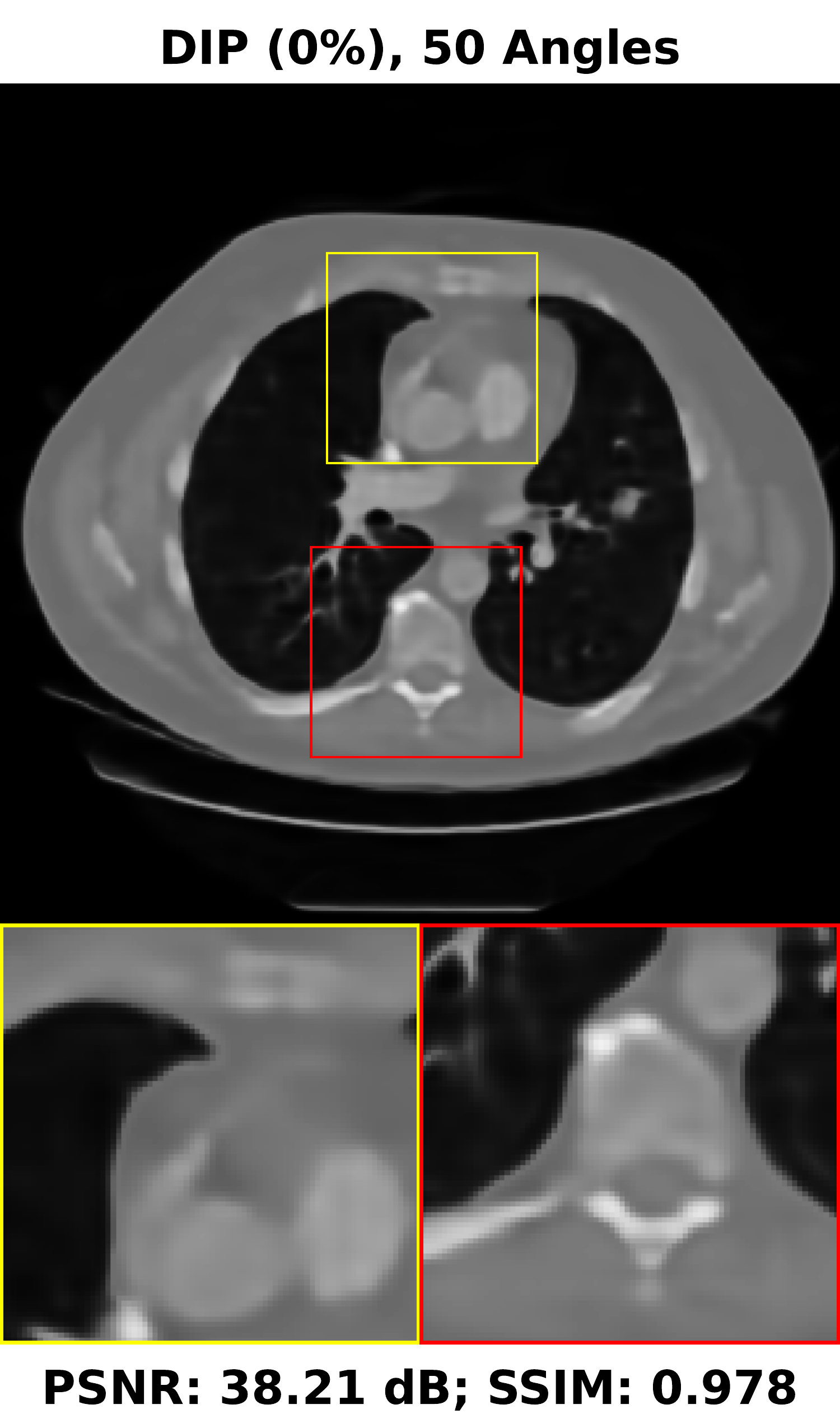}
    \end{subfigure}
    \hfill
    \begin{subfigure}[b]{0.245\textwidth}
      \includegraphics[width=\textwidth]{images/GTX.pdf}
    \end{subfigure}
    \hfill
    \begin{subfigure}[b]{0.245\textwidth}
      \includegraphics[width=\textwidth]{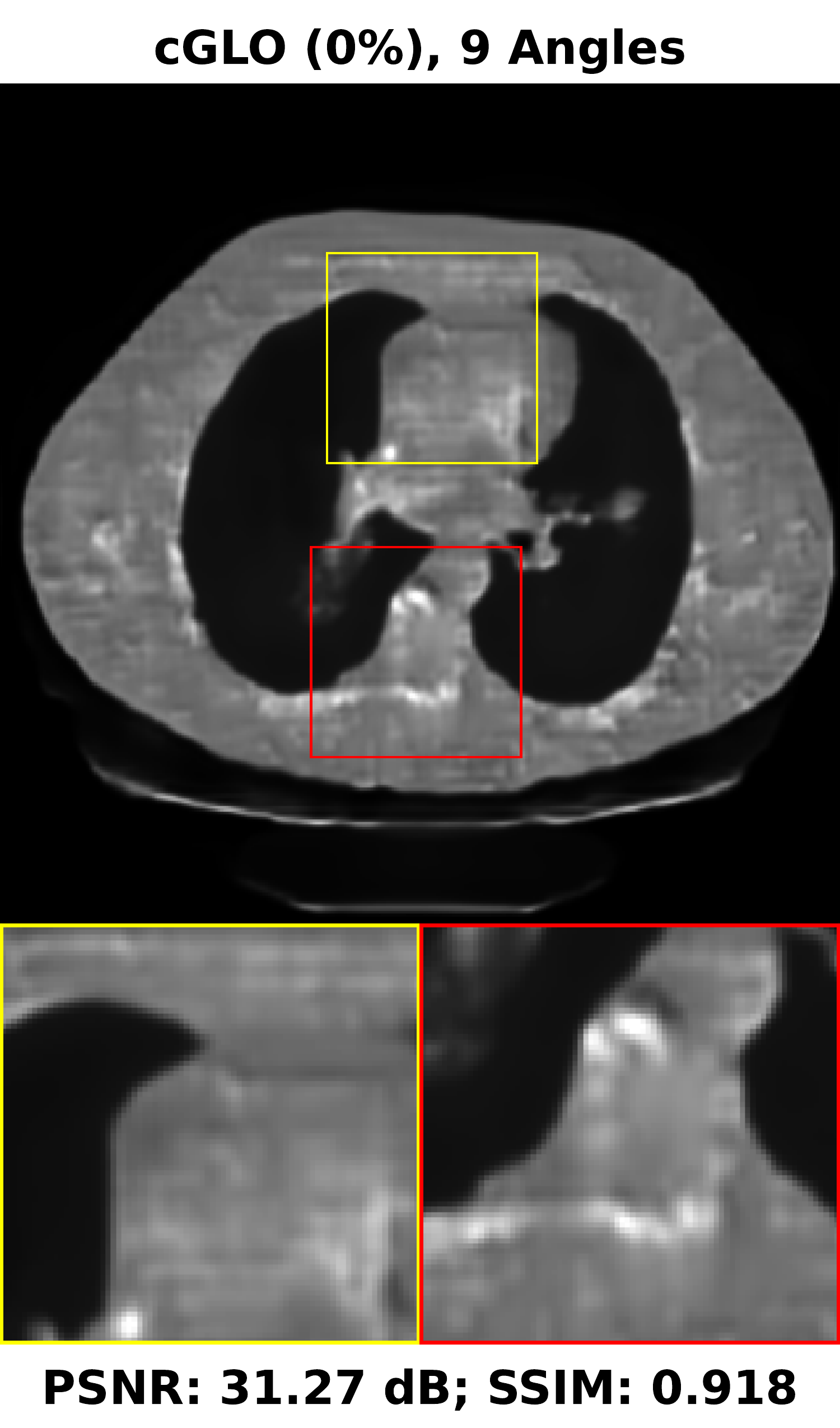}
    \end{subfigure}
    \hfill
    \begin{subfigure}[b]{0.245\textwidth}
      \includegraphics[width=\textwidth]{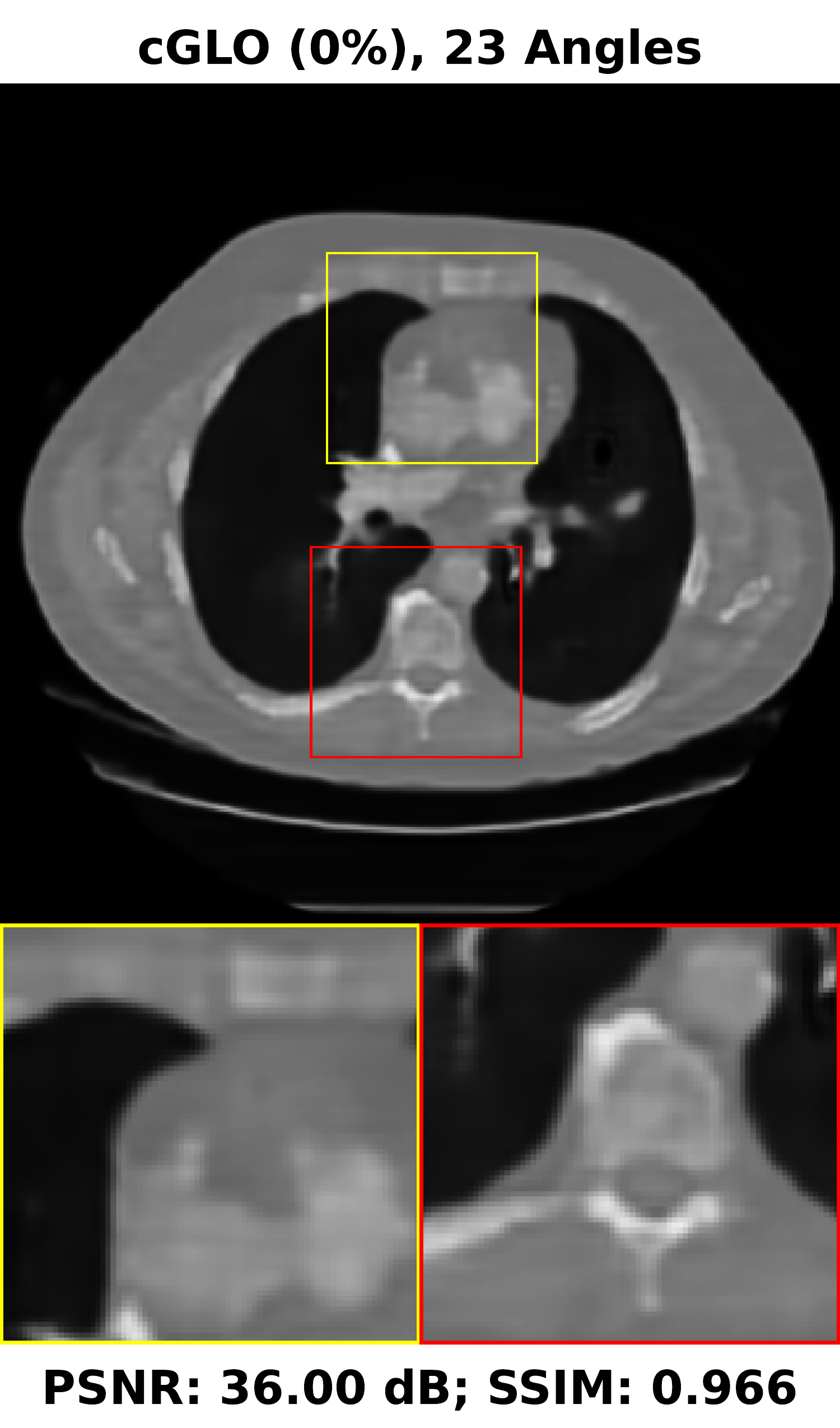}
    \end{subfigure}
    \hfill
    \begin{subfigure}[b]{0.245\textwidth}
      \includegraphics[width=\textwidth]{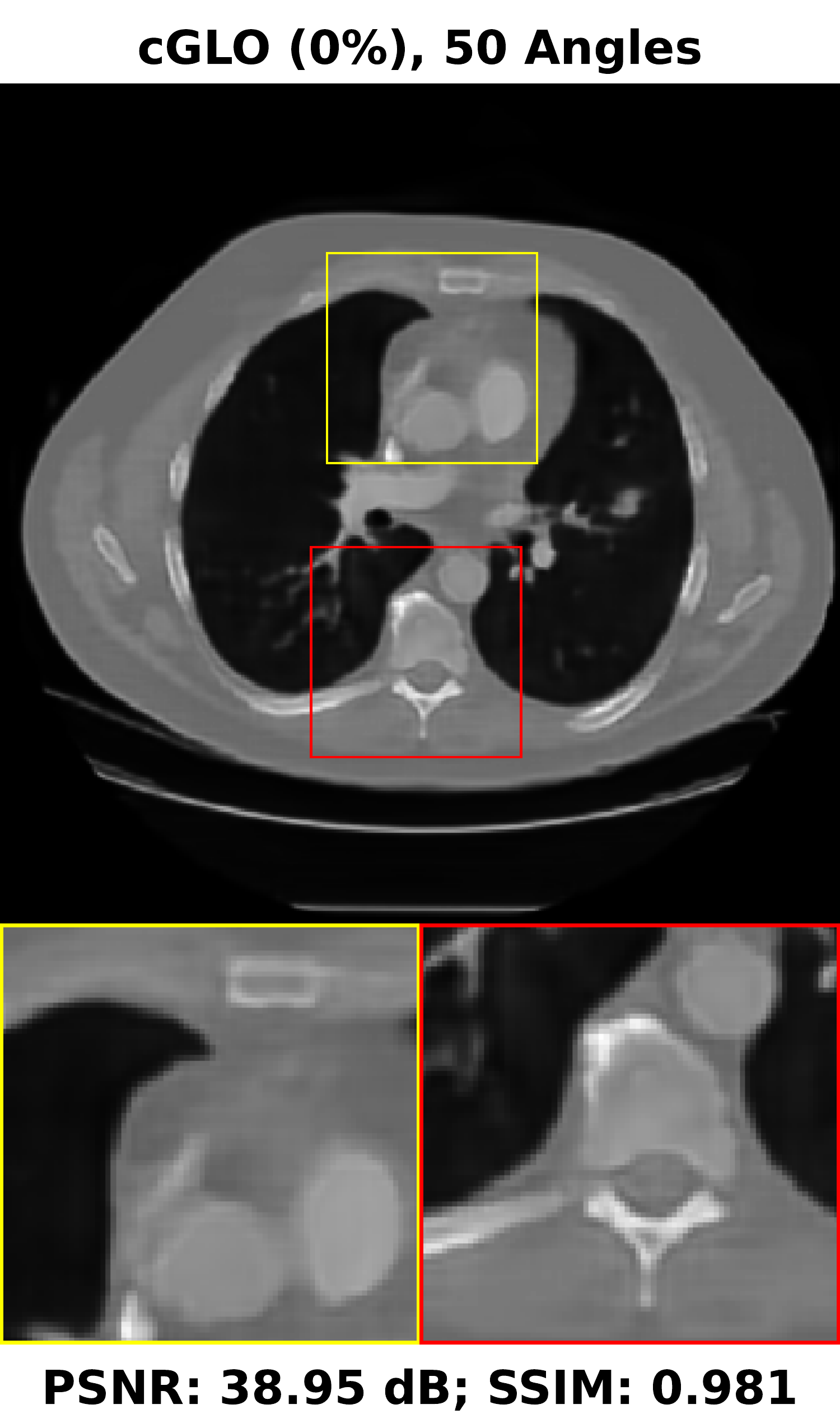}
    \end{subfigure}
    \caption{
        Examples of reconstructions given 9, 23 and 50 experimental viewing angles,
        obtained with FBP (upper row), DIP (middle row) and cGLO (lower row).
        Reconstructions are achieved without prior unsupervised training.
    }
    \label{fig:results_no_training}
  \end{figure}

  In this section, our method cGLO is compared with DIP and the conventional baseline FBP for
  reconstructions obtained \textit{without} any training dataset. Regarding this specific use
  case, DIP currently holds state-of-the-art performance.  These approaches are evaluated using
  quantitative pixel-wise (PSNR) and structural (SSIM) metrics which are detailed in Table
  \ref{tab:results_no_training}, alongside qualitative results consisting of examples of
  reconstructions from the LIDC test set and given in figure \ref{fig:results_no_training}. Our
  method is also tested against cSGM and MCG on use cases \textit{with} available training
  datasets of varying sizes. To the best of our knowledge, with prior unsupervised training,
  MCG achieves state-of-the-art reconstruction quality for sparse CT. Quantitative metrics for
  9, 23, and 50 experimental viewing angles are given in figure \ref{fig:curves} and detailed
  in Appendix D. Qualitative results consist of examples of reconstructions from cGLO, cSGM and
  MCG trained on the smallest sub-dataset, i.e.\ a 2\% portion of the LIDC dataset. These
  reconstructions are presented in figure \ref{fig:lidc_ex_2}. Additional examples, for
  experimental setups corresponding to different combinations of training dataset sizes and
  number of viewing angles are provided in Appendix E.

  \subsection{Without prior unsupervised training}

    As shown in Table \ref{tab:results_no_training}, on the use case with no training dataset
    available, our method cGLO outperforms DIP and FBP reconstruction methods for every set of
    experimental viewing angles, regarding both PSNR and SSIM metrics. Most notably given 9
    experimental viewing angles, cGLO offers a PSNR gain of $+3.15$ dB on the LIDC test set
    compared to DIP.

    \begin{table}[h!]
    \centering
    \begin{tabular}{c | c | cc | cc}
    \multirow{2}{*}{Method} &
    \multirow{2}{*}{Angles} &
    \multicolumn{2}{c|}{LIDC 320$\times$320} &
    \multicolumn{2}{c}{LDCT 512$\times$512} \\\cline{3-6}
    & & PSNR $\uparrow$ & SSIM $\uparrow$ &
        PSNR $\uparrow$ & SSIM $\uparrow$ \\
    \hline \hline
    FBP  & 9    & 25.99 \tiny $\pm 0.11$ & 0.366 \tiny $\pm 0.011$ &
                  26.31 \tiny $\pm 0.09$ & 0.454 \tiny $\pm 0.006$\\
    DIP  & 9    & \underline{28.26} \tiny $\pm 0.32$ & \underline{0.843} \tiny $\pm 0.009$ &
                  \underline{27.78} \tiny $\pm 0.39$ & \underline{0.842} \tiny $\pm 0.010$\\
    cGLO & 9    & \textbf{31.41} \tiny $\pm 0.34$ & \textbf{0.916} \tiny $\pm 0.005$ &
                  \textbf{30.28} \tiny $\pm 0.29$ & \textbf{0.906} \tiny $\pm 0.003$\\
    \hline
    FBP  & 23   & 31.46 \tiny $\pm 0.19$ & 0.605 \tiny $\pm 0.004$ &
                  30.79 \tiny $\pm 0.08$ & 0.623 \tiny $\pm 0.004$\\
    DIP  & 23   & \underline{35.50} \tiny $\pm 0.23$ & \underline{0.954} \tiny $\pm 0.002$ &
                  \underline{33.64} \tiny $\pm 0.41$ & \underline{0.935} \tiny $\pm 0.005$\\
    cGLO & 23   & \textbf{36.18} \tiny $\pm 0.39$ & \textbf{0.966} \tiny $\pm 0.002$ &
                  \textbf{35.99} \tiny $\pm 0.64$ & \textbf{0.956} \tiny $\pm 0.005$\\
    \hline
    FBP  & 50   & 35.31 \tiny $\pm 0.56$ & 0.763 \tiny $\pm 0.005$ &
                  33.11 \tiny $\pm 0.10$ & 0.724 \tiny $\pm 0.002$\\
    DIP  & 50   & \underline{37.40} \tiny $\pm 0.36$ & \underline{0.973} \tiny $\pm 0.002$ &
                  \underline{36.36} \tiny $\pm 0.31$ & \underline{0.960} \tiny $\pm 0.002$\\
    cGLO & 50   & \textbf{38.92} \tiny $\pm 0.27$ & \textbf{0.980} \tiny $\pm 0.001$ &
                  \textbf{38.75} \tiny $\pm 0.20$ & \textbf{0.973} \tiny $\pm 0.001$\\
    \hline
    \end{tabular}
    \vspace{10pt}
    \caption{
        PSNR and SSIM median $\pm$ half Inter Quartile Range (IQR) values for
        reconstructions of 100 slices from the LIDC and the LDCT test datasets given 9, 23
        and 50 experimental viewing angles. Best and second best results are respectively
        indicated as bold and underline for each angle and metric.
    }
    \label{tab:results_no_training}
    \end{table}

    Also, median PSNR and SSIM values reported in Table \ref{tab:results_no_training} indicate
    that the performance gap, at the advantage of cGLO over DIP, increases as the set of
    experimental viewing angles gets smaller. More specifically, the PSNR gap between cGLO and
    DIP more than double between experiments involving 50 ($+1.52$ dB) and 9 ($+3.15$ dB)
    viewing angles on the LIDC test set. In other words, our method is more robust for solving
    increasingly sparse CT with no prior training when compared to DIP.

    It can also be observed in figure \ref{fig:results_no_training} on the reconstruction
    example given 9 experimental viewing angles that, contrary to cGLO, DIP reproduces some of
    the deformations induced by the strong artifacts from FBP. This corroborates the larger
    difference between cGLO and DIP regarding SSIM metric for this experiment ($+0.073$). It is
    a direct effect of DIP network parameters first being initialized on the FBP
    reconstruction. However, by skipping this initialization step, DIP reconstructions and
    quantitative performances would be further degraded.

  \subsection{With prior unsupervised training}

    \subsubsection*{Quantitative performance}

      \begin{figure}[h!]

          \centering

          \begin{subfigure}[b]{\textwidth}

              \centering

              \includegraphics[width=0.29\textwidth]{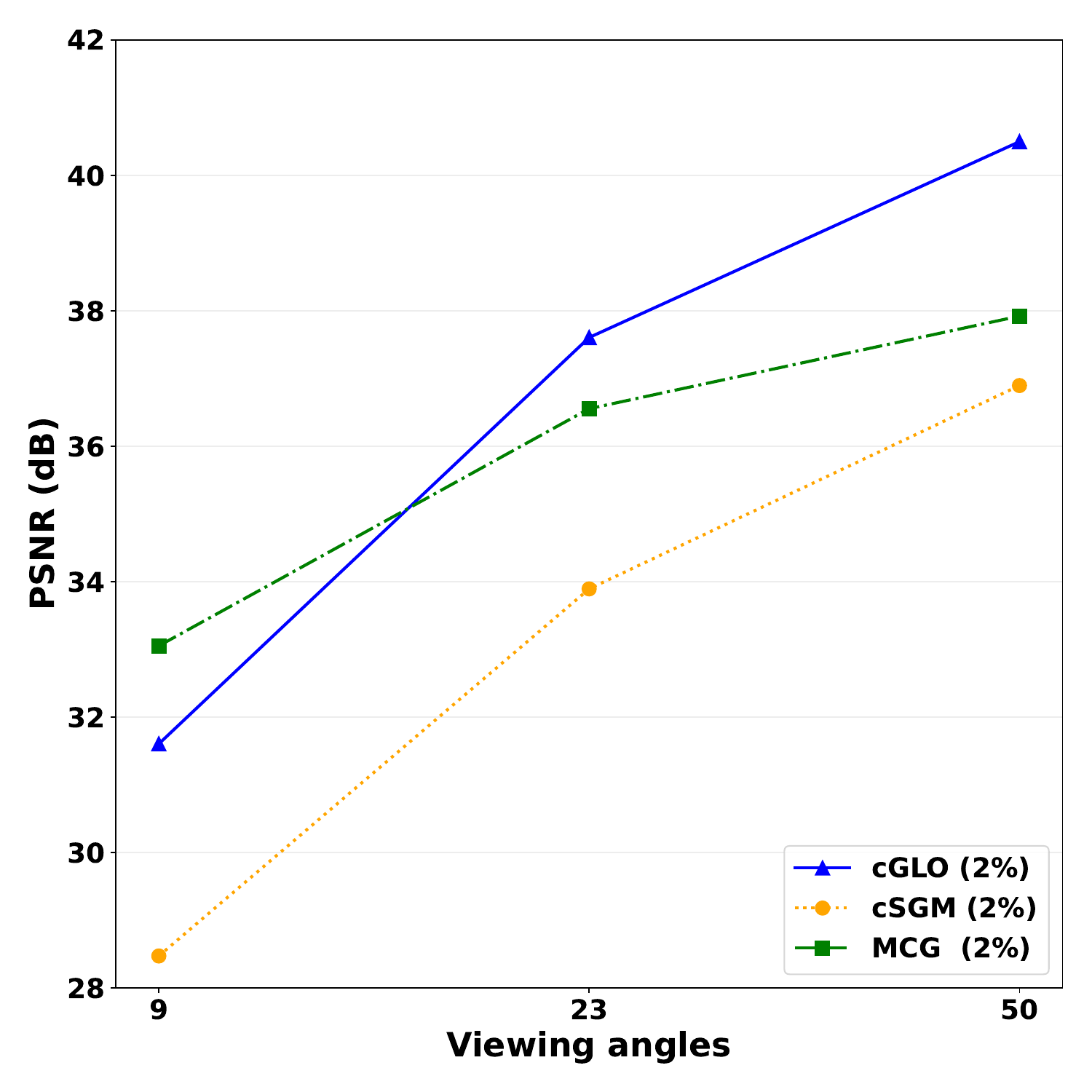}
              \includegraphics[width=0.29\textwidth]{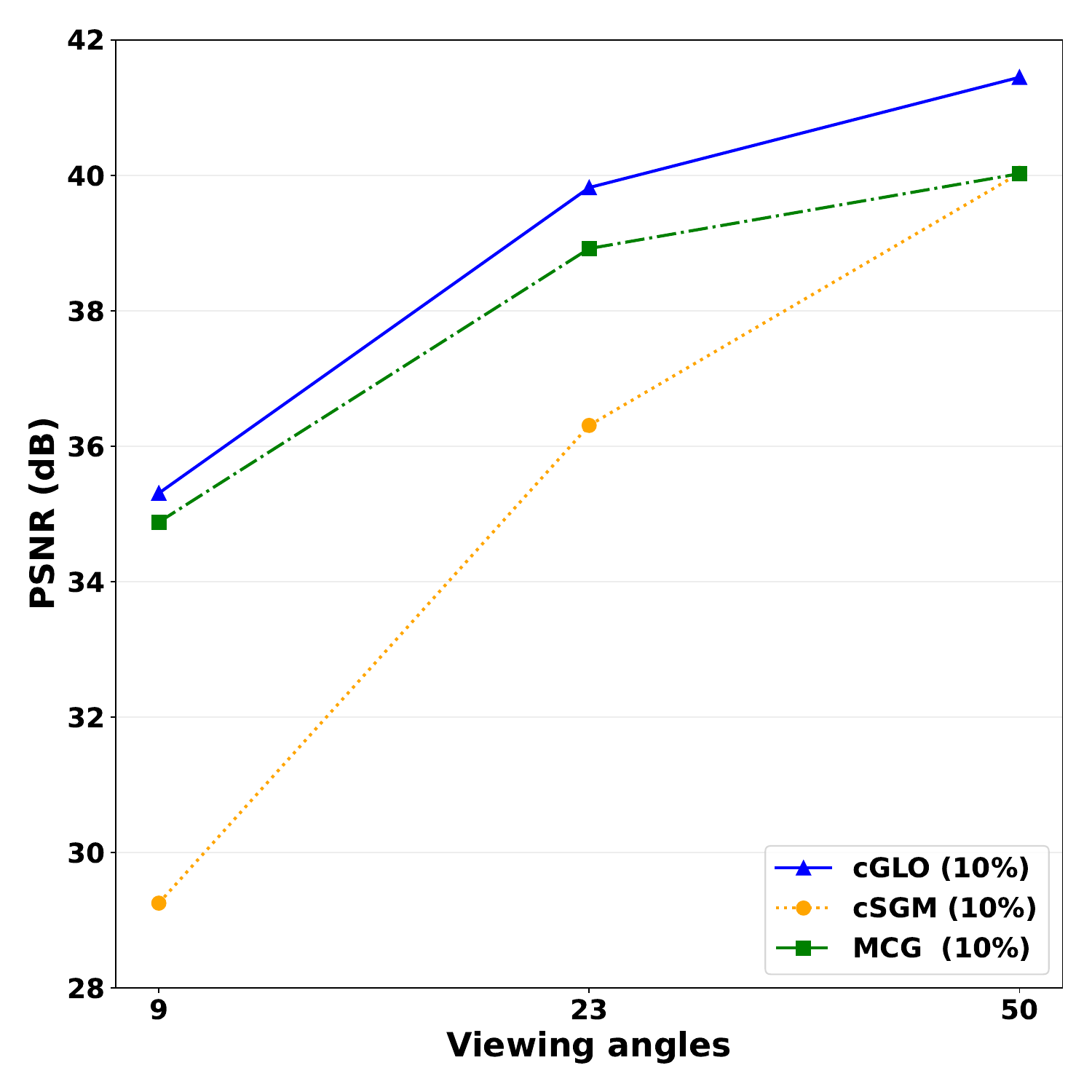}
              \includegraphics[width=0.29\textwidth]{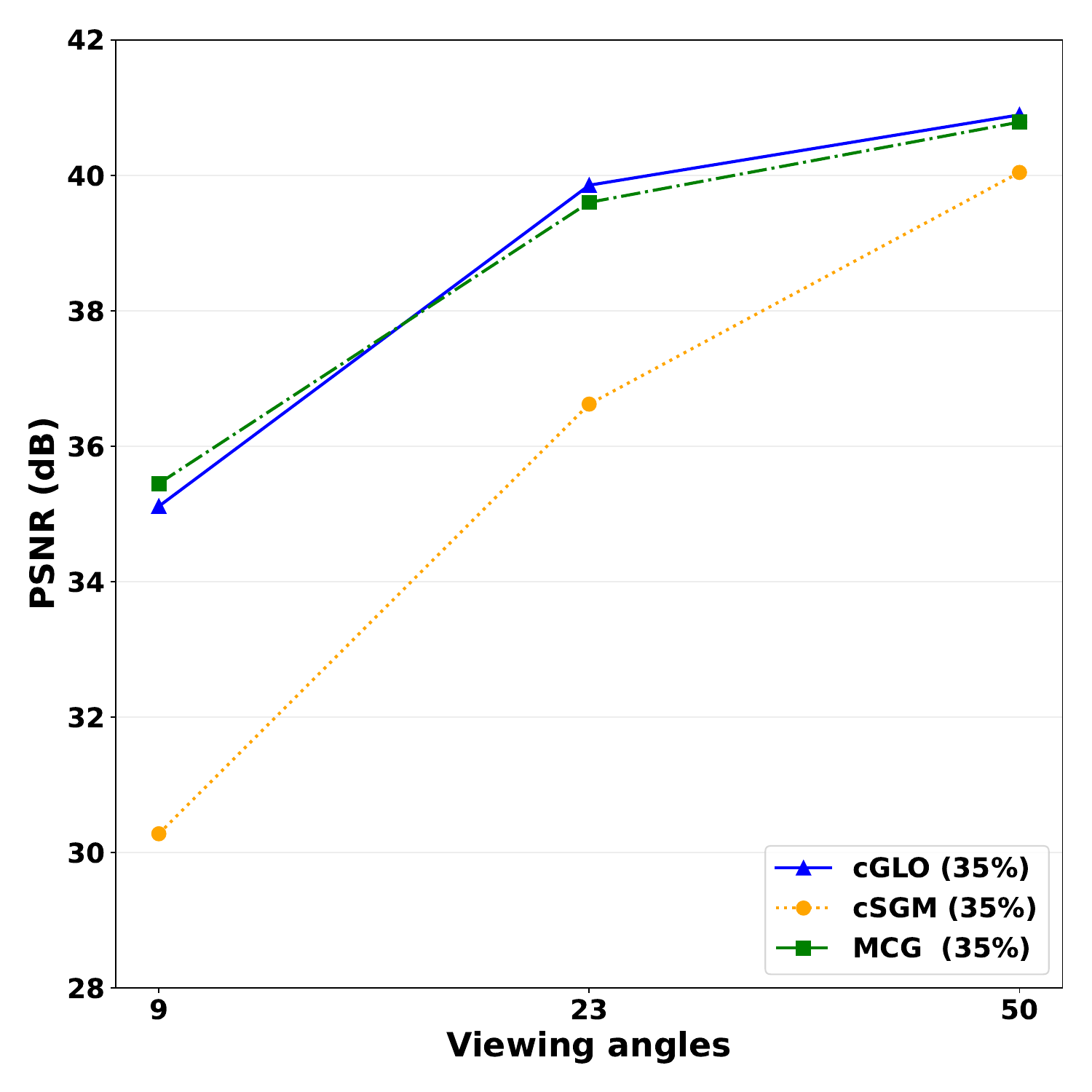}

              \includegraphics[width=0.29\textwidth]{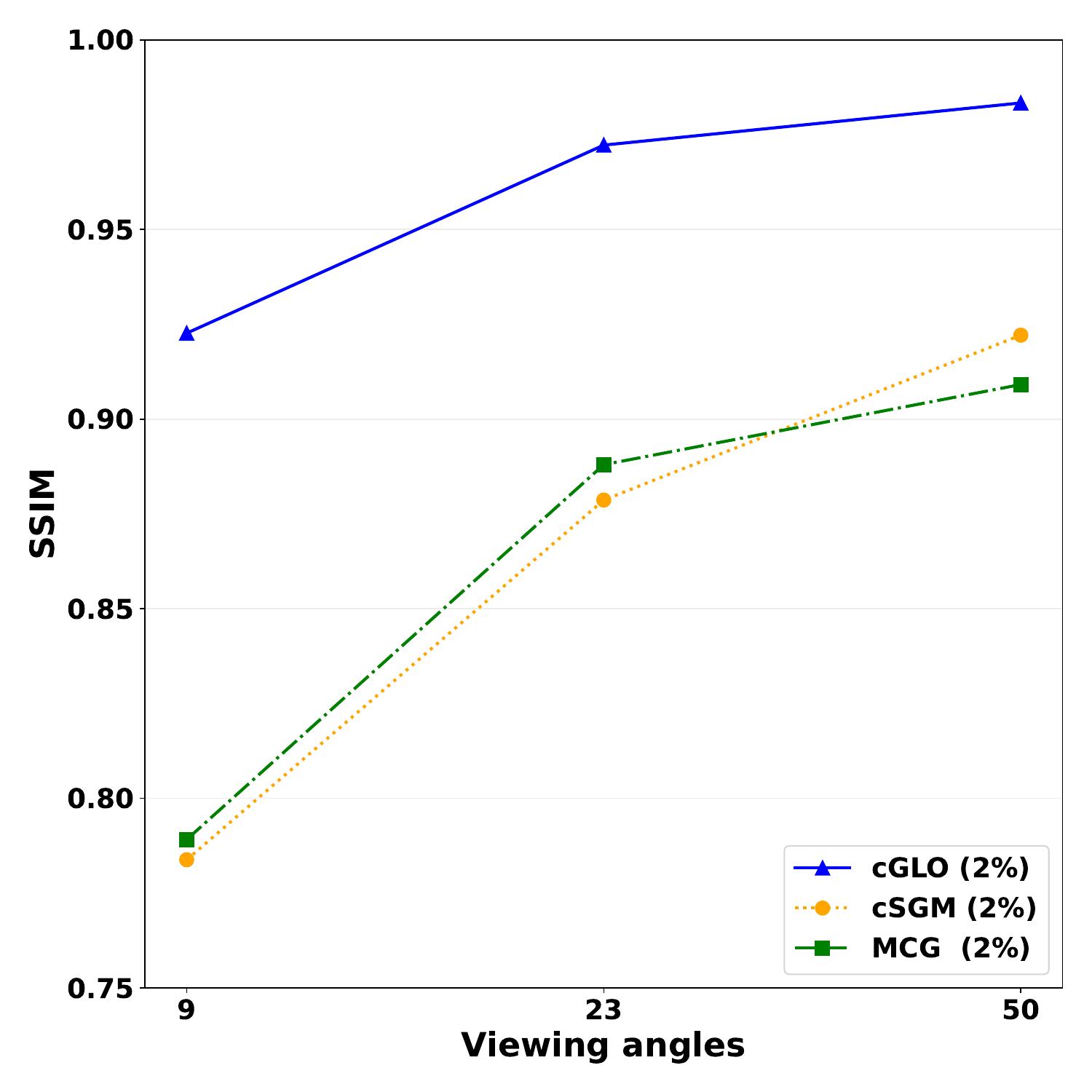}
              \includegraphics[width=0.29\textwidth]{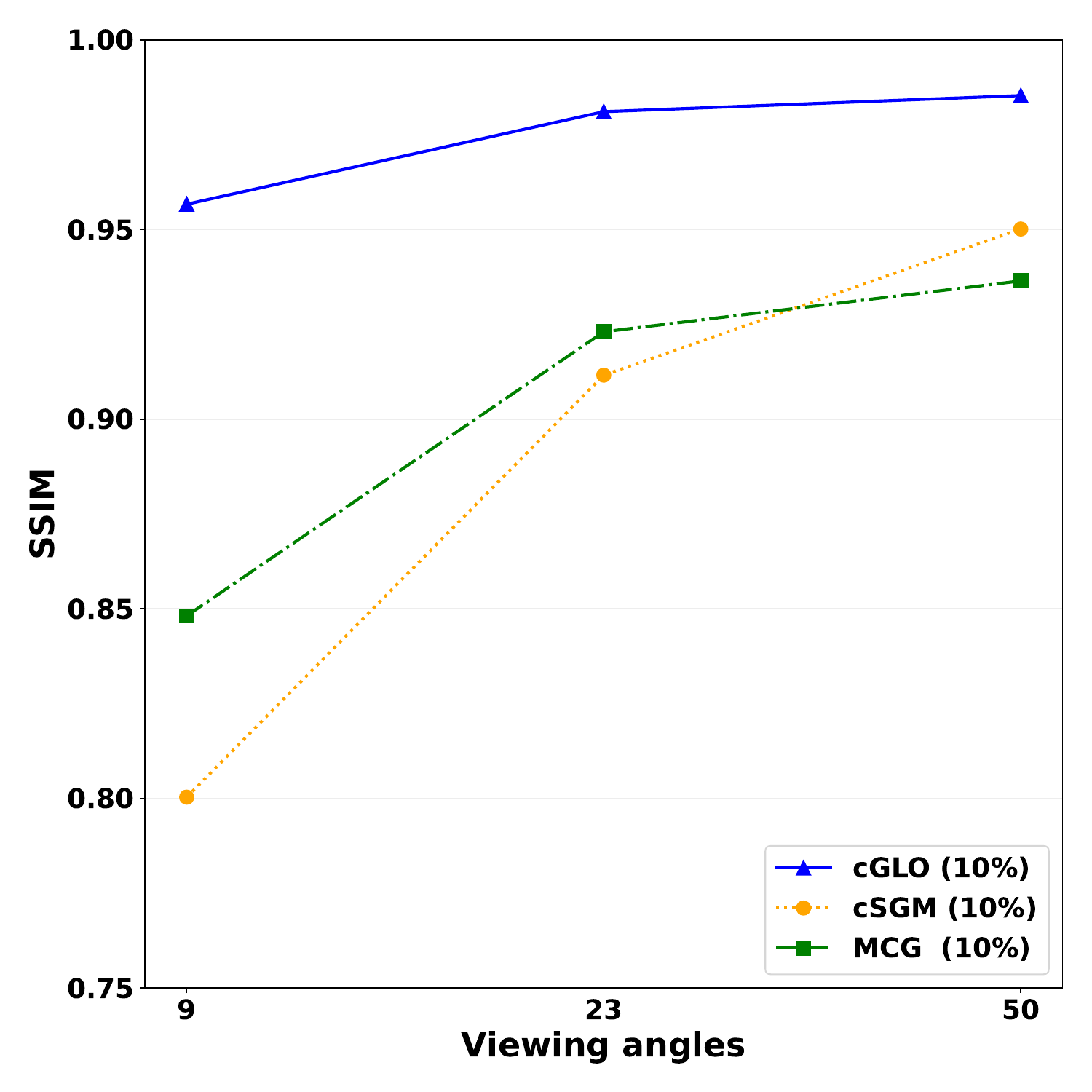}
              \includegraphics[width=0.29\textwidth]{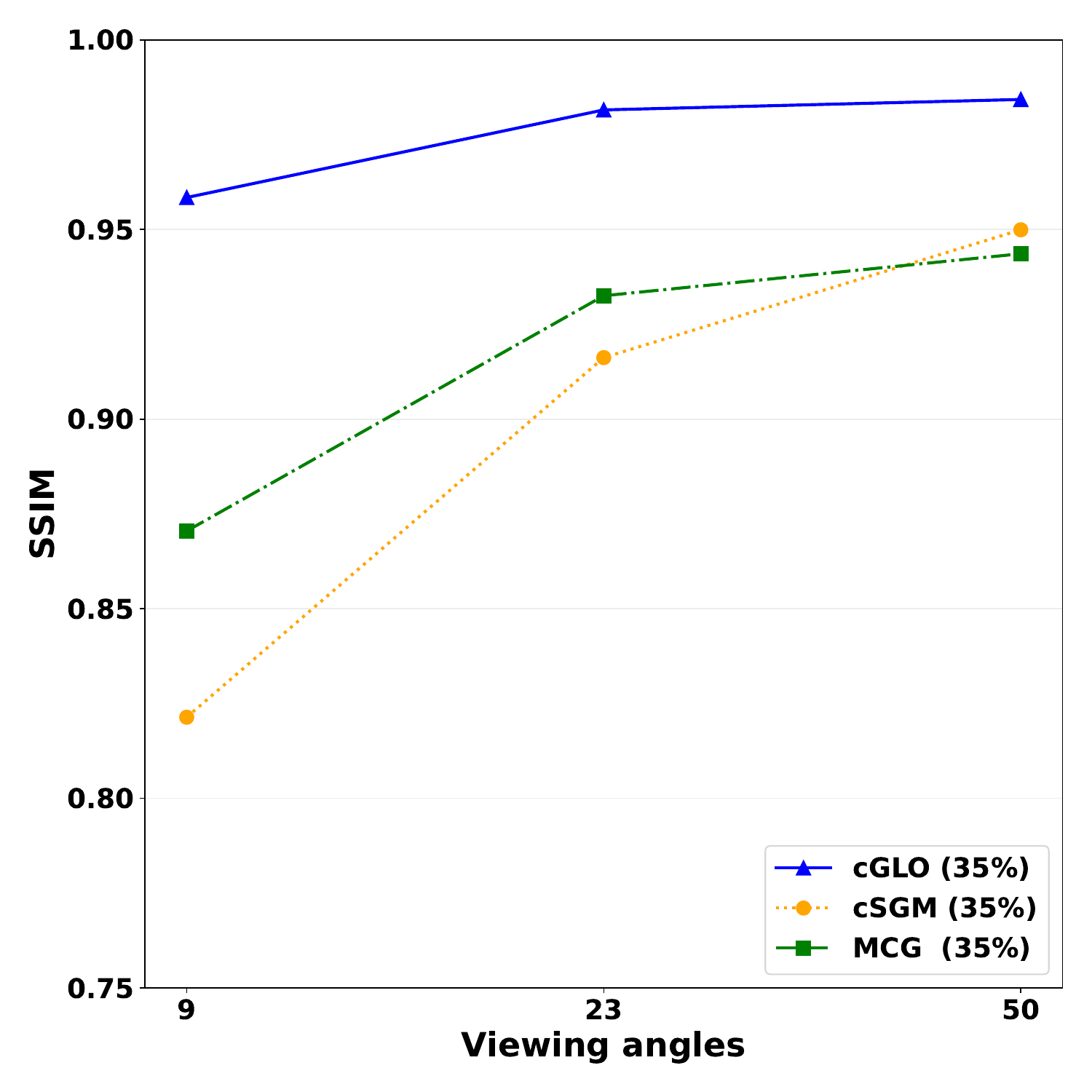}

              \caption{PSNR and SSIM - LIDC dataset}
              \label{fig:curves_lidc}

          \end{subfigure}

          \begin{subfigure}[b]{\textwidth}

              \centering

              \includegraphics[width=0.29\textwidth]{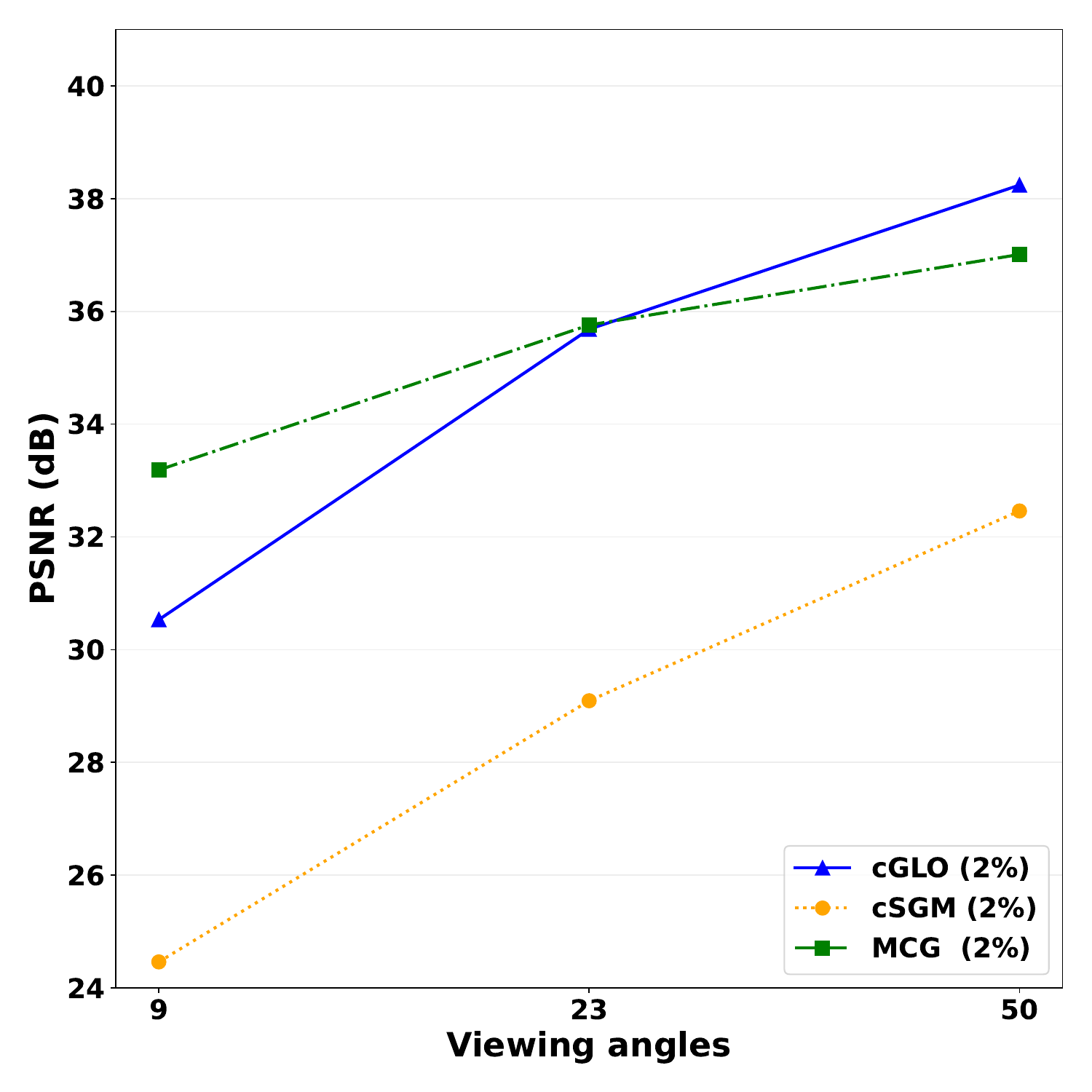}
              \includegraphics[width=0.29\textwidth]{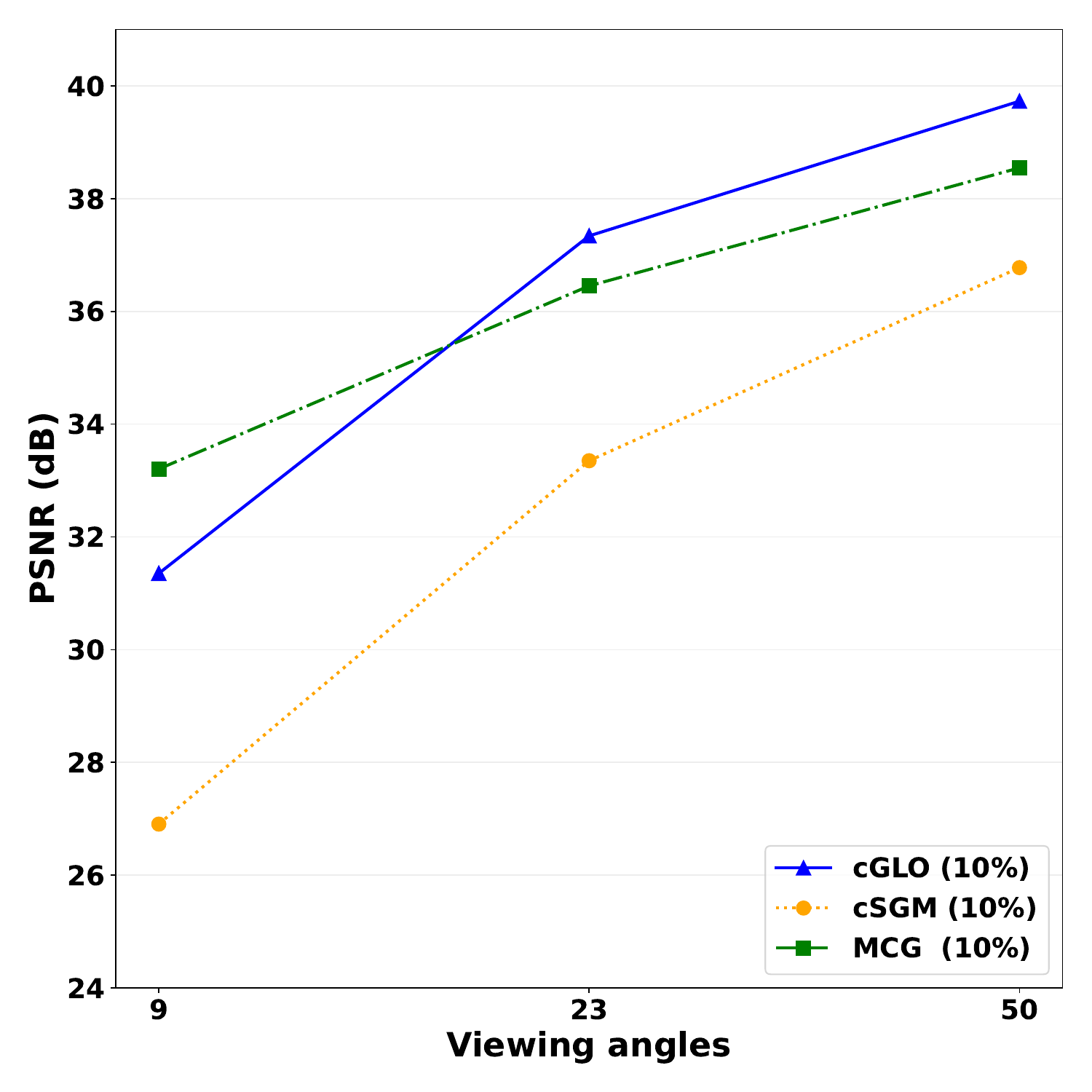}
              \includegraphics[width=0.29\textwidth]{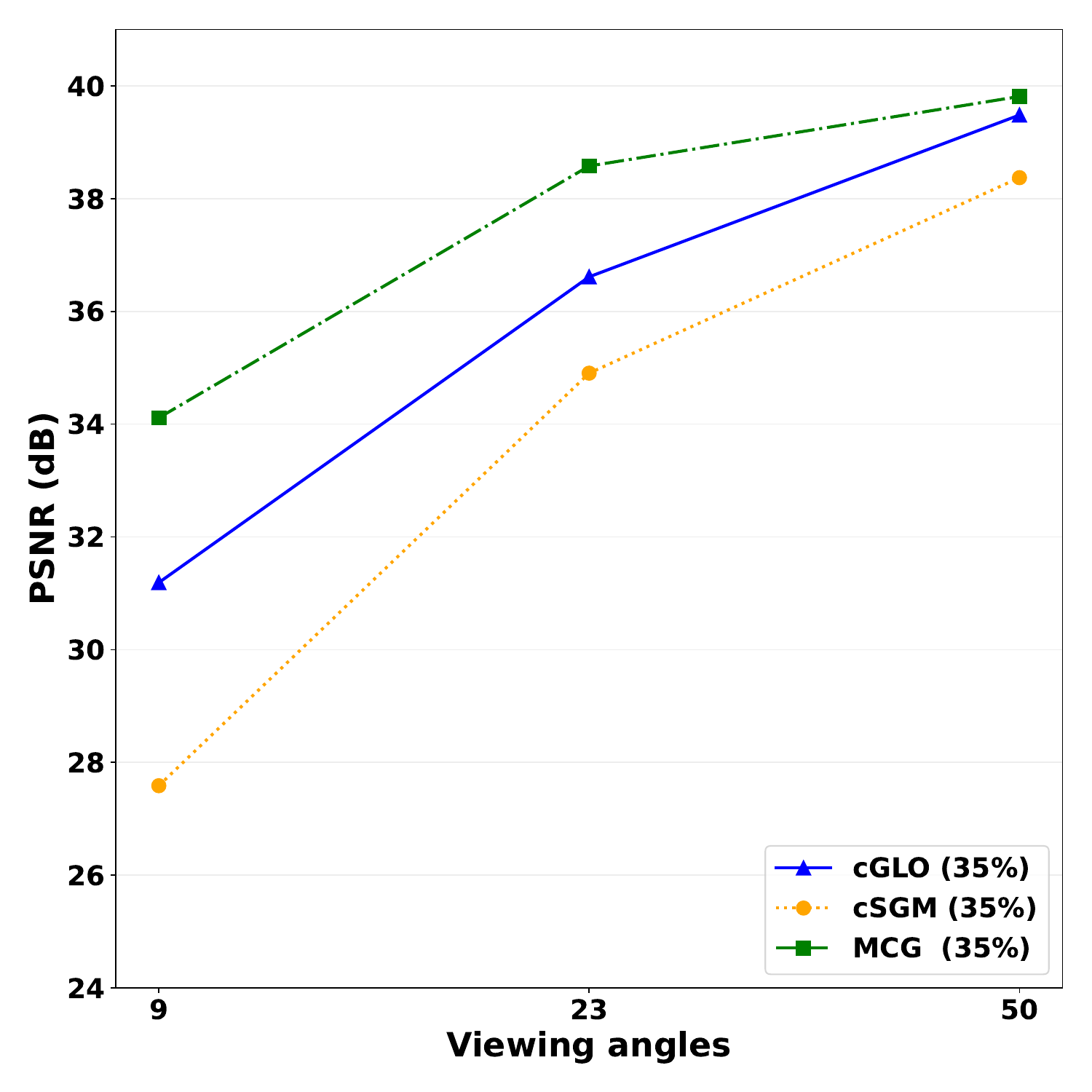}

              \includegraphics[width=0.29\textwidth]{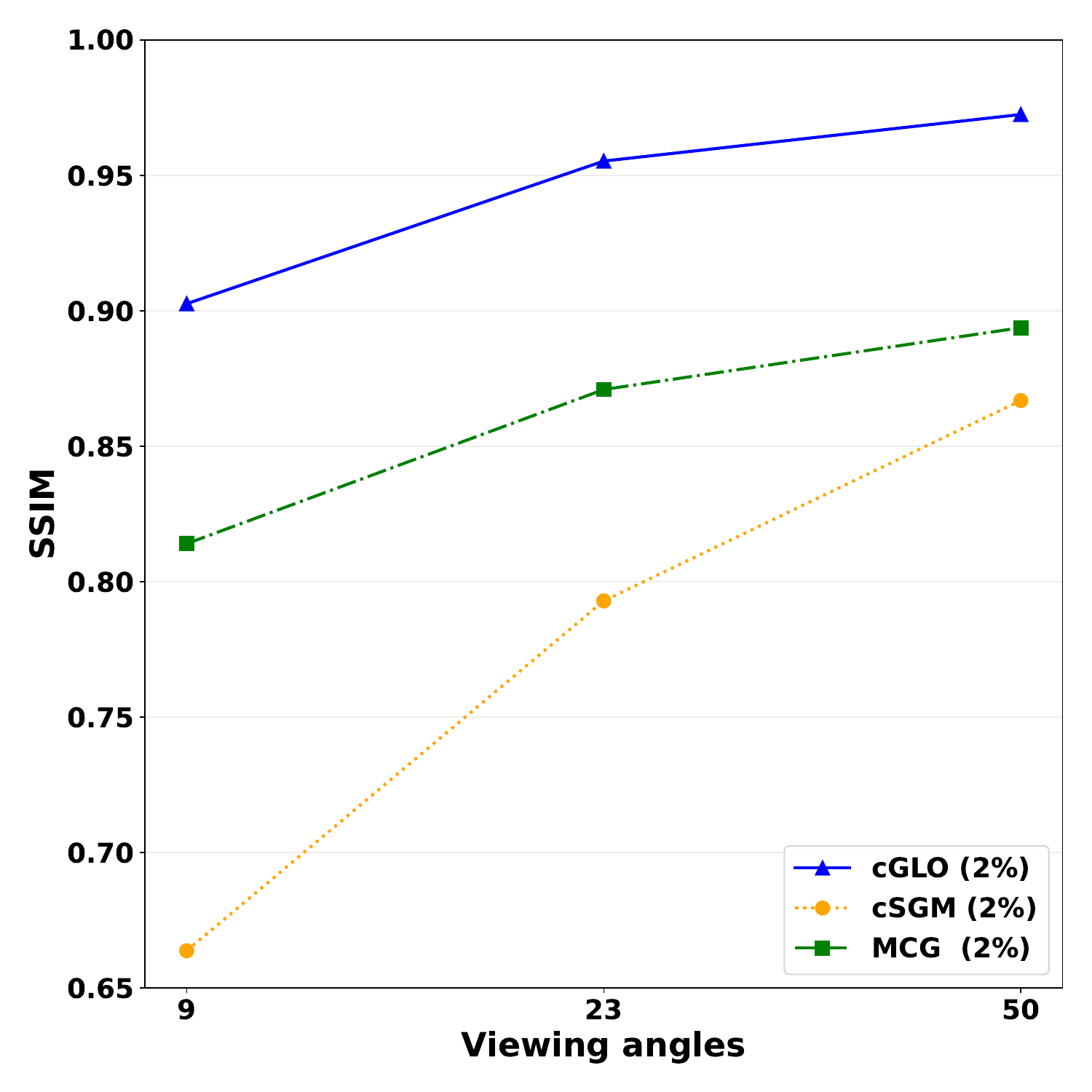}
              \includegraphics[width=0.29\textwidth]{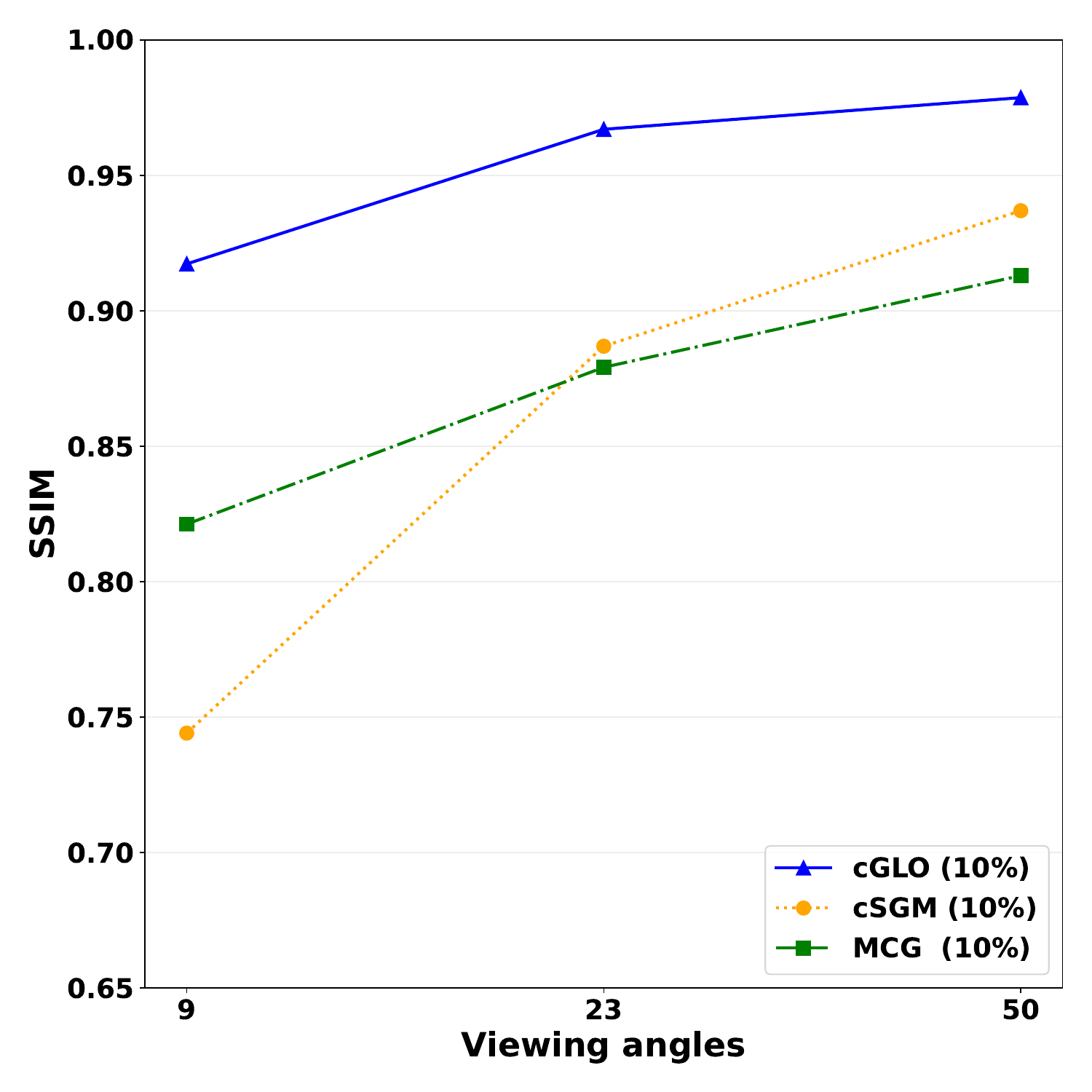}
              \includegraphics[width=0.29\textwidth]{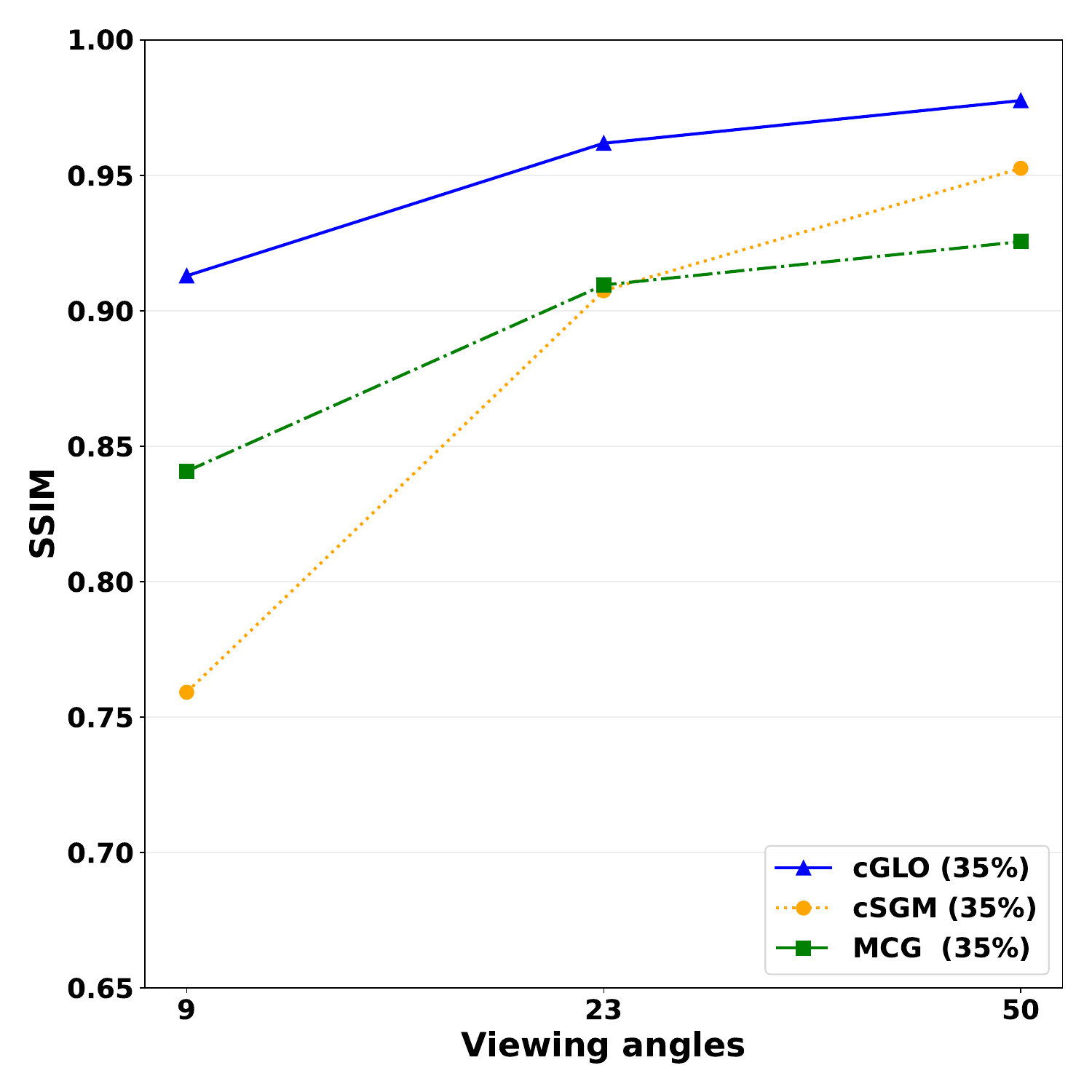}

              \caption{PSNR and SSIM - LDCT dataset}
              \label{fig:curves_ldct}

          \end{subfigure}

          \caption{
              PSNR and SSIM median values curves corresponding to reconstructions of slices
              from the LIDC (a) and the LDCT (b) test sets given 9, 23 and 50 experimental
              viewing angles. Each column is associated with one of the 2\%, 10\% and 35\%
              training sub-datasets presented in table \ref{tab:data}.
          }
          \label{fig:curves}

      \end{figure}

      \begin{figure}[h!]
      \centering
      \begin{subfigure}[b]{0.245\textwidth}
          \includegraphics[width=\textwidth]{images/GTX.pdf}
      \end{subfigure}
      \hfill
      \begin{subfigure}[b]{0.245\textwidth}
          \includegraphics[width=\textwidth]{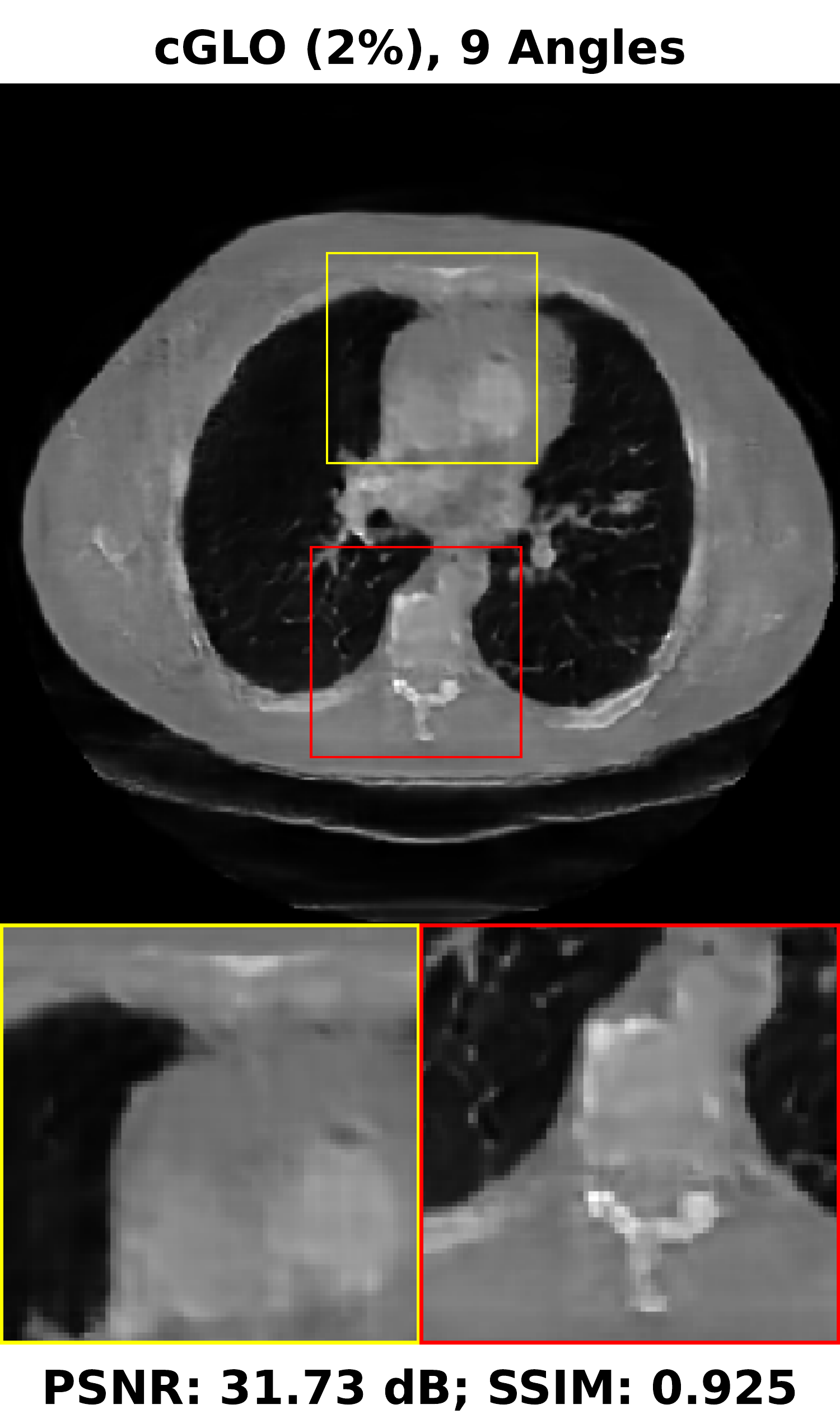}
      \end{subfigure}
      \hfill
      \begin{subfigure}[b]{0.245\textwidth}
          \includegraphics[width=\textwidth]{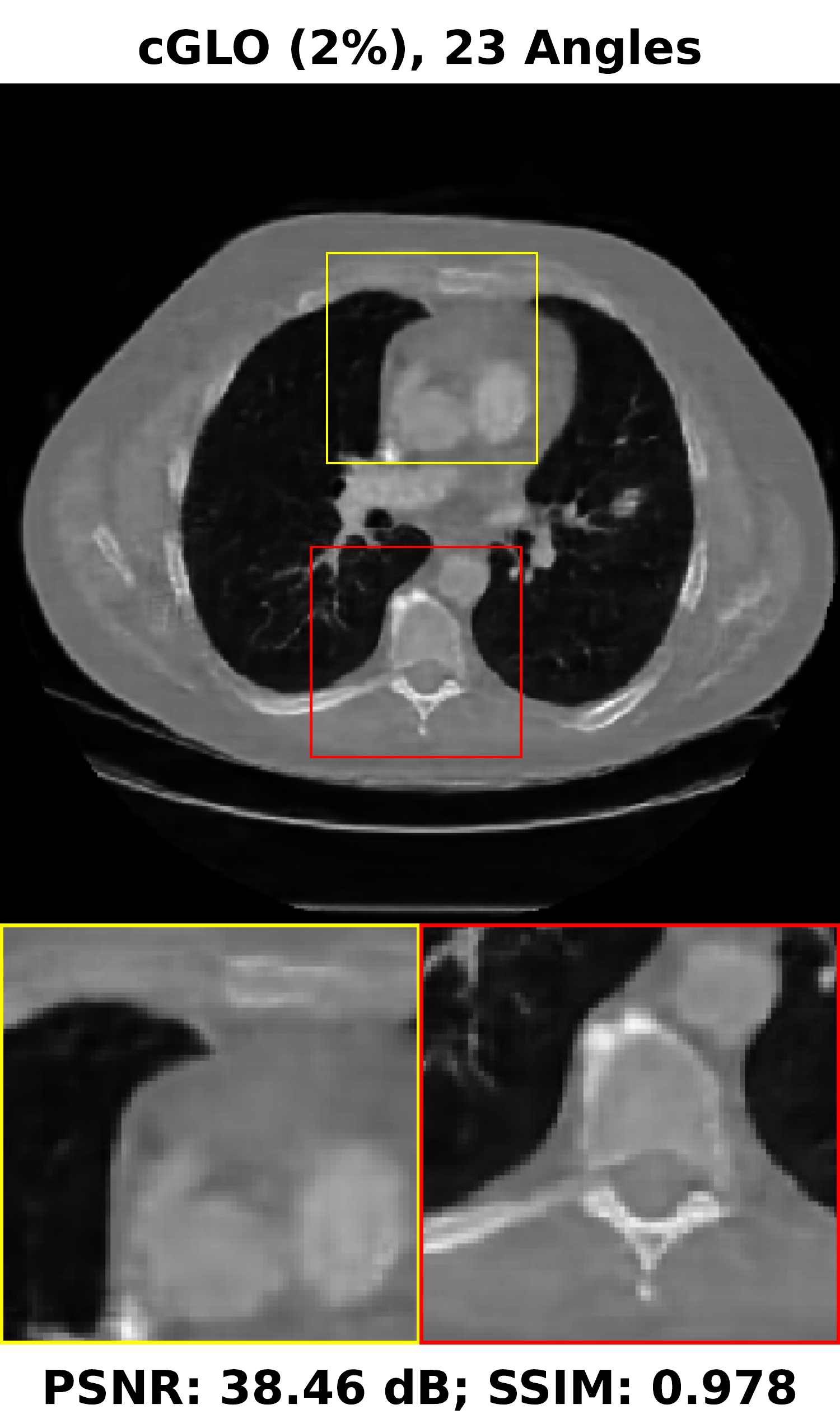}
      \end{subfigure}
      \hfill
      \begin{subfigure}[b]{0.245\textwidth}
          \includegraphics[width=\textwidth]{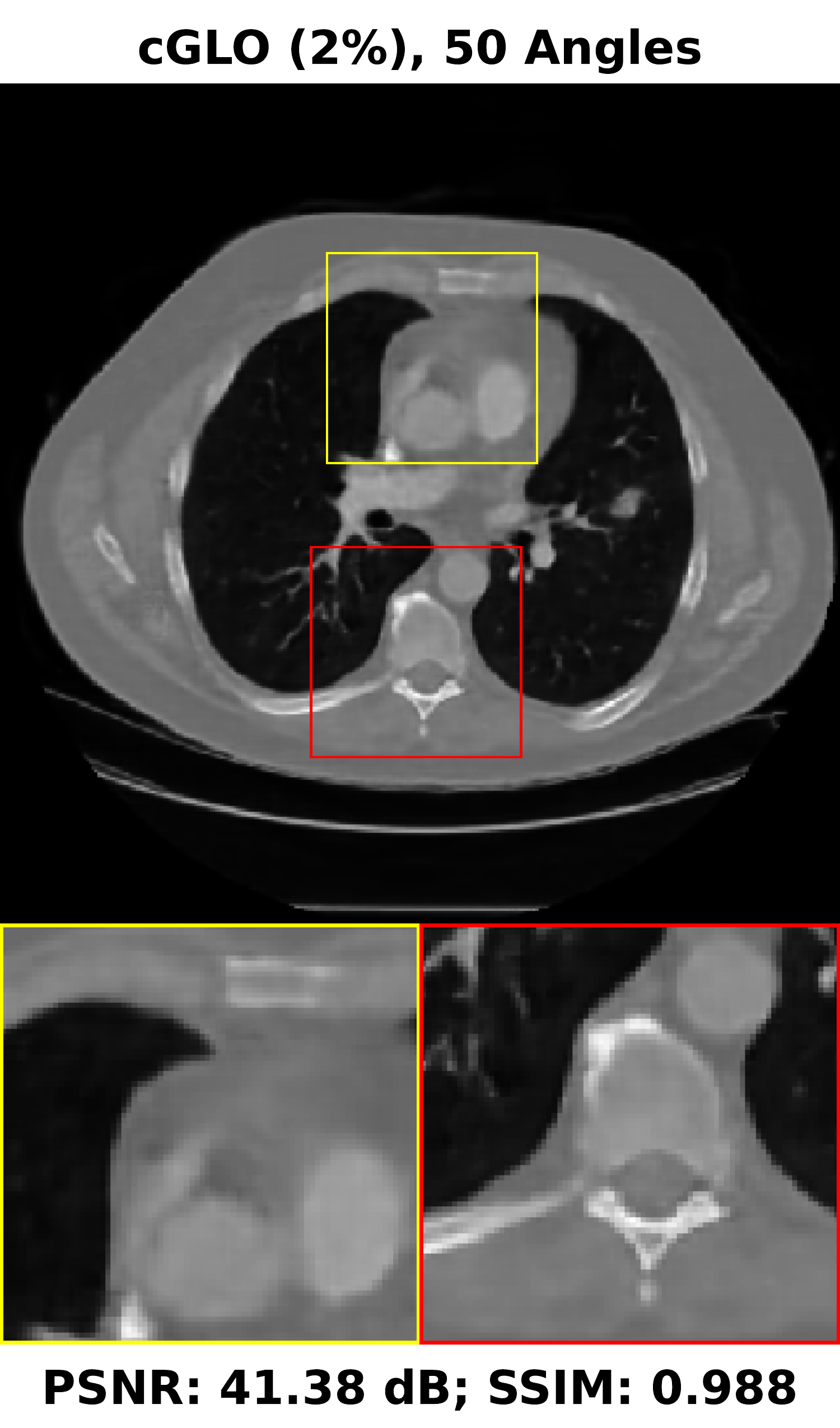}
      \end{subfigure}
      \hfill
      \begin{subfigure}[b]{0.245\textwidth}
          \includegraphics[width=\textwidth]{images/GTX.pdf}
      \end{subfigure}
      \hfill
      \begin{subfigure}[b]{0.245\textwidth}
          \includegraphics[width=\textwidth]{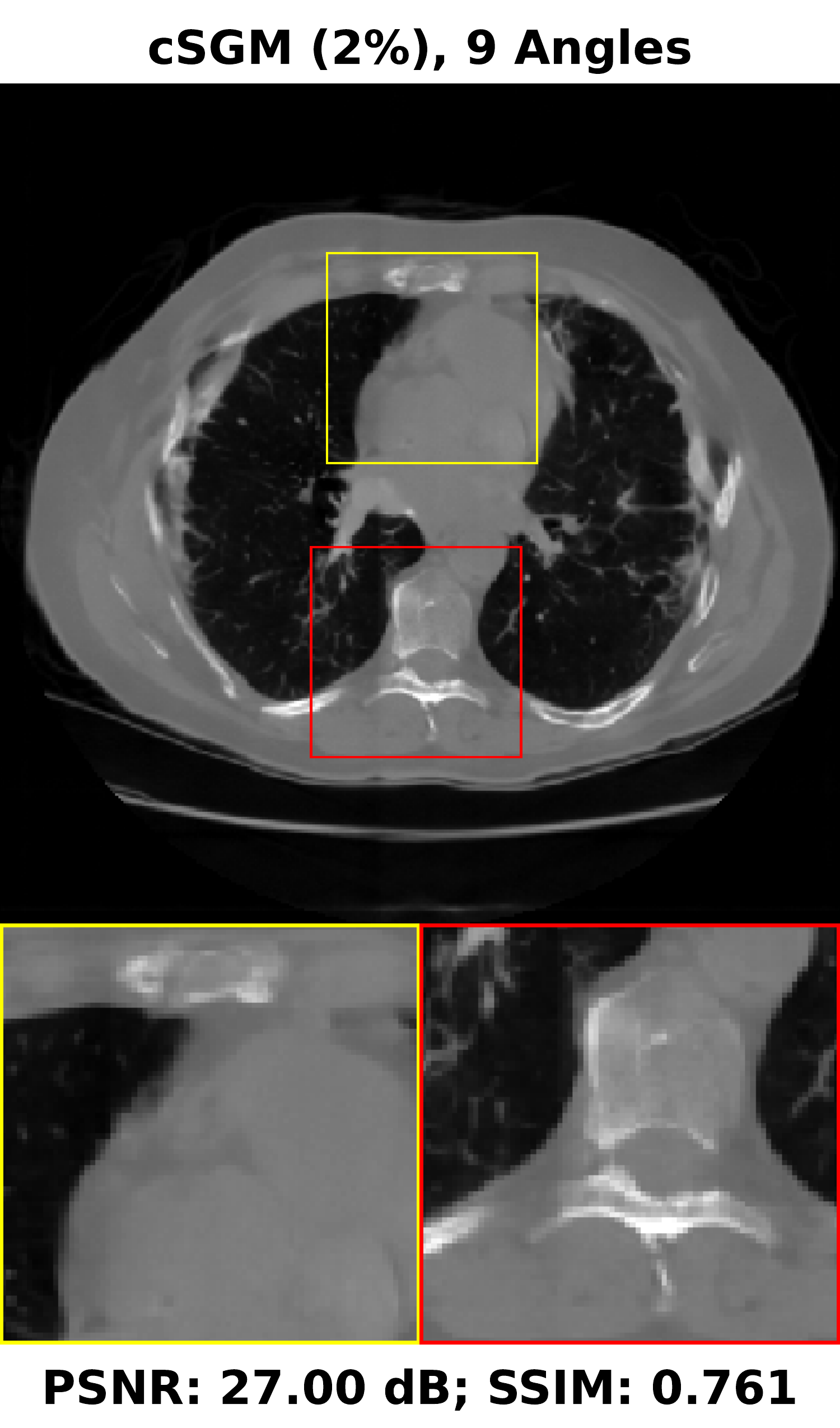}
      \end{subfigure}
      \hfill
      \begin{subfigure}[b]{0.245\textwidth}
          \includegraphics[width=\textwidth]{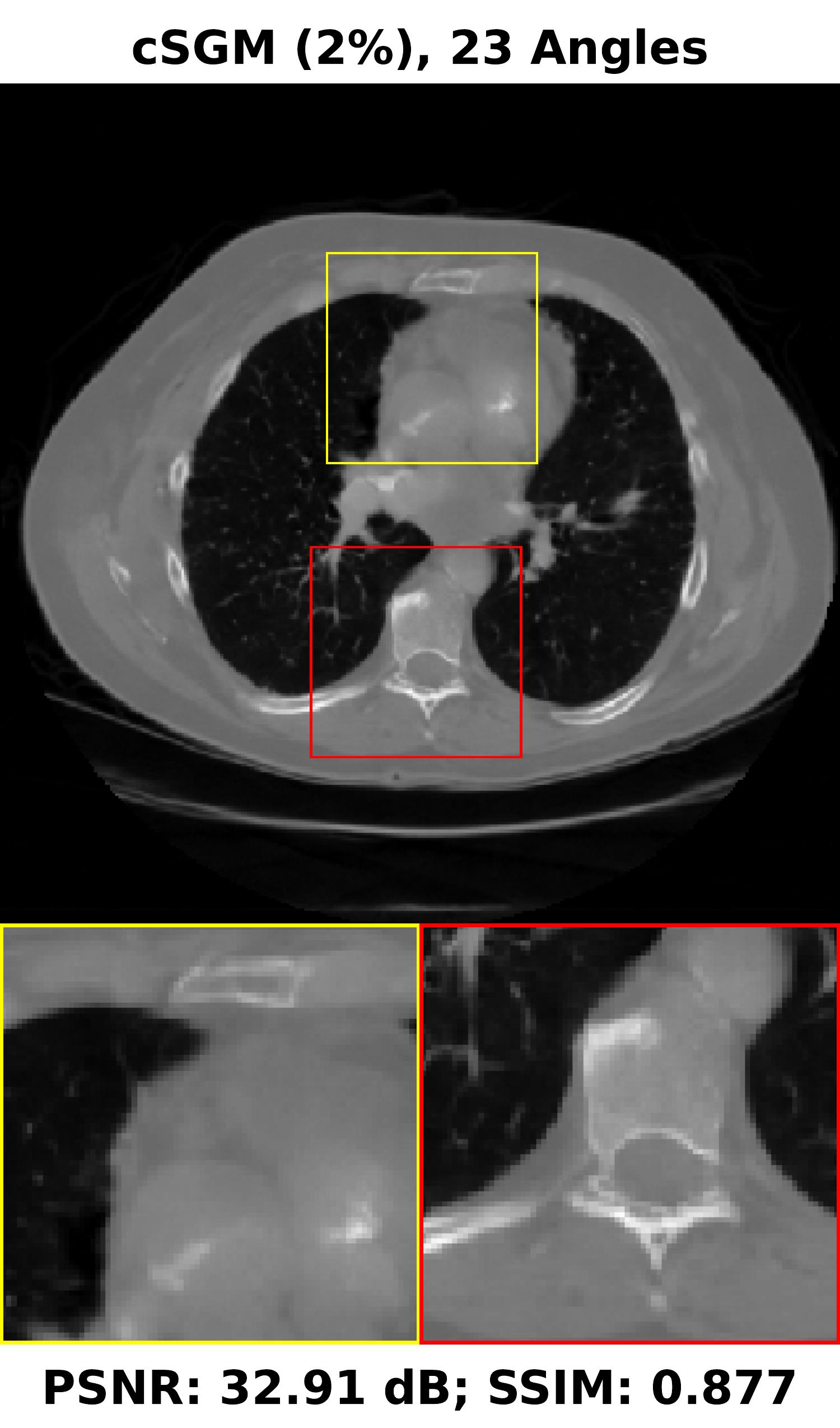}
      \end{subfigure}
      \hfill
      \begin{subfigure}[b]{0.245\textwidth}
          \includegraphics[width=\textwidth]{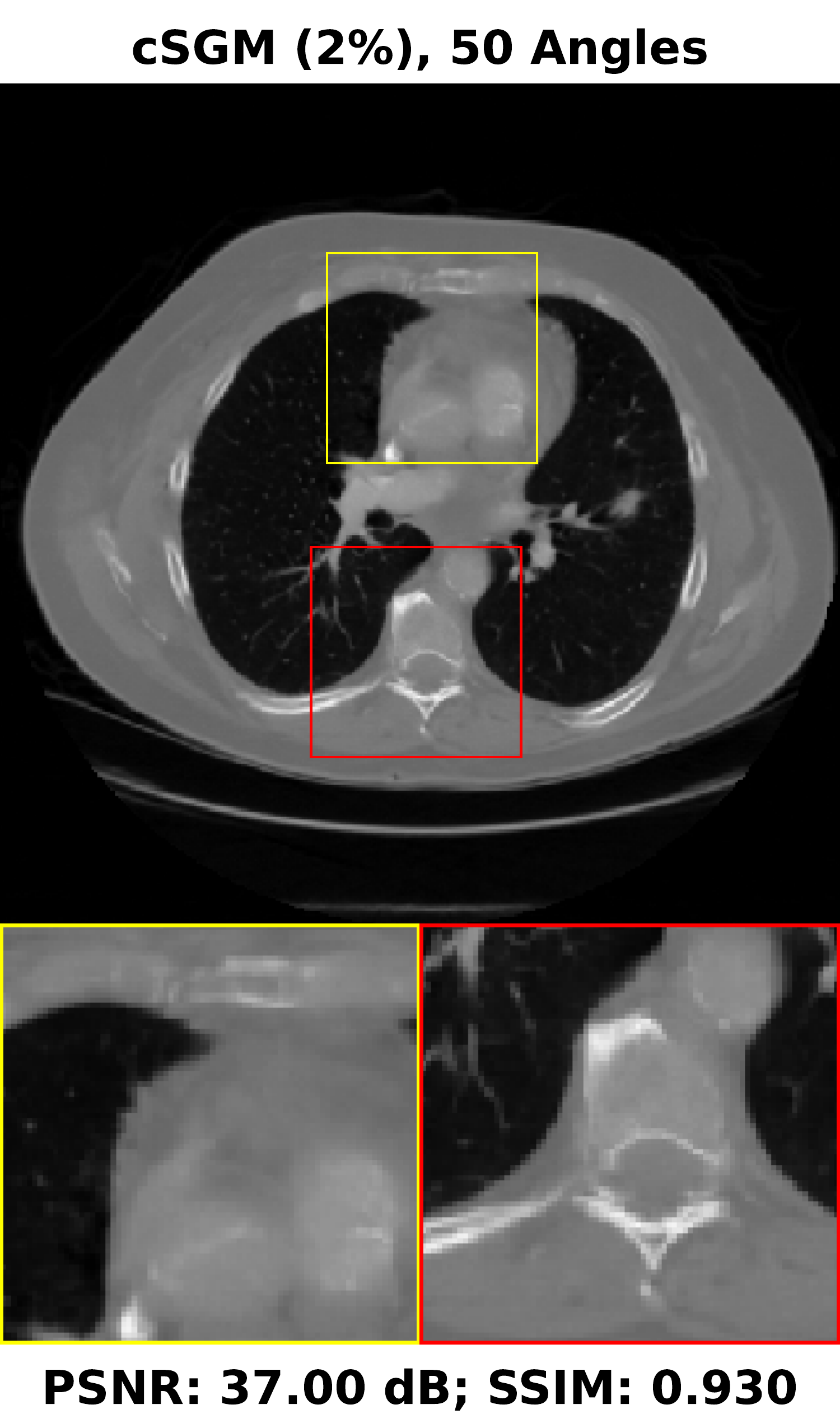}
      \end{subfigure}
      \hfill
      \begin{subfigure}[b]{0.245\textwidth}
          \includegraphics[width=\textwidth]{images/GTX.pdf}
      \end{subfigure}
      \hfill
      \begin{subfigure}[b]{0.245\textwidth}
          \includegraphics[width=\textwidth]{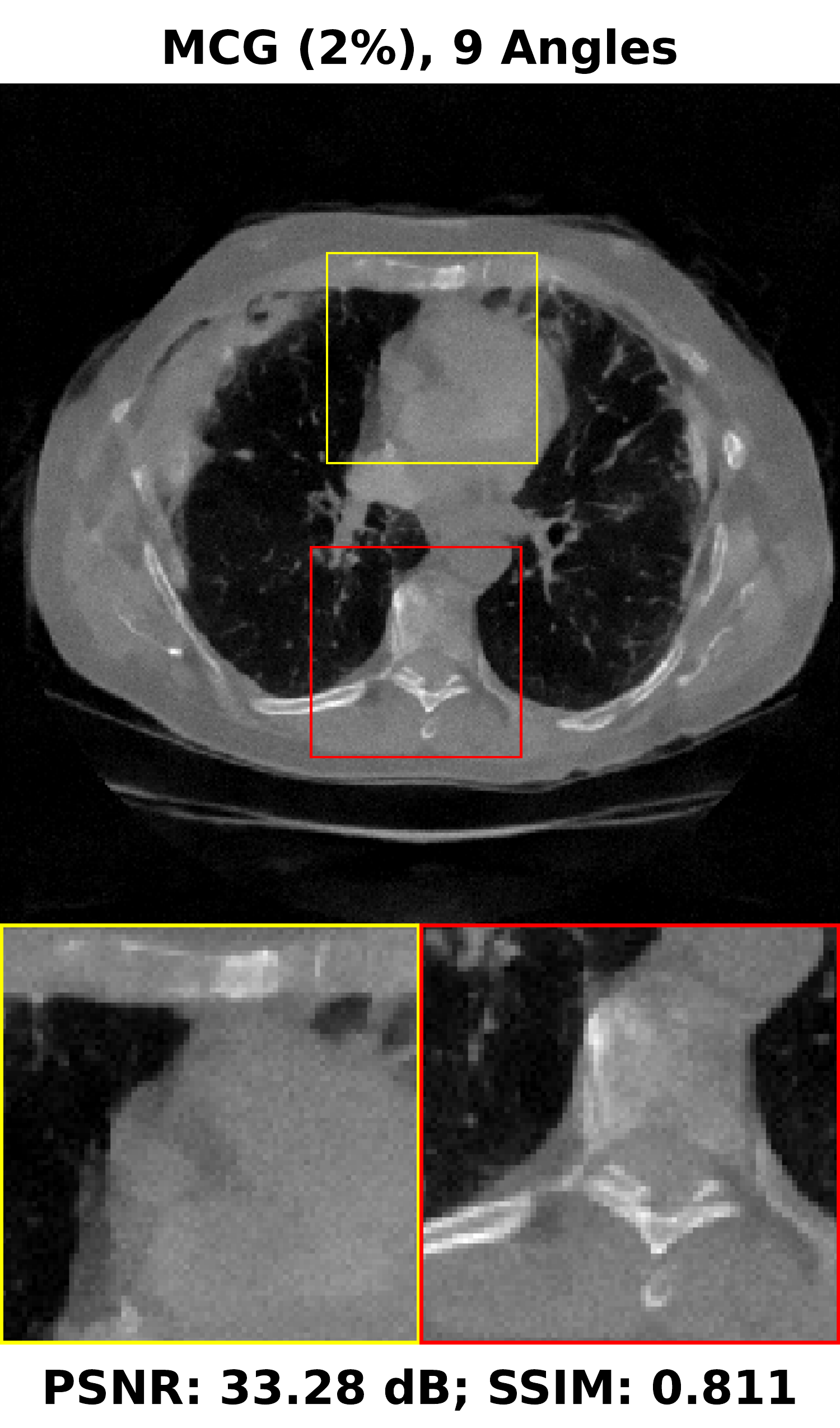}
      \end{subfigure}
      \hfill
      \begin{subfigure}[b]{0.245\textwidth}
          \includegraphics[width=\textwidth]{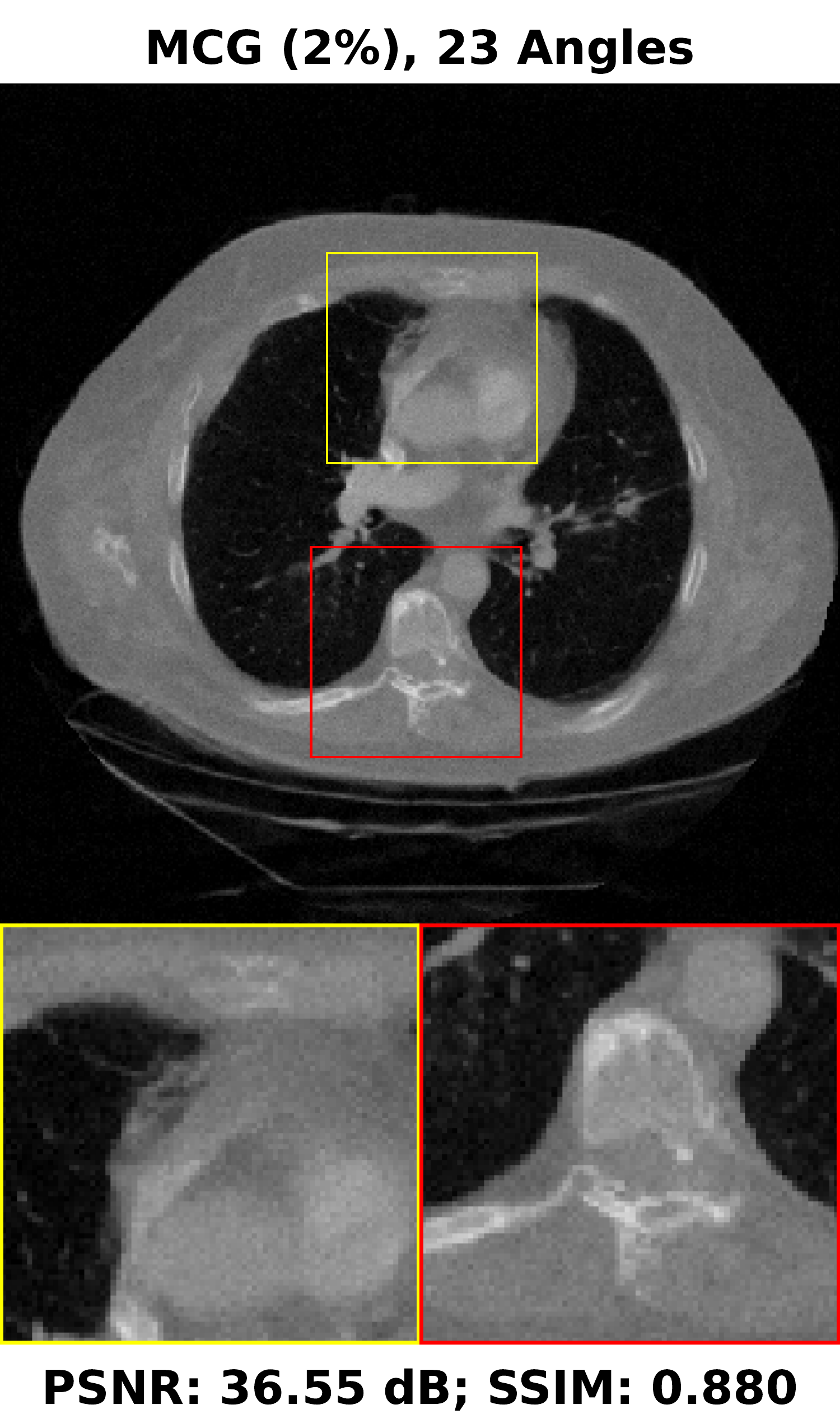}
      \end{subfigure}
      \hfill
      \begin{subfigure}[b]{0.245\textwidth}
          \includegraphics[width=\textwidth]{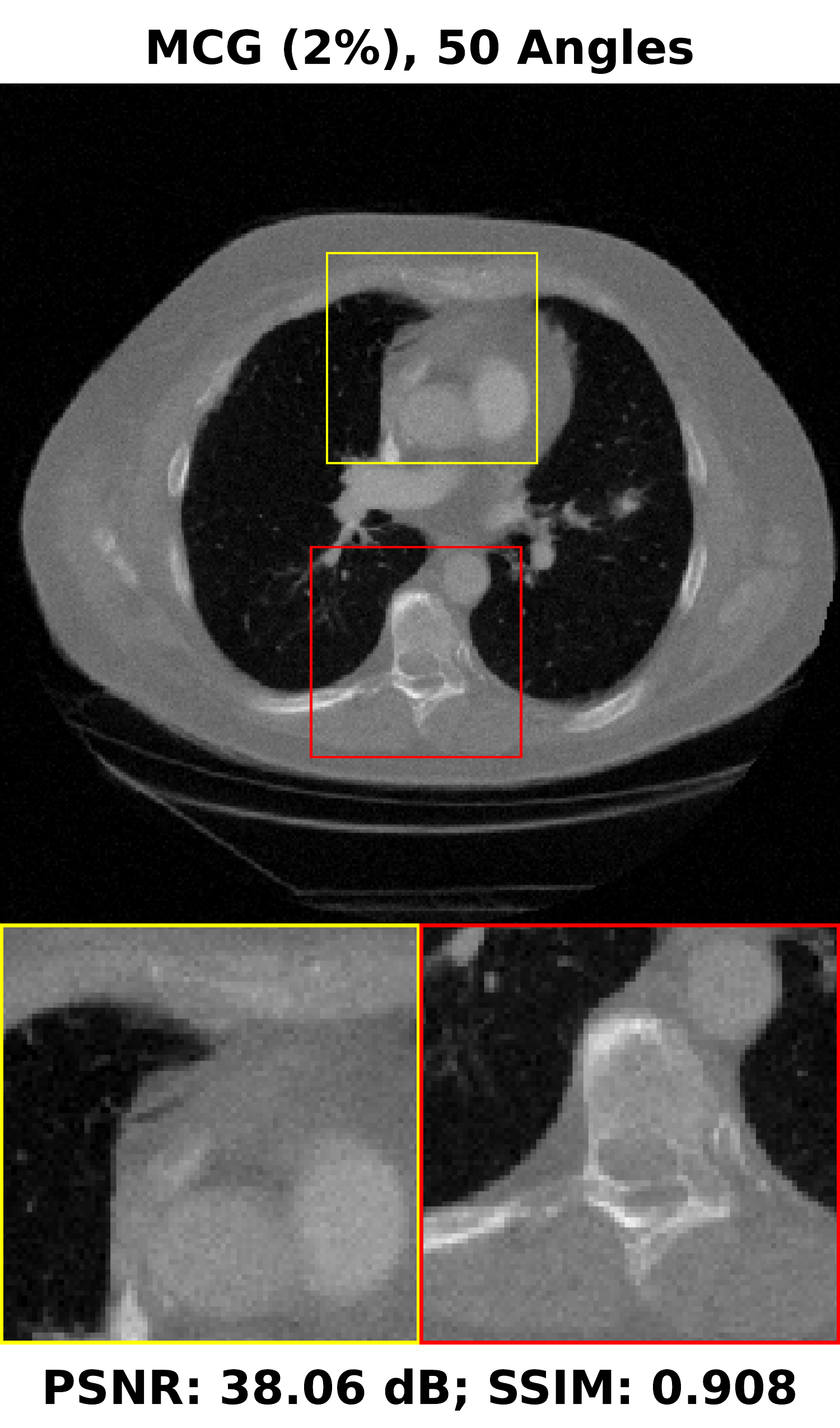}
      \end{subfigure}
      \caption{
          Examples of reconstructions given 9, 23 and 50 experimental viewing angles,
          obtained with cGLO (upper row), cSGM (middle row) and MCG (lower row).
          Methods are trained on the sub-dataset consisting of a 2\% portion of the LIDC
          dataset.
      }
      \label{fig:lidc_ex_2}
      \end{figure}

      As presented in figure \ref{fig:curves}, cGLO outperforms cSGM for all data and viewing
      angles configurations regarding both median PSNR and SSIM values. The largest performance
      gaps are observed for experiments given 9 viewing angles, with PSNRs ranging between
      $+3.13$ dB and $+6.06$ dB respectively for the 2\% and 10\% LIDC training sub-dataset.
      In addition, median PSNR and especially median SSIM curves in figure \ref{fig:curves}
      show an increasing advantage of cGLO over cSGM as the set of experimental viewing angles
      gets sparser and/or the training dataset gets smaller. Specifically for LIDC experiments,
      the lead in favor of cGLO over cSGM regarding SSIM starts at $+0.034$ (2\% sub-dataset and
      9 viewing angles) to reach $+0.139$ (35\% sub-dataset and 50 viewing angles).
      Furthermore, cGLO achieves higher median SSIM metrics values in all experimental setups,
      even in abundant data configurations with large training sub-datasets and many
      experimental viewing angles when compared to both cSGM and MCG. When compared to MCG on
      the same configurations as above, cGLO demonstrates a SSIM gain ranging from $+0.041$ to
      $+0.134$.  Figure \ref{fig:curves} also displays a dominance of MCG over cGLO regarding
      PSNR metrics on use cases with abundant data, e.g.\ with prior training on the 35\% LDCT
      sub-dataset with a superiority of $+0.33$ dB, $+1.97$ dB and $+2.92$ dB for PSNRs given
      respectively 50, 23 and 9 viewing angles.  On the smaller 2\% and 10\% training
      sub-datasets, cGLO equals or outperforms MCG regarding PSNR given experimental setups
      with more than 23 viewing angles. For instance on the LIDC experiments, a gain rising
      from $+0.9$ dB (10\% sub-dataset and 23 viewing angles) to $+2.58$ dB (2\% sub-dataset and
      50 viewing angles) can be observed.

      The effects of training dataset sizes and viewing angles are consistent between
      experiments but are especially visible on the curves describing SSIM metrics for
      reconstructions from the LDCT test set in last row of panels of figure \ref{fig:curves}.
      Considering one method (one color), higher vertical gaps between panels of the same row
      show greater dependency on training dataset sizes, while steeper slopes within each panel
      indicate a superior impact of viewing angles on performance. In this regard, cSGM is the
      most affected by a diminishing number of viewing angles and/or smaller training datasets.
      For example, when considering configurations with abundant data, e.g.\ the LIDC 35\%
      sub-dataset, cSGM PSNR values plummet from $40.04$ dB to $30.28$ dB implying a $-9.76$ dB
      decrease between 50 and 9 viewing angles. As intended by its design, MCG demonstrates in
      the same context the most robustness (specifically regarding PSNRs) against sparser
      experimental setups with a $-5.34$ dB loss. cGLO sustains performances close to MCG with a
      $-5.78$ dB drop. Finally, our method is the most parsimonious/robust concerning training
      dataset sizes. Indeed, in the configuration with abundant viewing angles (50), cGLO
      PSNRs for LIDC experiments fall from $40.89$ dB to $40.50$ dB, i.e.\ a small loss
      of $-0.39$ dB when compared to MCG ($-2.87$ dB) and cSGM ($-3.14$ dB).

      The prominent points of the quantitative performances comparison between cGLO, cSGM and
      MCG are the following. For every configurations, cGLO offers better reconstructions than
      cSGM and MCG regarding the SSIM metric and also surpasses cSGM when considering PSNR
      values. Further cGLO is less dependent on training dataset sizes than MCG which is more
      robust to reduced viewing angles. Consequently, when comparing PSNR values, cGLO provides
      better performances than MCG in scenarios with more than 23 angles and restricted access
      to training data, e.g.\ less than 10\% of the LIDC dataset.

    \subsubsection*{Effect of large training datasets on cGLO}

      As illustrated in figure \ref{fig:curves}, contrary to cSGM and MCG, as the training
      datasets becomes larger, cGLO PSNR performances stagnate and even slightly worsen. More
      specifically for PSNRs, averaged over angle configurations, a drop is observed when cGLO
      is trained on the 35\% rather than the 10\% sub-dataset, from $38.86$ dB to $38.62$ dB
      for LIDC, and from $36.14$ dB to $35.74$ dB for LDCT. Given a fixed representation
      capacity for our decoder architecture, i.e.\ a fixed number of input channels
      $\mathbf{C}$, the model becomes under-parameterized after a given training dataset size
      is reached. Since the model parameters and the latent codes are jointly optimized in Eq.\
      \eqref{cglo_init} with respect to a data-likelihood objective function, the model error
      distribution is widely spread over the entire training dataset \cite{theis_note_2016}.
      Consequently, when the data variability overcomes the model capacity, each additional
      training example further deteriorates the average prediction quality. Ideally, the model
      capacity should be tuned to best fit each experimental setup.  However, the experiments
      realized in this paper demonstrate that even with a fixed decoder architecture, cGLO
      produces superior reconstructions for multiple experimental setups.

    \subsubsection*{Reconstruction structural quality}

      Figure \ref{fig:lidc_ex_2} compares reconstructions computed with cGLO, cSGM and MCG from
      the LIDC test dataset. Reconstruction quality clearly degrades as the number of
      experimental viewing angles diminishes. The type of degradation, however, differs between
      cGLO and the score-based methods cSGM and MCG. While cGLO reconstructions lack sharp
      details and get noisier for very sparse CT or when trained on very small datasets, cSGM
      and MCG tend to alter the structural integrity of the slices.  This observation
      corroborates and explains the total dominance of cGLO over all the methods when comparing
      median SSIM curves rather than median PSNR curves in figure \ref{fig:curves}.

      The fact that cSGM/MCG and cGLO yield different reconstruction structures for more
      difficult experimental setups can be explained by the way each model employs experimental
      data. cSGM uses a proximal optimization step to mix experimental data with unconditioned
      samples. Details are provided in Appendix B (Eq.\ \eqref{proximal_step}). This leads to a
      cost function expressing a balance between data likelihood and prior, which only enforces
      a loose constraint on the conformity of the proposed reconstruction to experimental data.
      When the number of viewing angles diminishes, there are increasing structural
      deformations as shown in figure \ref{fig:lidc_ex_2}. MCG correction step (detailed in
      Appendix B Eq.\ \eqref{manifold_iteration}) prevents the inference mechanism to deviate
      from data consistency for sparse experimental setups. However, when trained on
      insufficient data quantities, the assumption that the score-model approximate the data
      manifold may not hold anymore. On the contrary, the cost function of cGLO, described in
      equation \eqref{cglo_objective}, enforces a strict constraint on experimental data. With
      decreasing viewing angles, the energy landscape of this cost function becomes less
      smooth, leading to the presence of local minima and harder convergence of the
      optimization process. As seen in figure \ref{fig:lidc_ex_2}, this leads to degraded
      reconstructions with more noise and less details at 9 viewing angles for example.
      However, contrary to cSGM and MCG, cGLO does not tend to create hallucinated structures.

\section{Conclusion}

    In this work we presented cGLO, a novel unsupervised method to solve IIPs built upon the
    general framework of GLO \cite{bojanowski_optimizing_2018}. Contrary to supervised
    strategies, our method does not require a fixed experimental setup. Analogously to DIP
    \cite{baguer_computed_2020}, our strategy benefits from the structural prior induced by
    convolutional decoders without any preceding training. Furthermore, cGLO takes advantage of
    a shared objective function by reconstructing multiple slices simultaneously at examination
    time and may be initialized beforehand on an unsupervised training dataset of any size to
    improve reconstruction quality. In other words, our approach is a very flexible
    plug-and-play reconstruction technique that can be used \textit{with or without}
    unsupervised training data and \textit{without} any assumption on the IIP at hand.

    cGLO was tested on full-dose sparse view CT which is a widely studied reconstruction
    problem arising in medical imaging and compared to the current state-of-the-art approaches,
    MCG \cite{chung_manifold_2022} for unsupervised cases and DIP for cases with no prior
    training. Experiments conducted in this paper cover a wide range of realistic setups with
    varying amount of available training data and experimental viewing angles. Quantitative
    results regarding pixel-wise metric (PSNR) demonstrate that cGLO is a parsimonious and
    robust reconstruction method as it offers similar to better performances when compared to
    MCG on small data regimes with sufficient viewing angles and surpasses DIP in every
    experimental setups. Moreover, for the entire set of conducted experiments, cGLO offers the
    best performances with respect to the structural metric (SSIM). Indeed, reconstruction
    examples illustrate that cGLO exhibits a propensity to preserve the structural integrity of
    reconstructions and it avoids the inclusion of hallucinated pattern, even for very
    ill-posed IIPs, thanks to its straightforward conditioning to experimental measurements.

    While the performance of cGLO has been tested on a tomographic reconstruction problem, the
    framework developed in this article can be readily applied to other ill-posed IIPs, such as
    MRI reconstruction from a sparsely sampled k-space. Furthermore, since the cGLO approach
    only requires the knowledge of the forward operator, it could be extended to solve
    non-linear IIPs. One of the main interests of cGLO is that it does not require any backward
    operator, such as FBP for tomographic reconstruction. Consequently, it is a method of
    choice for IIPs where such operators do not exist, such as multi-material CT
    reconstruction. Future work will explore the application of cGLO to such IIPs. Since cGLO
    efficiently learns correlations between the output channels of its decoder network, further
    developments will also focus on multi-task applications, such as jointly learning
    reconstruction and segmentation.

\clearpage

\bibliographystyle{unsrt}

\clearpage

\section*{Appendix A: Computed Tomography}

    \begin{wrapfigure}{r}{0.32\textwidth}
        \vspace{-20pt}
        \begin{center}
            \includegraphics[width=0.31\textwidth]{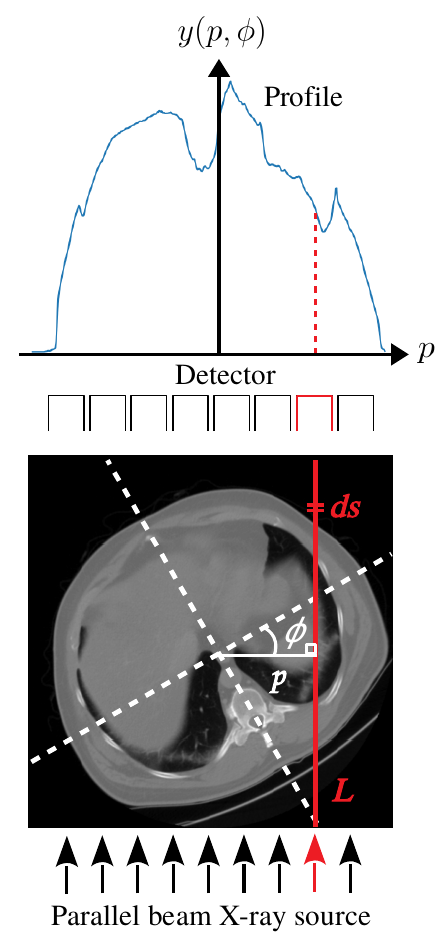}
        \end{center}
        \vspace{-10pt}
        \caption{
            Parallel beam geometry profile acquisition for one viewing angle
        }
        \vspace{-40pt}
        \label{fig:radon}
    \end{wrapfigure}

    Given a narrow beam of X-ray photons of energy $E \in [0, \varepsilon]$, with a normalized
    dose spectrum $S(E)$, tracing through a material $m$ with homogeneous density distribution
    $\rho_m$, the beam intensity decreases in accord with the Beer-Lambert law:

    \begin{equation}
    \label{spectrum_beam_attenuation}
        \frac{I}{I_0} = \int_0^\varepsilon S(E) \exp \Biggl[
            - \biggl(\frac{\mu_m}{\rho_m}\biggr)_E \, \rho_m L
        \Biggr]dE
    \end{equation}

    where $I_0$ and $I$ are respectively the input and observed intensities, $L$ is the beam
    path length and $\mu_m$ the linear attenuation coefficient of material $m$. The ratio
    $\Bigl(\frac{\mu_m}{\rho_m}\Bigr)_E$ corresponds to the mass attenuation of material $m$.
    Depending on the energy level $E$, this macroscopic attenuation is obtained by adding up
    the effect of several absorption and scattering mechanisms, e.g.\ Rayleigh and Compton
    scattering, photoelectric absorption or pair production.

    In medical tomodensitometry, the quantity of interest is the heterogeneous density
    distribution reflecting the patient internals.  However, further assumptions are made to
    recover this distribution from experimental measurements: bones and other tissues are
    reduced to one equivalent material with water mass attenuation and the X-ray beam is
    considered monoenergetic.

    Now, by considering an horizontal plane section through the patient body, i.e.\ a slice
    $x$, and by moving the source and detector as indicated in figure \ref{fig:radon}, it is
    possible to obtain a profile $y$, on the detector axis $p$, for the viewing angle $\phi$:

    \begin{equation}
    \label{viewing_angle_profile}
        y(p, \phi) = -\log\biggl(\frac{I}{I_0}\biggr)
                   = \int_L x\, ds
    \end{equation}

    For continuous $p$ and $\phi$ the profiles $y$ may be identified with the two-dimensional
    Radon transform \cite{deans_radon_1983}, also called the forward operator
    $\mathcal{T}$, of the slice $x$, such that $y = \mathcal{T}(x)$. However, in practical use
    cases, projection profiles are acquired for a set of incremental values $\Phi = \phi_1,
    ..., \phi_N$, which constitutes a sampling of the Radon transform.  Moreover, shot noise
    following a Poisson process $\tau$ usually appears in experimental profiles for lower
    doses. Resulting experimental measurements can then be described by:

    \begin{equation}
    \label{experimental_profiles}
        y_\Phi = \mathcal{T}_\Phi(x) + \tau
    \end{equation}

    Given a number of viewing angles close to the Nyquist sampling criterion
    \cite{natterer_mathematical_2001}, and sufficiently low noise, the inverse of the
    sampled Radon transform $\mathcal{T}_\Phi^\dagger$ can be computed with standard methods,
    e.g.\ FBP, to retrieve the slice density map without producing artifacts. By stacking
    several slices, the two-dimensional information may be converted to three-dimensional
    information.

    In the case of insufficient viewing angles, e.g.\ sparse CT, or in presence of too much
    noise, e.g.\ low-dose CT, the inverse problem becomes ill-posed. It is classically solved
    using iterative optimization algorithm \cite{beck_fast_2009} using a \textit{Maximum A
    Posteriori} (MAP) formulation of the problem:

    \begin{equation}
    \label{maximum_a_posteriori}
        x^* = \argmin_x \lVert \mathcal{T}_\Phi(x)
            - y_\Phi \rVert^2_2 + \mathcal{R}(x)
    \end{equation}

    where $\mathcal{R}$ is a real valued function used to regularize the optimization process
    by injecting prior information on the desired output $x^*$. Most common operators
    encountered in the literature to model this information are $\lVert x \rVert_1$, $\lVert x
    \rVert_2$ and Total Variation (TV), $\mathrm{TV}(x) = \lVert \nabla x \rVert_1$.

\clearpage

\section*{Appendix B: Score-Based Generative Models (SGMs)}

  \subsection*{Unconditioned sampling with SGMs}

    The method developed by Song et al.\ \cite{song_score-based_2021} is generative. With a
    dataset of i.i.d.\ samples $\mathbf{x} \in X$ from an unknown distribution
    $p(\mathbf{x})$, it aims at generating new data samples. To this end, it
    defines a continuous diffusion process, that progressively perturbs samples from
    $p(\mathbf{x})$ into samples from a tractable prior distribution. It is formulated as a
    linear SDE as follows:

    \begin{equation}
    \label{sde}
        \mathrm{d}\mathbf{x}_t = f(t) \mathbf{x}_t \mathrm{d}t
                               + g(t) \mathrm{d}\mathbf{w}_t
    \end{equation}

    where $t\in[0,1]$, $\mathrm{d}t$ is an infinitesimal time step, $f$ and $g$ are real
    valued functions, $\mathbf{w}_t$ is a standard Wiener process, and $\mathbf{x}_t$
    represents the samples through the perturbation process. The marginal probability
    distribution of $\mathbf{x}_t$, $p_t(\mathbf{x}_t)$ can be further derived from Eq.\
    \eqref{sde}, as well as the transition distribution from $\mathbf{x}_0$ to
    $\mathbf{x}_t$, $p_{0t}(\mathbf{x}_t | \mathbf{x}_0)$. $f$ and $g$ are carefully
    designed so that, for any $p_0(\mathbf{x}) \equiv p(\mathbf{x})$, Eq.\ \eqref{sde}
    ensures convergence at $t=1$ towards a distribution $p_1(\mathbf{x})$ close to a
    Gaussian noise.

    As demonstrated by Anderson \cite{anderson_reverse-time_1982}, the reverse of the
    diffusion process, in Eq.\ \eqref{sde}, is also a diffusion process which can be
    written as the following reverse-time SDE:

    \begin{equation}
    \label{reverse_sde}
        \mathrm{d}\mathbf{x}_t = [
            f(t) \mathbf{x}_t
          + g(t)^2 \nabla_{\mathbf{x}_t}\mathrm{log}p_t(\mathbf{x}_t)
        ] \mathrm{d}t + g(t) \mathrm{d}\mathbf{\overline{w}}_t
    \end{equation}

    where $\mathbf{\overline{w}}_t$ is a standard Wiener process in the reverse-time
    direction, and $\mathrm{d}t$ indicates an infinitesimal negative time step. Solving
    Eq.\ \eqref{reverse_sde} from $t=1$ to $t=0$ corresponds to a continuous denoising,
    yielding data samples $\mathbf{x}_0 \sim p_0(\mathbf{x}) \equiv p(\mathbf{x})$ from
    noise samples \cite{song_score-based_2021}. However, it requires the score function
    $\nabla_{\mathbf{x}_t} \mathrm{log}p_t(\mathbf{x}_t)$ of $p_t(\mathbf{x}_t)$.  Using a
    time-dependent neural network $s_\theta(\mathbf{x},t)$ called the score model, it is
    learned on $X$ with denoising score matching. As described in
    \cite{vincent_connection_2011} and \cite{song_score-based_2021}, the score model is
    trained by approximating the known score functions of the transition distributions from
    $\mathbf{x}_0$ to $\mathbf{x}_t$, $p_{0t}(\mathbf{x}_t | \mathbf{x}_0)$. Optimized
    parameters $\theta^*$ ensure that $s_{\theta^*}(\mathbf{x},t) \approx
    \nabla_{\mathbf{x}_t} \mathrm{log}p_t(\mathbf{x}_t)$ according to denoising score
    matching theory. The score model can then be plugged into Eq.\ \eqref{reverse_sde},
    yielding:

    \begin{equation}
    \label{score_reverse_sde}
        \mathrm{d}\mathbf{x}_t = [
            f(t) \mathbf{x}_t + g(t)^2 s_{\theta^*}(\mathbf{x},t)
        ] \mathrm{d}t + g(t) \mathrm{d}\mathbf{\overline{w}}_t
    \end{equation}

    \sloppy from here, one can sample from $p_0(\mathbf{x})$ using a sequence of time steps
    ${0=t_0<t_1<...<t_N=1}$ with standard iterative sampling techniques:

    \begin{equation}
    \label{unconditioned_iteration}
        \mathbf{x}_{t_{i-1}} = \boldsymbol{h}(
            \mathbf{x}_{t_i},
            \mathbf{z}_i, s_{\theta^*}(\mathbf{x}_{t_i},t_i)
        )
    \end{equation}

    where $\mathbf{z}_i \sim \mathcal{N}(\mathbf{0}, \boldsymbol{I})$ and $\boldsymbol{h}$
    denotes the iterative function related to the chosen sampling algorithm, such as
    annealed Langevin dynamics \cite{song_generative_2019} or Predictor-Corrector samplers
    \cite{song_score-based_2021}.

  \subsection*{Conditioned sampling with SGMs (cSGM)}

    Although it is natural to \textit{unconditionally} sample from Eq.\
    \eqref{score_reverse_sde}, conditioning the sampling process to measurements, i.e.\
    being able to sample from the posterior distribution $p(\mathbf{x}|\mathbf{y})$, is not
    trivial. Building on the unconditional case, Song et al.\ \cite{song_solving_2022}
    introduce perturbed measurements $\mathbf{y}_t$ given the experimental observation
    $\mathbf{y}$. They design an iterative sampling algorithm that promotes, at each
    time-step $t_i$, consistency of \textit{conditioned} samples $\mathbf{x}'_{t_i}$
    simultaneously with the perturbed measurements $\mathbf{y}_{t_i}$ and the
    \textit{unconditioned} samples $\mathbf{x}_{t_i}$ by solving a proximal optimization
    step:

    \begin{equation}
    \label{proximal_step}
        \mathbf{x}'_{t_i} = \argmin_{w}\lbrace
        (1-\lambda) \lVert w - \mathbf{x}_{t_i} \rVert^2_{\mathcal{T}_{\phi_e}}
        + \lambda \min_{u} \lVert w - u \rVert^2_{\mathcal{T}_{\phi_e}}
        \rbrace \quad s.t. \quad \mathcal{T}_{\phi_e}(u) = \mathbf{y}_{t_i}
    \end{equation}

    where $\mathcal{T}_{\phi_e}$ is the Radon transform sampled on $\phi_e$, the set of
    experimental viewing angles, and $\lVert.\rVert^2_\mathcal{T} = \lVert \mathcal{T(.)}
    \rVert^2_2$. More details on the Radon transform and CT can be found
    in Appendix A. The hyper-parameter $\lambda \in [0, 1]$ balances consistency regarding
    experimental measurements ($\lambda \rightarrow 1$) and \textit{unconditioned} samples
    ($\lambda \rightarrow 0$).

    Song et al.\ \cite{song_solving_2022} demonstrated that Eq.\ \eqref{proximal_step} has
    a tractable closed-formed solution, such that in practice, \textit{conditioned} samples
    are computed through:

    \begin{equation}
    \label{conditioned_sampling}
        \mathbf{x}'_{t_i} = \mathcal{T}_\Phi^\dagger[
            P_{\phi_e} (
                \lambda \mathbf{y}_{t_i}
              + (1 - \lambda) \mathcal{T}_{\phi_e}(\mathbf{x}_{t_i})
            )
          + P_{\phi_s} \circ \mathcal{T}_{\phi_s}(\mathbf{x}_{t_i})
        ]
    \end{equation}

    where the set $\phi_e$ of experimental viewing angles is completed with simulated
    viewing angles $\phi_s$ to constitute an appropriate sampling, $\Phi = \phi_e \cup
    \phi_s$, above the Nyquist criterion, so that the inverse Radon transform
    $\mathcal{T}_\Phi^\dagger$ is well-defined. $P$ denotes a profile padding operator
    ensuring dimensionality consistency.

    Given \textit{conditioned} samples $\mathbf{x}'_{t_i}$, at time-step $t_i$, the same
    iterative sampling strategy as Eq.\ \eqref{unconditioned_iteration} is applied to draw
    \textit{unconditioned} samples $\mathbf{x}_{t_i - 1}$:

    \begin{equation}
    \label{conditioned_iteration}
        \mathbf{x}_{t_{i-1}} = \boldsymbol{h}(
            \mathbf{x}'_{t_i},
            \mathbf{z}_i, s_{\theta^*}(\mathbf{x}_{t_i},t_i)
        )
    \end{equation}

    where $s_{\theta^*}(\mathbf{x},t)$ is the score model trained on $X$ in an unsupervised
    manner, i.e.\ without assuming any measurement process. The reconstruction is
    ultimately computed by iterating sequentially on steps described in Eq.\
    \eqref{conditioned_iteration} and Eq.\ \eqref{conditioned_sampling}.

  \subsection*{Manifold Constraint on the Gradient (MCG)}

    Chung et al.\ \cite{chung_manifold_2022} refined this process by interpreting the
    diffusion process as a transport from the data manifold $\mathbf{x}_0$ to the
    hypersphere where pure noise is concentrated $\mathbf{x}_1$ through interpolating
    manifolds that correspond to the subsequent steps $0 < t < 1$. More specifically, given
    a discrete implementation of the diffusion process reparameterized as $\mathbf{x}_i =
    a_i\mathbf{x}_0 + b_i\mathbf{z}$ with $\mathbf{z} \sim \mathcal{N}(\mathbf{0},
    \boldsymbol{I})$, the application of Tweedie's formula \cite{robbins_empirical_1992}
    gives the following posterior expectation:

    \begin{equation}
    \label{tweedie_formula}
        \hat{\mathbf{x}}_0
      = \mathbb{E}[\mathbf{x}_0 | \mathbf{x}_i]
      = \frac{1}{a_i}(x_i + b_i^2\nabla_{\mathbf{x}_i}\mathrm{log}p_i(\mathbf{x}_i))
    \end{equation}

    From here, Chung et al.\ \cite{chung_manifold_2022} prove that, by assuming
    local linearity of the data manifold $\mathcal{M}$, it is possible to
    design a correction term for data fidelity that does not leave $\mathcal{M}$, yielding
    \textit{corrected} samples:

    \begin{equation}
    \label{manifold_iteration}
        \mathbf{x}^c_{t_i} = \mathbf{x}_t
      - \alpha \nabla_{\mathbf{x}_{t_i}}
        \lVert \mathbf{y}_{\phi_e}\ -
        \mathcal{T}_{\phi_e}(\hat{\mathbf{x}}_0) \rVert^2_2
    \end{equation}

    where $\alpha$ is a hyper-parameter, $\mathbf{y_{\phi_e}}$ and $\mathcal{T}_{\phi_e}$
    are respectively the set of experimental measurements and the Radon transform sampled
    on experimental viewing angles. The \textit{corrected} samples are additionally
    projected onto measurement space to further prevent deviation from data
    consistency. By applying subsequently the steps described in Eq.\
    \eqref{tweedie_formula} and Eq.\ \eqref{manifold_iteration}, MCG diffusion is guided to
    lie on the data manifold resulting in more faithful reconstruction compared to cSGM.

\clearpage

\section*{Appendix C: Effect of the interpolation factor}

\begin{figure}[h!]
  \centering
      \begin{subfigure}[b]{0.49\textwidth}
          \includegraphics[width=\textwidth]{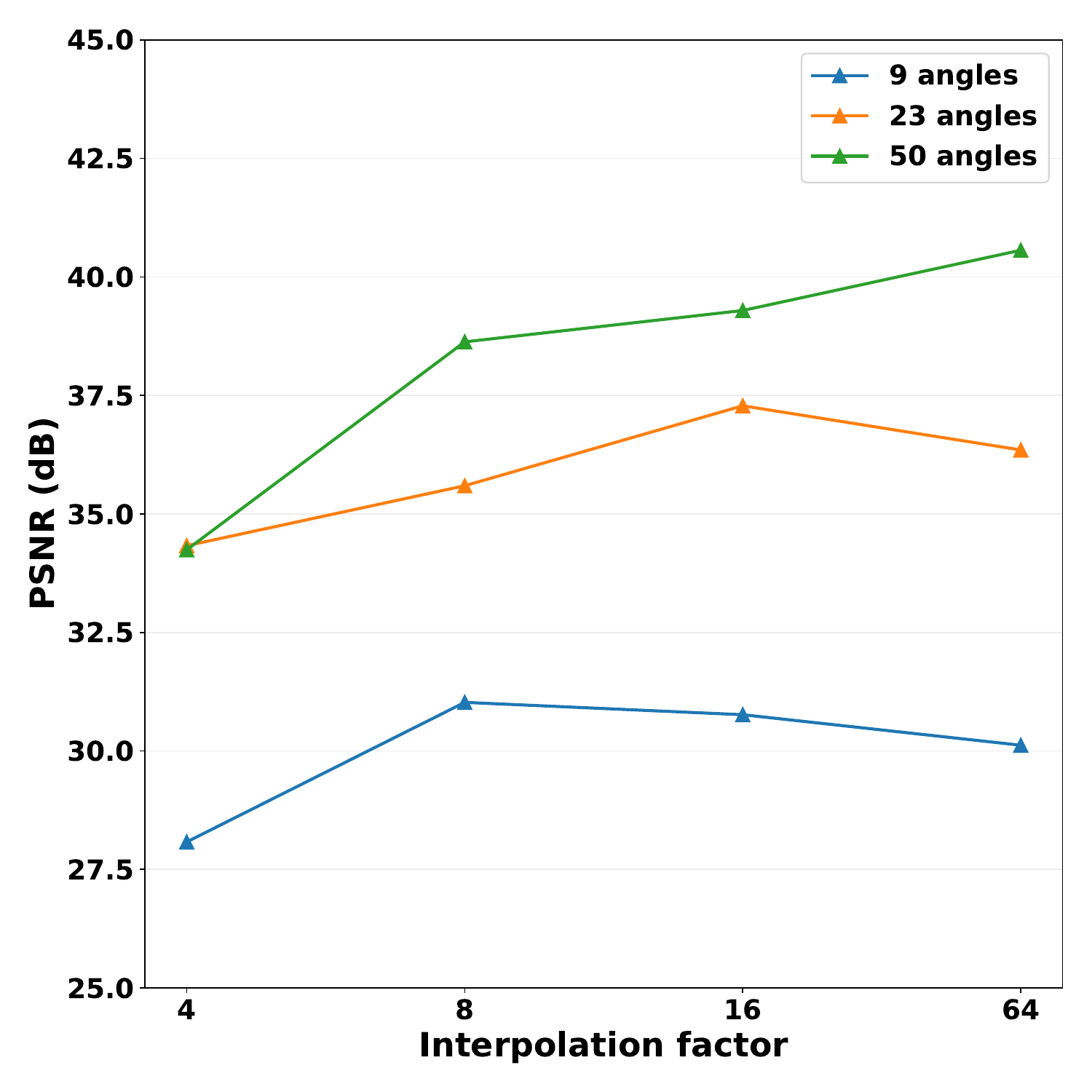}
      \end{subfigure}
      \hfill
      \begin{subfigure}[b]{0.49\textwidth}
          \includegraphics[width=\textwidth]{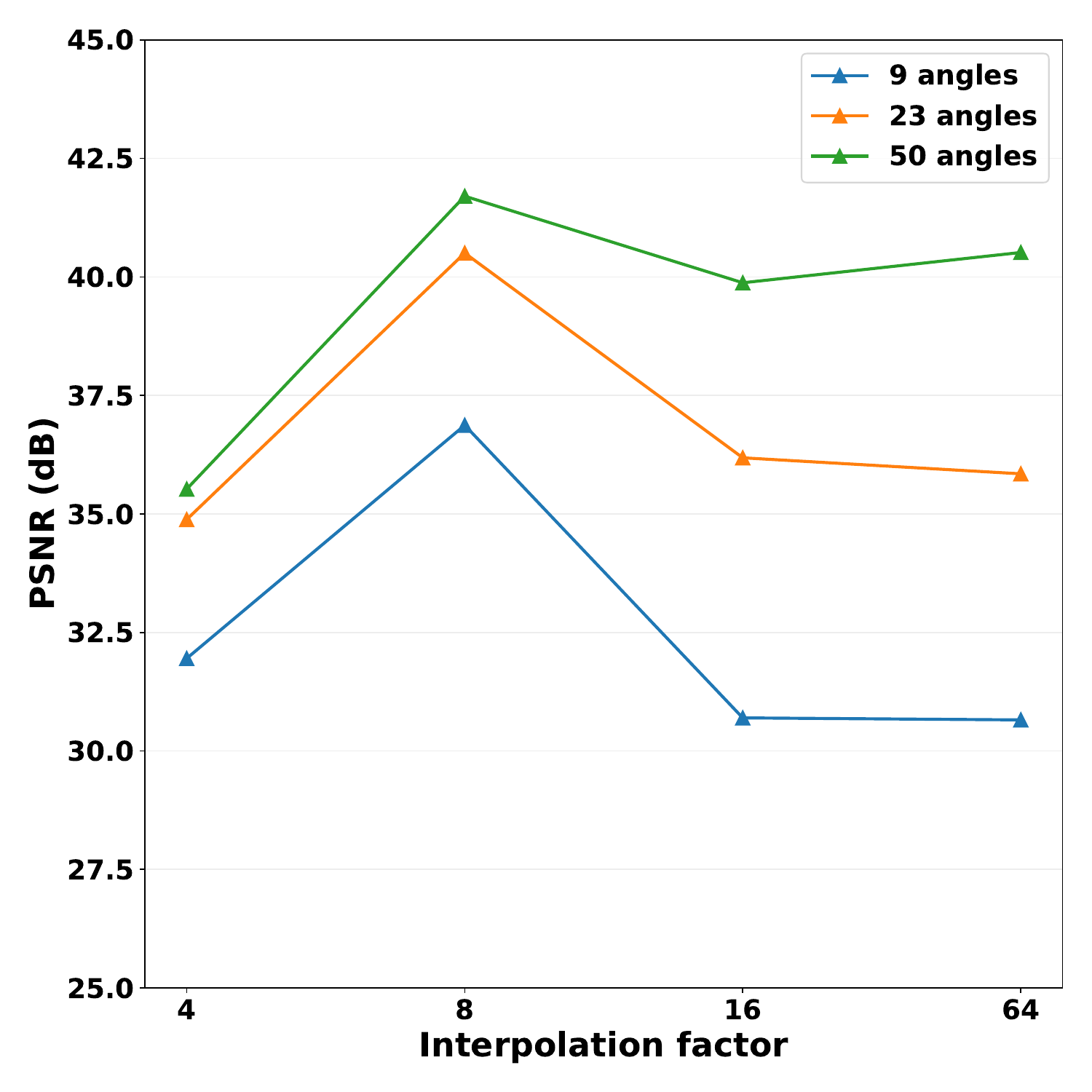}
      \end{subfigure}
  \caption{
      Median PSNR values for reconstructions of slices from the LIDC test set given : 9,
      23 and 50 experimental viewing angles. They are obtained with cGLO without any training
      dataset (left) and trained on the 35\% portion of the LIDC dataset (right).
  }
  \label{fig:interp_exp}
  \end{figure}

  In this study, the impact of the number of experimental profiles provided during
  the optimization process described in Eq.\ \eqref{cglo_objective} on reconstruction
  performances is examined. For this purpose, 5 independent reconstructions, one for each
  patient of the LIDC test set, are conducted for several interpolation factors: 4, 8, 16 and
  64. Independently for each patient, the augmentation is achieved by inserting new
  artificial profiles computed by linear interpolation between the existing experimental
  profiles along the axis orthogonal to the slices plane. The interpolation factor
  corresponds to the number by which the initial profiles quantity is multiplied in the
  process.  Reconstruction quality is evaluated using median PSNR values over the entire LIDC
  test set for varying experimental setups with: 9, 23 and 50 viewing angles. Reconstructions
  computed using artificial profiles are discarded. These quantitative metrics are given in
  figure \ref{fig:interp_exp} for two experiments, one with no prior training and the other
  with the decoder trained on the sub-dataset consisting of a 35\% portion of the LIDC
  dataset.

  The results presented in figure \ref{fig:interp_exp} indicate that the interpolation
  factor has a significant impact on reconstruction quality by inducing great variations
  (several dBs) for each curve, i.e.\ each experimental setup. Concerning experiments
  \textit{with} prior unsupervised training, our model displays a consistent behavior across
  angles with a peak performance for a factor of 8. Also, a greater performance gap benefiting
  higher numbers of viewing angles can be observed for increasing interpolation factors both
  for experiments \textit{with} and \textit{without} prior training.

  The interpolation factor used in our method can be seen as a hyper-parameter balancing the
  learned prior in Eq.\ \eqref{cglo_init}, and the data-likelihood enforcing experimental
  profile consistency during the reconstruction task at examination time in Eq.\
  \eqref{cglo_objective}. Indeed, for sufficiently important interpolation factors (greater
  than 16), performances \textit{with} prior training become similar to those
  \textit{without} any training. It implies that beyond a given interpolation factor value,
  decoder initialization, i.e.\ prior information learned on the unsupervised training
  sub-dataset, is completely overwritten by the second optimization.

  Consequently the interpolation factor should be adapted for each use case, with higher
  values for experimental setups involving more viewing angles and a potential upper limit
  depending on the quantity of training data at hand.

\clearpage

\section*{Appendix D: Additional results tables}

    \begin{table}[h]
    \centering
    \begin{tabular}{c | c | cc | cc}
    \multirow{2}{*}{Method} &
    \multirow{2}{*}{Data} &
    \multicolumn{2}{c|}{LIDC 320$\times$320} &
    \multicolumn{2}{c}{LDCT 512$\times$512} \\\cline{3-6}
    & & PSNR $\uparrow$ & SSIM $\uparrow$ &
        PSNR $\uparrow$ & SSIM $\uparrow$ \\
    \hline \hline
    cSGM & 2\%  & 28.47 \tiny $\pm 1.04$ & 0.784 \tiny $\pm 0.027$ &
                  24.46 \tiny $\pm 1.10$ & 0.664 \tiny $\pm 0.059$\\
     MCG & 2\%  & \textbf{33.05} \tiny $\pm 1.35$ & \underline{0.789} \tiny $\pm 0.020$ &
                  \textbf{33.19} \tiny $\pm 1.15$ & \underline{0.814} \tiny $\pm 0.033$\\
    cGLO & 2\%  & \underline{31.60} \tiny $\pm 1.30$ & \textbf{0.923} \tiny $\pm 0.016$ &
                  \underline{30.53} \tiny $\pm 0.98$ & \textbf{0.903} \tiny $\pm 0.014$\\
    \hline
    cSGM & 10\% & 29.25 \tiny $\pm 1.00$ & 0.800 \tiny $\pm 0.020$ &
                  26.91 \tiny $\pm 1.08$ & 0.744 \tiny $\pm 0.041$\\
     MCG & 10\% & \underline{34.88} \tiny $\pm 1.12$ & \underline{0.848} \tiny $\pm 0.013$ &
                  \textbf{33.20} \tiny $\pm 1.07$ & \underline{0.821} \tiny $\pm 0.024$\\
    cGLO & 10\% & \textbf{35.31} \tiny $\pm 0.94$ & \textbf{0.957} \tiny $\pm 0.008$ &
                  \underline{31.35} \tiny $\pm 0.99$ & \textbf{0.917} \tiny $\pm 0.017$\\
    \hline
    cSGM & 35\% & 30.28 \tiny $\pm 1.13$ & 0.821 \tiny $\pm 0.031$ &
                  27.59 \tiny $\pm 1.31$ & 0.759 \tiny $\pm 0.038$\\
     MCG & 35\% & \textbf{35.45} \tiny $\pm 1.19$ & \underline{0.871} \tiny $\pm 0.017$ &
                  \textbf{34.11} \tiny $\pm 1.11$ & \underline{0.841} \tiny $\pm 0.019$\\
    cGLO & 35\% & \underline{35.11} \tiny $\pm 0.97$ & \textbf{0.958} \tiny $\pm 0.006$ &
                  \underline{31.19} \tiny $\pm 1.08$ & \textbf{0.913} \tiny $\pm 0.016$\\
    \hline
    \end{tabular}
    \vspace{10pt}
    \caption{
        PSNR and SSIM median $\pm$ half Inter Quartile Range (IQR) values
        for reconstructions of slices from the LIDC and the LDCT test datasets given
        9 experimental viewing angles. Best and second best results are respectively
        indicated as bold and underline for each sub-dataset and metric.
    }
    \label{tab:results_9_angles}
    \end{table}

    \begin{table}[h]
    \centering
    \begin{tabular}{c | c | cc | cc}
    \multirow{2}{*}{Method} &
    \multirow{2}{*}{Data} &
    \multicolumn{2}{c|}{LIDC 320$\times$320} &
    \multicolumn{2}{c}{LDCT 512$\times$512} \\\cline{3-6}
    & & PSNR $\uparrow$ & SSIM $\uparrow$ &
        PSNR $\uparrow$ & SSIM $\uparrow$ \\
    \hline \hline
    cSGM & 2\%  & 33.89 \tiny $\pm 1.01$ & 0.879 \tiny $\pm 0.019$ &
                  29.09 \tiny $\pm 0.61$ & 0.793 \tiny $\pm 0.018$\\
     MCG & 2\%  & \underline{36.55} \tiny $\pm 1.23$ & \underline{0.888} \tiny $\pm 0.012$ &
                  \textbf{35.76} \tiny $\pm 1.41$ & \underline{0.871} \tiny $\pm 0.015$\\
    cGLO & 2\%  & \textbf{37.60} \tiny $\pm 1.67$ & \textbf{0.972} \tiny $\pm 0.009$ &
                  \underline{35.68} \tiny $\pm 1.03$ & \textbf{0.955} \tiny $\pm 0.010$\\
    \hline
    cSGM & 10\% & 36.31 \tiny $\pm 0.79$ & 0.912 \tiny $\pm 0.015$ &
                  33.35 \tiny $\pm 1.02$ & \underline{0.887} \tiny $\pm 0.017$\\
     MCG & 10\% & \underline{38.92} \tiny $\pm 0.92$ & \underline{0.923} \tiny $\pm 0.010$ &
                  \underline{36.45} \tiny $\pm 1.28$ & 0.879 \tiny $\pm 0.013$\\
    cGLO & 10\% & \textbf{39.82} \tiny $\pm 1.26$ & \textbf{0.981} \tiny $\pm 0.006$ &
                  \textbf{37.34} \tiny $\pm 1.40$ & \textbf{0.967} \tiny $\pm 0.010$\\
    \hline
    cSGM & 35\% & 36.62 \tiny $\pm 1.61$ & 0.916 \tiny $\pm 0.024$ &
                  34.90 \tiny $\pm 1.08$ & 0.907 \tiny $\pm 0.015$\\
     MCG & 35\% & \underline{39.60} \tiny $\pm 1.17$ & \underline{0.933} \tiny $\pm 0.014$ &
                  \textbf{38.58} \tiny $\pm 1.37$ & \underline{0.910} \tiny $\pm 0.013$\\
    cGLO & 35\% & \textbf{39.85} \tiny $\pm 1.19$ & \textbf{0.982} \tiny $\pm 0.005$ &
                  \underline{36.61} \tiny $\pm 1.45$ & \textbf{0.962} \tiny $\pm 0.010$\\
    \hline
    \end{tabular}
    \vspace{10pt}
    \caption{
        PSNR and SSIM median $\pm$ half Inter Quartile Range (IQR) values
        for reconstructions of slices from the LIDC and the LDCT test datasets given
        23 experimental viewing angles. Best and second best results are respectively
        indicated as bold and underline for each sub-dataset and metric.
    }
    \label{tab:results_23_angles}
    \end{table}

    \begin{table}[h]
    \centering
    \begin{tabular}{c | c | cc | cc}
    \multirow{2}{*}{Method} &
    \multirow{2}{*}{Data} &
    \multicolumn{2}{c|}{LIDC 320$\times$320} &
    \multicolumn{2}{c}{LDCT 512$\times$512} \\\cline{3-6}
    & & PSNR $\uparrow$ & SSIM $\uparrow$ &
        PSNR $\uparrow$ & SSIM $\uparrow$ \\
    \hline \hline
    cSGM & 2\%  & 36.90 \tiny $\pm 0.83$ & \underline{0.922} \tiny $\pm 0.016$ &
                  32.46 \tiny $\pm 0.56$ & 0.867 \tiny $\pm 0.015$\\
     MCG & 2\%  & \underline{37.92} \tiny $\pm 0.85$ & 0.909 \tiny $\pm 0.008$ &
                  \underline{37.01} \tiny $\pm 1.46$ & \underline{0.894} \tiny $\pm 0.009$\\
    cGLO & 2\%  & \textbf{40.50} \tiny $\pm 1.56$ & \textbf{0.983} \tiny $\pm 0.007$ &
                  \textbf{38.24} \tiny $\pm 1.05$ & \textbf{0.972} \tiny $\pm 0.007$\\
    \hline
    cSGM & 10\% & \underline{40.03} \tiny $\pm 1.05$ & \underline{0.950} \tiny $\pm 0.016$ &
                  36.78 \tiny $\pm 0.96$ & \underline{0.937} \tiny $\pm 0.013$\\
     MCG & 10\% & 40.02 \tiny $\pm 1.10$ & 0.936 \tiny $\pm 0.009$ &
                  \underline{38.55} \tiny $\pm 1.38$ & 0.913 \tiny $\pm 0.011$\\
    cGLO & 10\% & \textbf{41.45} \tiny $\pm 1.53$ & \textbf{0.985} \tiny $\pm 0.006$ &
                  \textbf{39.73} \tiny $\pm 1.31$ & \textbf{0.979} \tiny $\pm 0.007$\\
    \hline
    cSGM & 35\% & 40.04 \tiny $\pm 1.45$ & \underline{0.950} \tiny $\pm 0.019$ &
                  38.37 \tiny $\pm 1.05$ & \underline{0.953} \tiny $\pm 0.010$\\
     MCG & 35\% & \underline{40.79} \tiny $\pm 1.21$ & 0.943 \tiny $\pm 0.008$ &
                  \textbf{39.81} \tiny $\pm 1.42$ & 0.939 \tiny $\pm 0.010$\\
    cGLO & 35\% & \textbf{40.89} \tiny $\pm 1.38$ & \textbf{0.984} \tiny $\pm 0.006$ &
                  \underline{39.48} \tiny $\pm 1.43$ & \textbf{0.978} \tiny $\pm 0.008$\\
    \hline
    \end{tabular}
    \vspace{10pt}
    \caption{
        PSNR and SSIM median $\pm$ half Inter Quartile Range (IQR) values
        for reconstructions of slices from the LIDC and the LDCT test datasets given
        50 experimental viewing angles. Best and second best results are respectively
        indicated as bold and underline for each sub-dataset and metric.
    }
    \label{tab:results_50_angles}
    \end{table}

\clearpage

\section*{Appendix E: Additional reconstruction examples}

    \begin{figure}[h!]
    \centering
        \begin{subfigure}[b]{0.245\textwidth}
            \includegraphics[width=\textwidth]{images/GTX.pdf}
        \end{subfigure}
        \hfill
        \begin{subfigure}[b]{0.245\textwidth}
            \includegraphics[width=\textwidth]{images/img_csgm_lidc/csgm_x2_a9.pdf}
        \end{subfigure}
        \hfill
        \begin{subfigure}[b]{0.245\textwidth}
            \includegraphics[width=\textwidth]{images/img_csgm_lidc/csgm_x2_a23.pdf}
        \end{subfigure}
        \hfill
        \begin{subfigure}[b]{0.245\textwidth}
            \includegraphics[width=\textwidth]{images/img_csgm_lidc/csgm_x2_a50.pdf}
        \end{subfigure}

        \begin{subfigure}[b]{0.245\textwidth}
            \includegraphics[width=\textwidth]{images/GTX.pdf}
        \end{subfigure}
        \hfill
        \begin{subfigure}[b]{0.245\textwidth}
            \includegraphics[width=\textwidth]{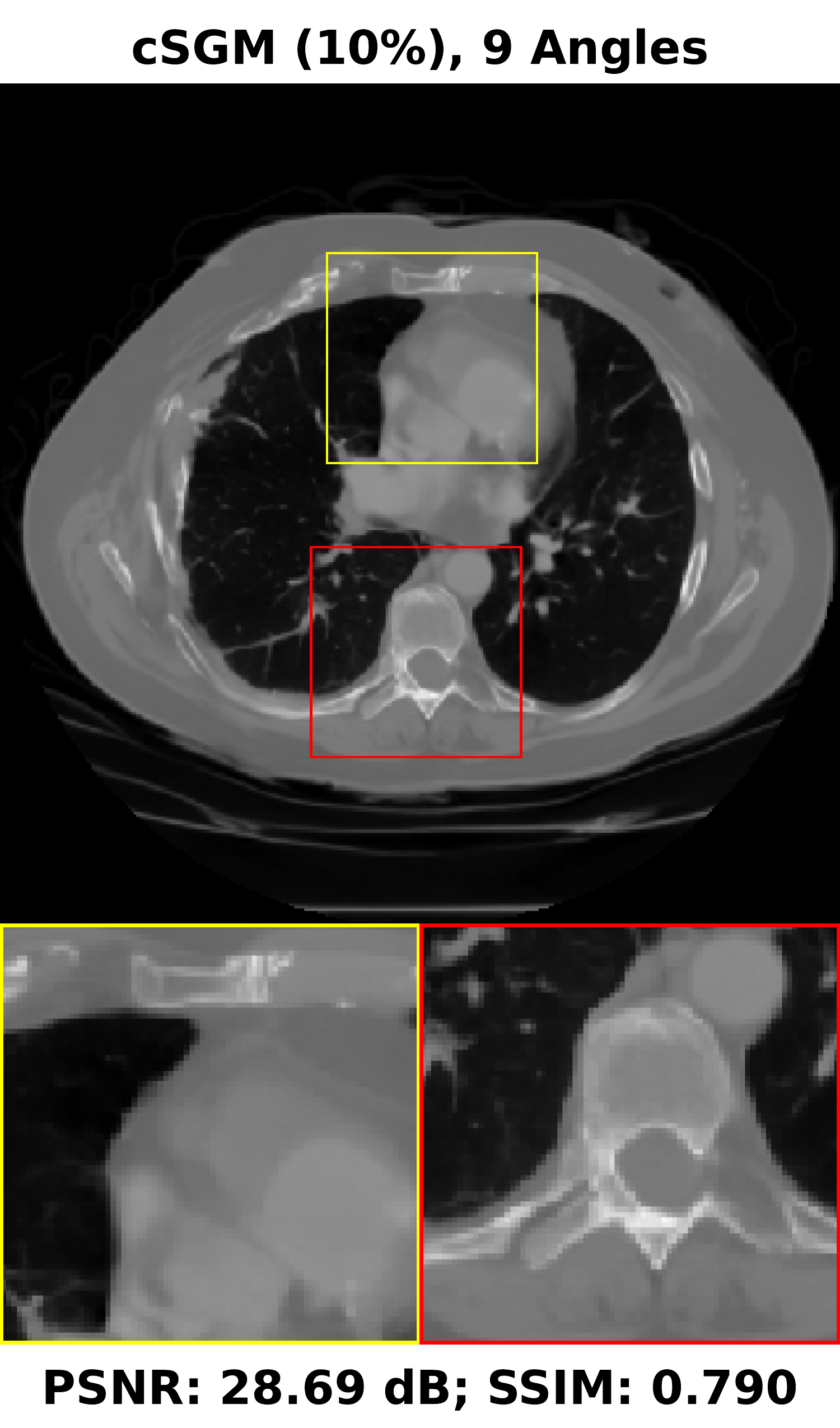}
        \end{subfigure}
        \hfill
        \begin{subfigure}[b]{0.245\textwidth}
            \includegraphics[width=\textwidth]{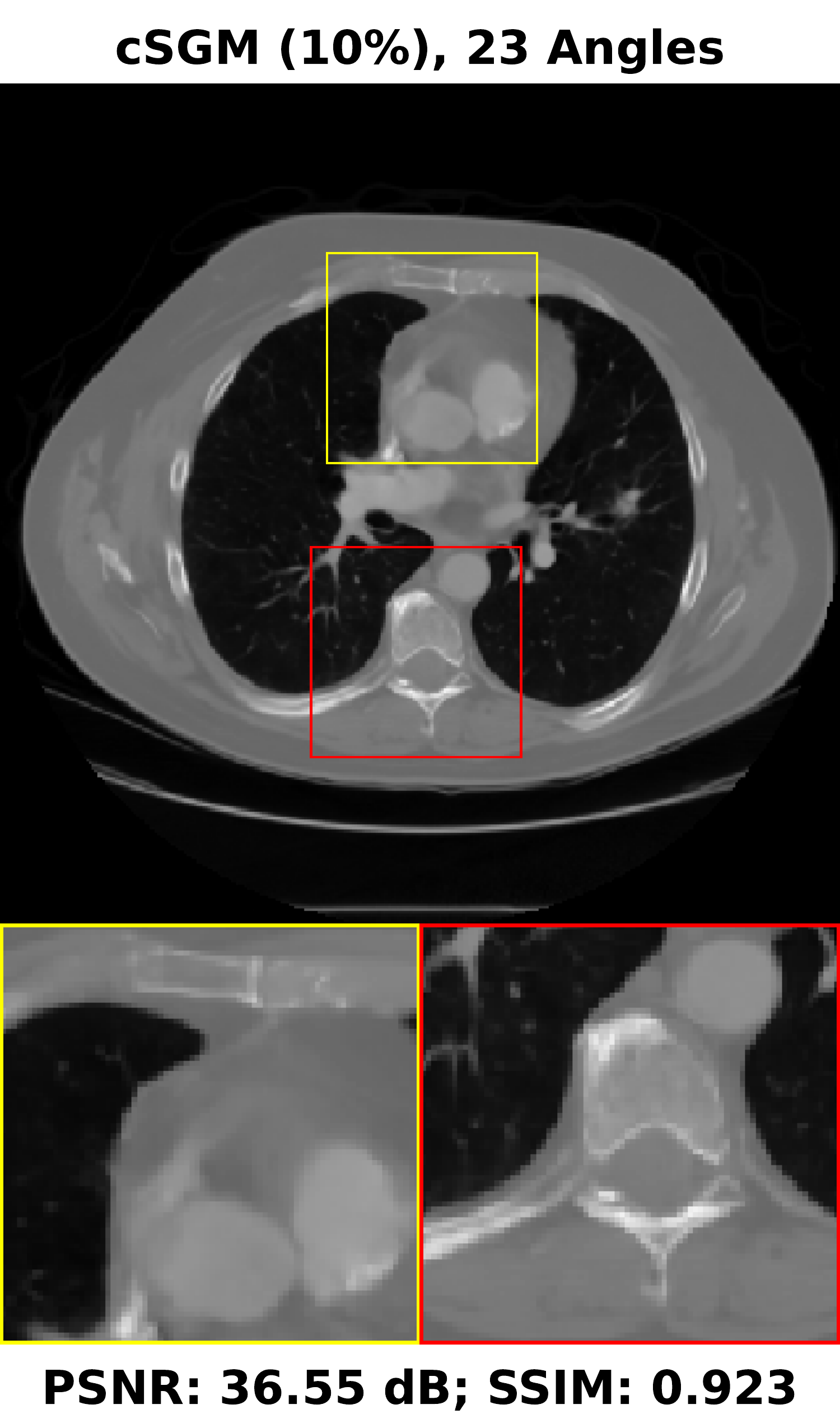}
        \end{subfigure}
        \hfill
        \begin{subfigure}[b]{0.245\textwidth}
            \includegraphics[width=\textwidth]{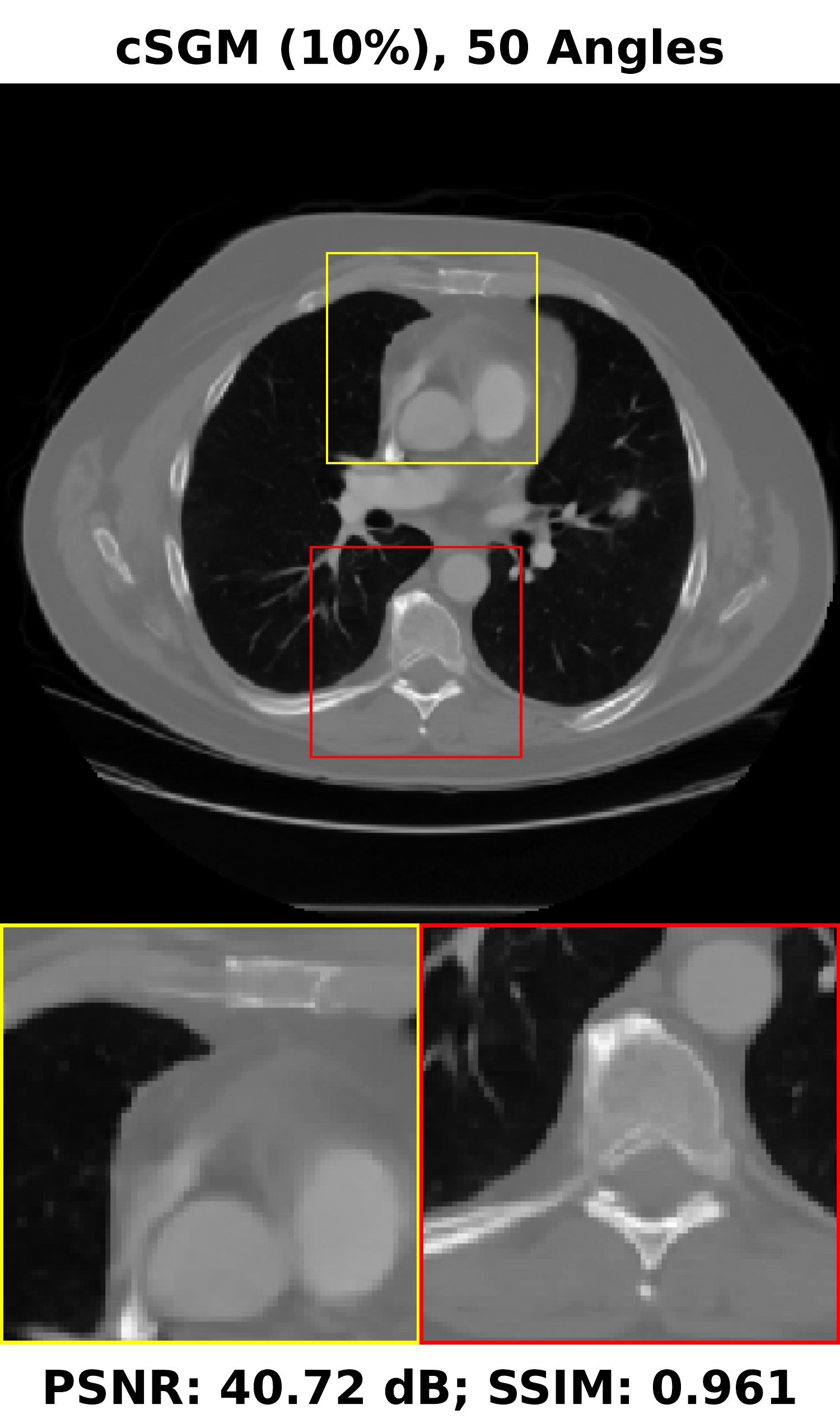}
        \end{subfigure}

        \begin{subfigure}[b]{0.245\textwidth}
            \includegraphics[width=\textwidth]{images/GTX.pdf}
        \end{subfigure}
        \hfill
        \begin{subfigure}[b]{0.245\textwidth}
            \includegraphics[width=\textwidth]{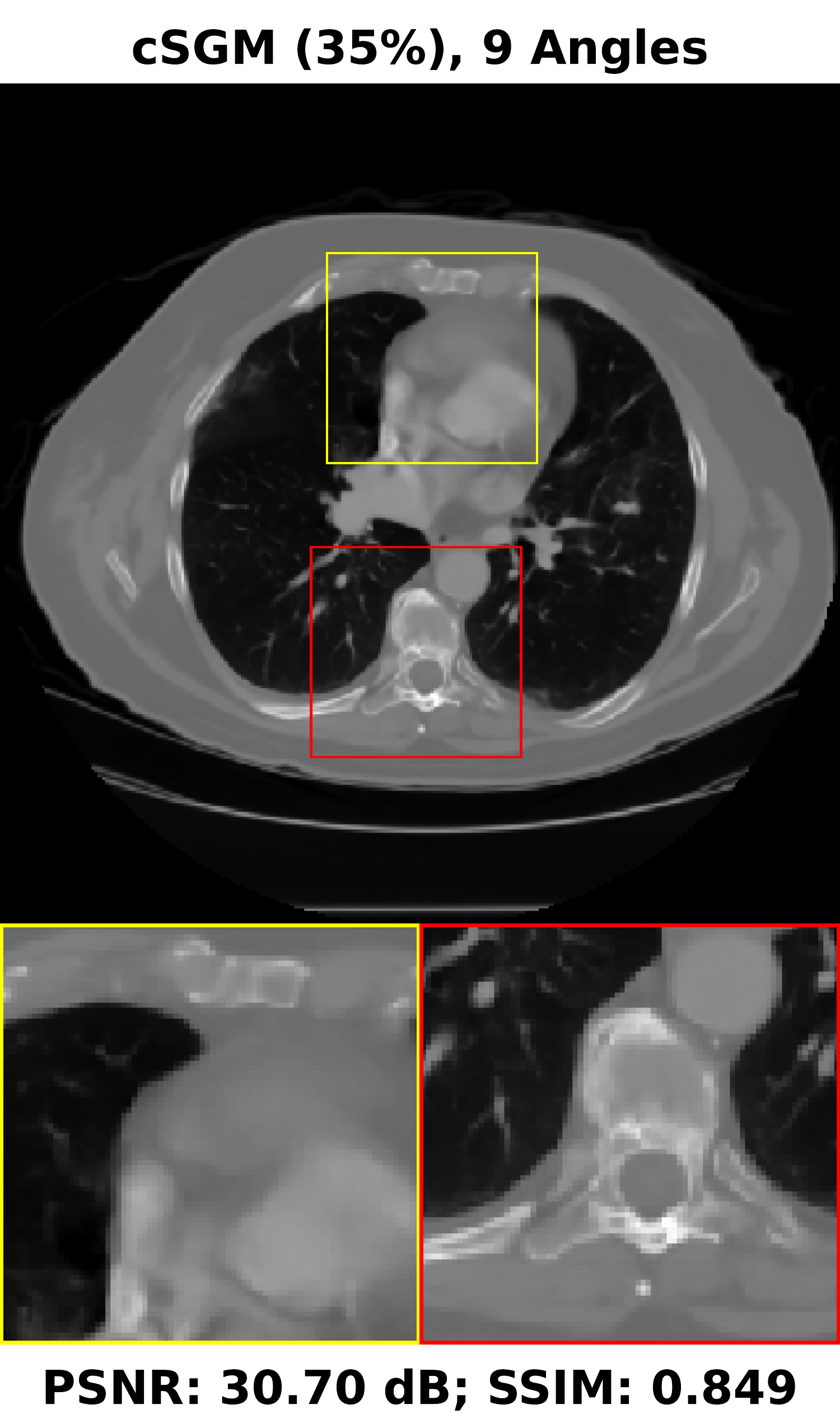}
        \end{subfigure}
        \hfill
        \begin{subfigure}[b]{0.245\textwidth}
            \includegraphics[width=\textwidth]{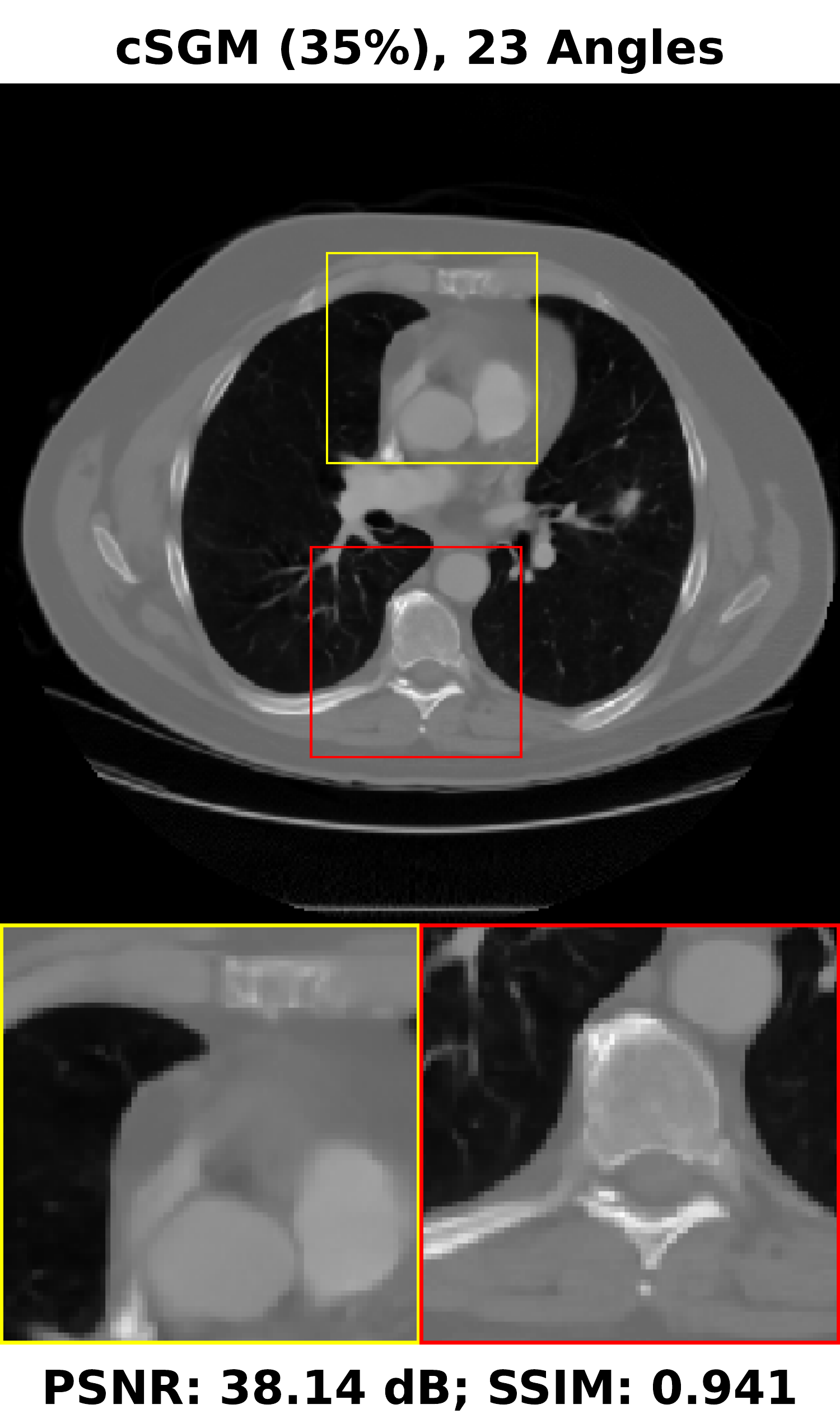}
        \end{subfigure}
        \hfill
        \begin{subfigure}[b]{0.245\textwidth}
            \includegraphics[width=\textwidth]{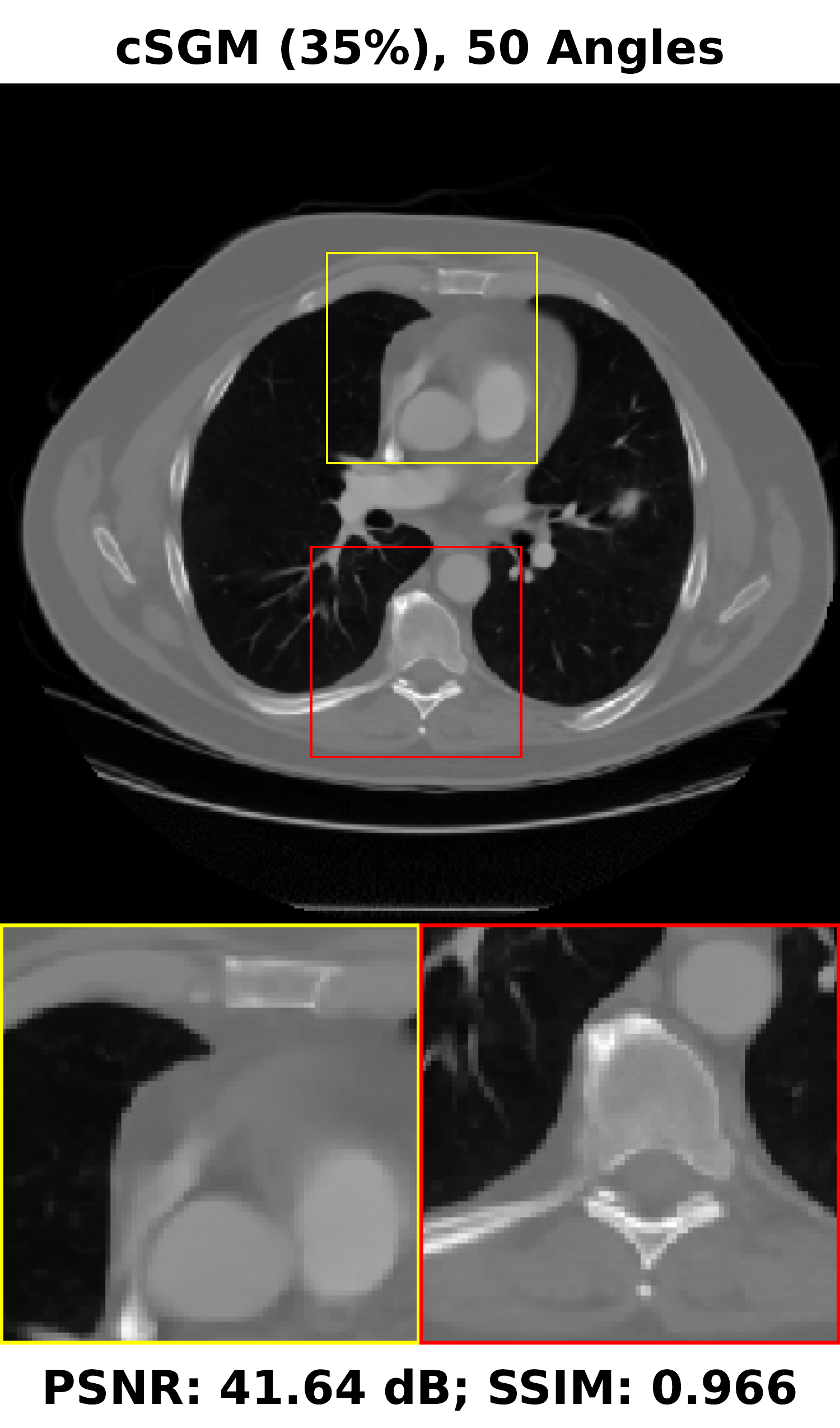}
        \end{subfigure}
    \caption{
        Examples of reconstructions of slices from the LIDC test set given : 9,
        23 and 50 experimental viewing angles. They are obtained with cSGM
        trained on the 2\%, 10\% and 35\% LIDC sub-datasets.
    }
    \label{fig:csgm_lidc_ex}
    \end{figure}

    \begin{figure}[h!]
    \centering
        \begin{subfigure}[b]{0.245\textwidth}
            \includegraphics[width=\textwidth]{images/GTX.pdf}
        \end{subfigure}
        \hfill
        \begin{subfigure}[b]{0.245\textwidth}
            \includegraphics[width=\textwidth]{images/img_mcg_lidc/mcg_x2_a9.pdf}
        \end{subfigure}
        \hfill
        \begin{subfigure}[b]{0.245\textwidth}
            \includegraphics[width=\textwidth]{images/img_mcg_lidc/mcg_x2_a23.pdf}
        \end{subfigure}
        \hfill
        \begin{subfigure}[b]{0.245\textwidth}
            \includegraphics[width=\textwidth]{images/img_mcg_lidc/mcg_x2_a50.pdf}
        \end{subfigure}

        \begin{subfigure}[b]{0.245\textwidth}
            \includegraphics[width=\textwidth]{images/GTX.pdf}
        \end{subfigure}
        \hfill
        \begin{subfigure}[b]{0.245\textwidth}
            \includegraphics[width=\textwidth]{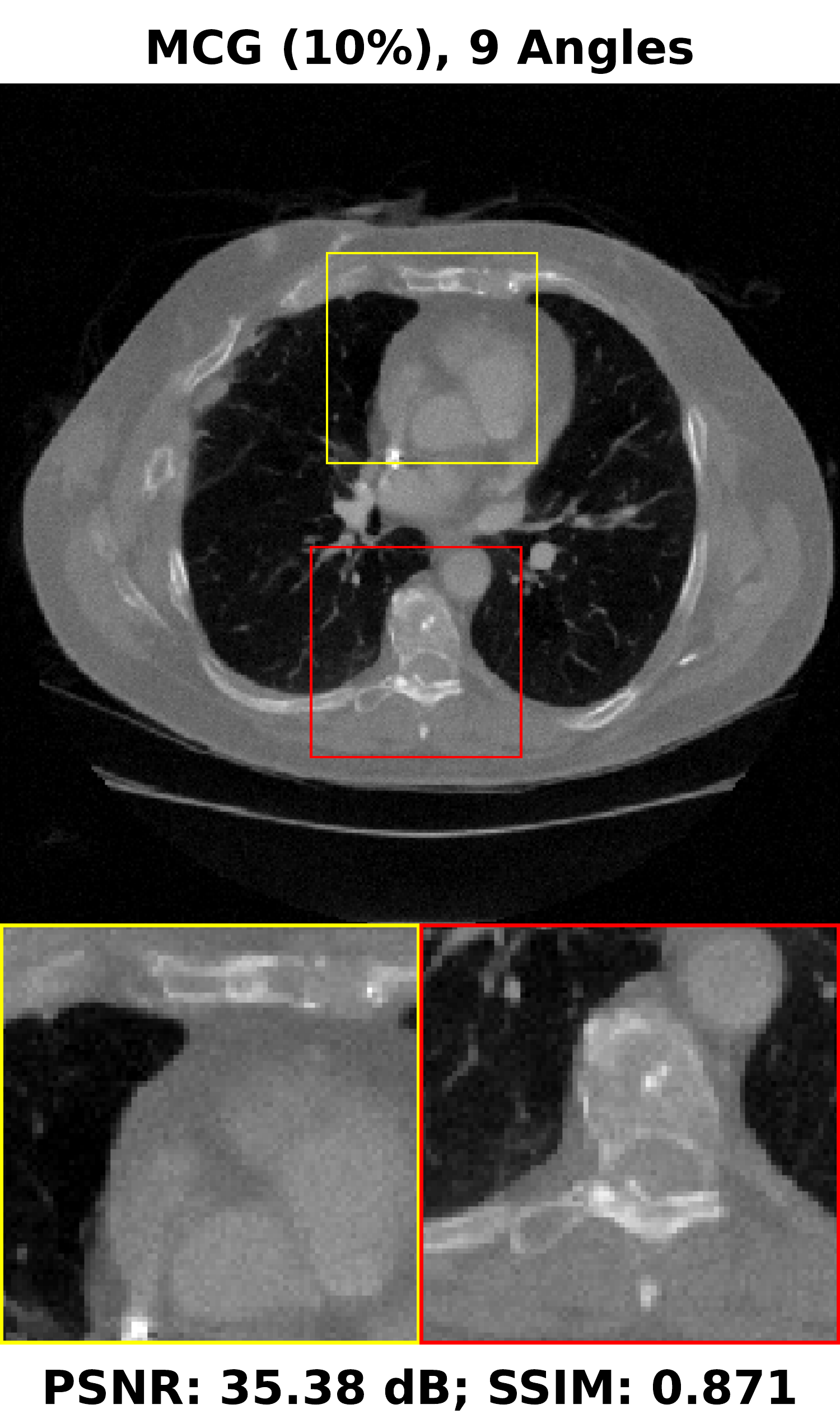}
        \end{subfigure}
        \hfill
        \begin{subfigure}[b]{0.245\textwidth}
            \includegraphics[width=\textwidth]{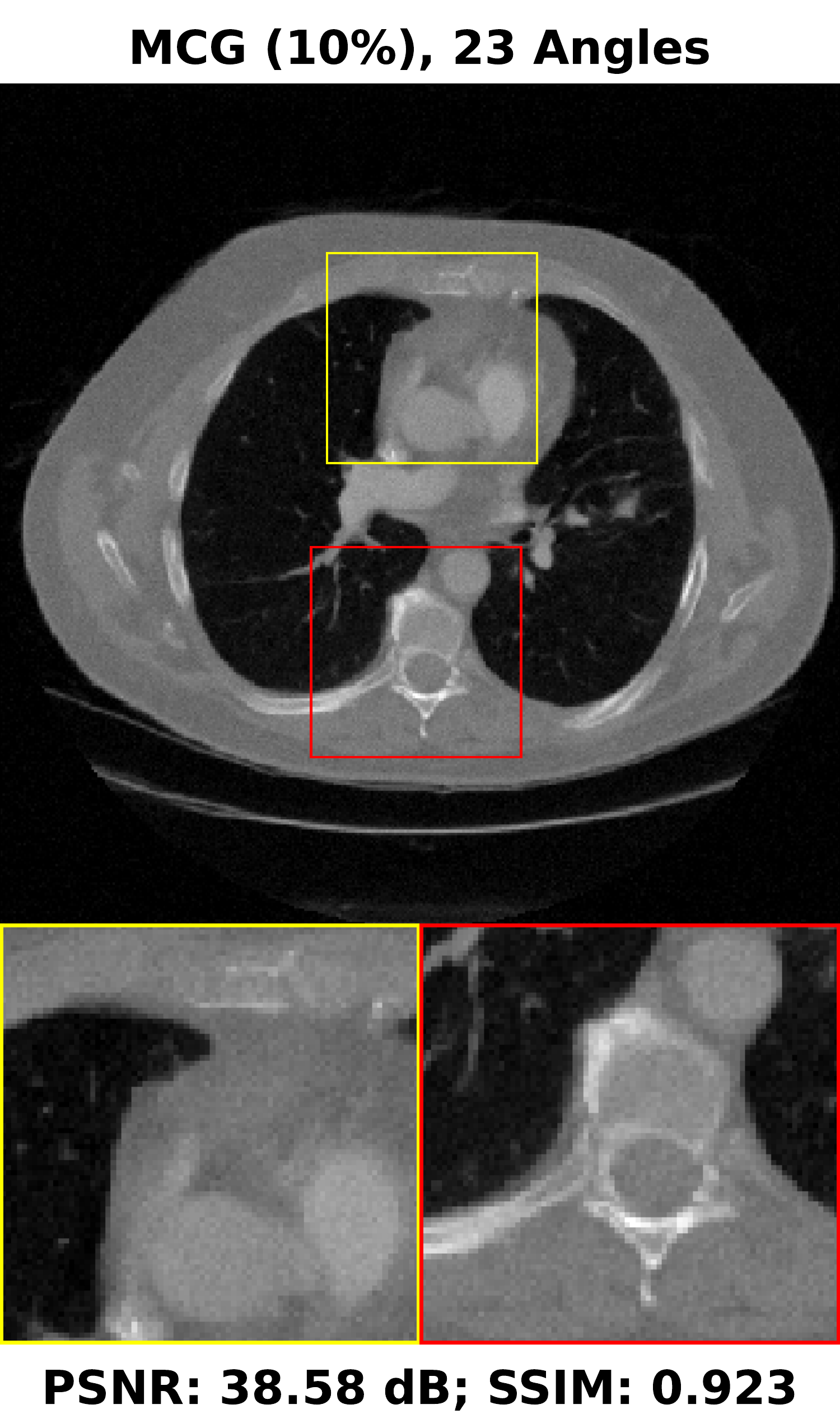}
        \end{subfigure}
        \hfill
        \begin{subfigure}[b]{0.245\textwidth}
            \includegraphics[width=\textwidth]{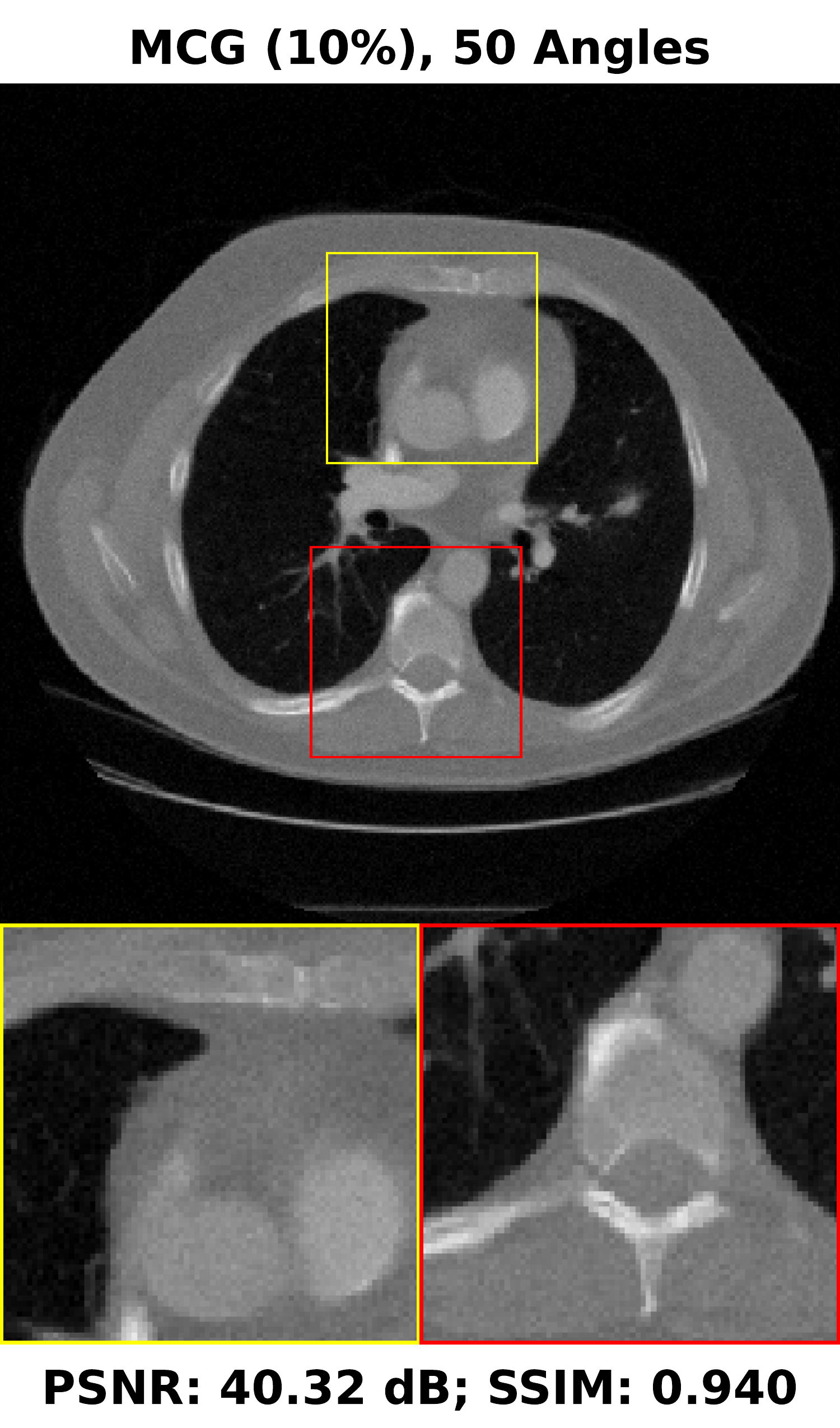}
        \end{subfigure}

        \begin{subfigure}[b]{0.245\textwidth}
            \includegraphics[width=\textwidth]{images/GTX.pdf}
        \end{subfigure}
        \hfill
        \begin{subfigure}[b]{0.245\textwidth}
            \includegraphics[width=\textwidth]{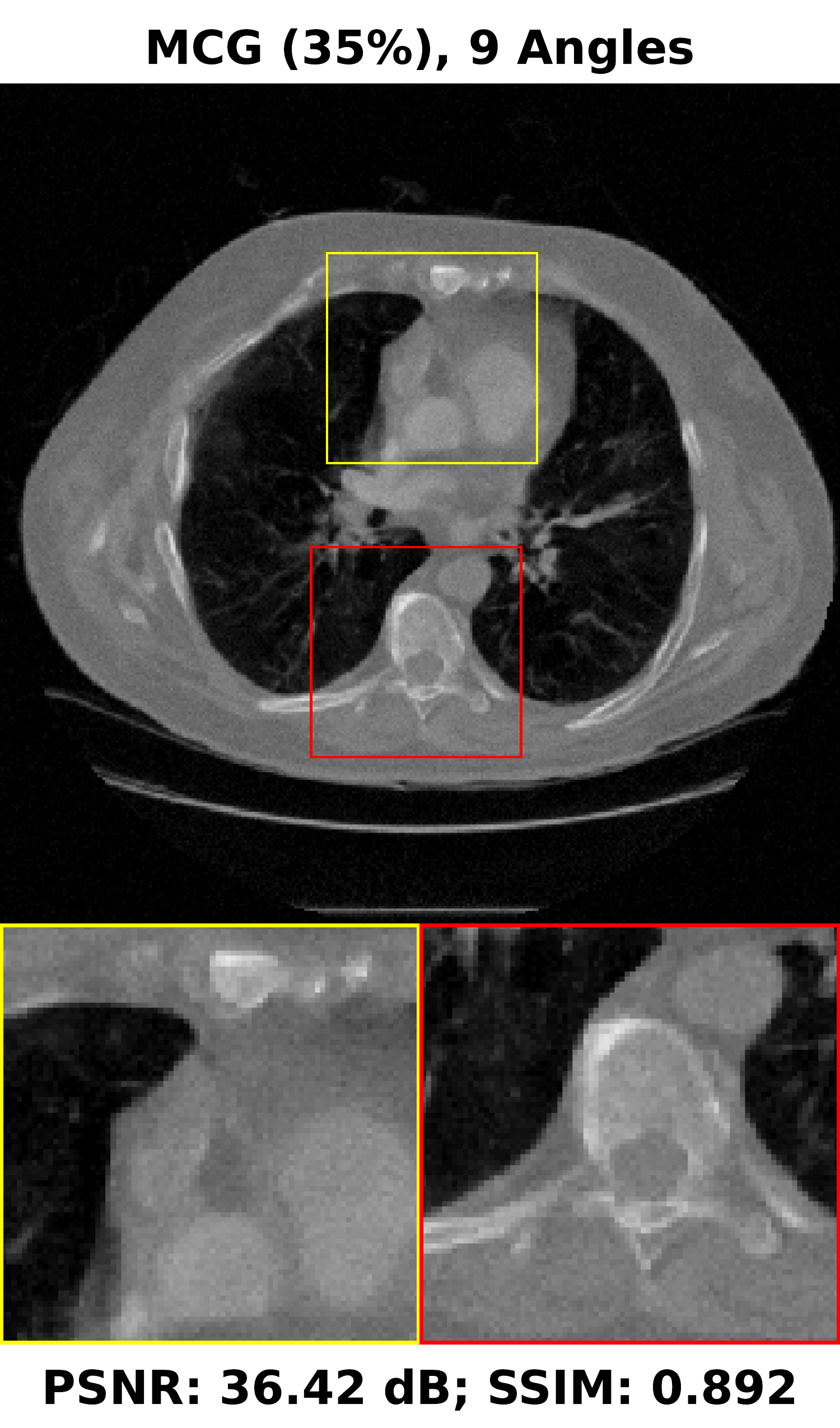}
        \end{subfigure}
        \hfill
        \begin{subfigure}[b]{0.245\textwidth}
            \includegraphics[width=\textwidth]{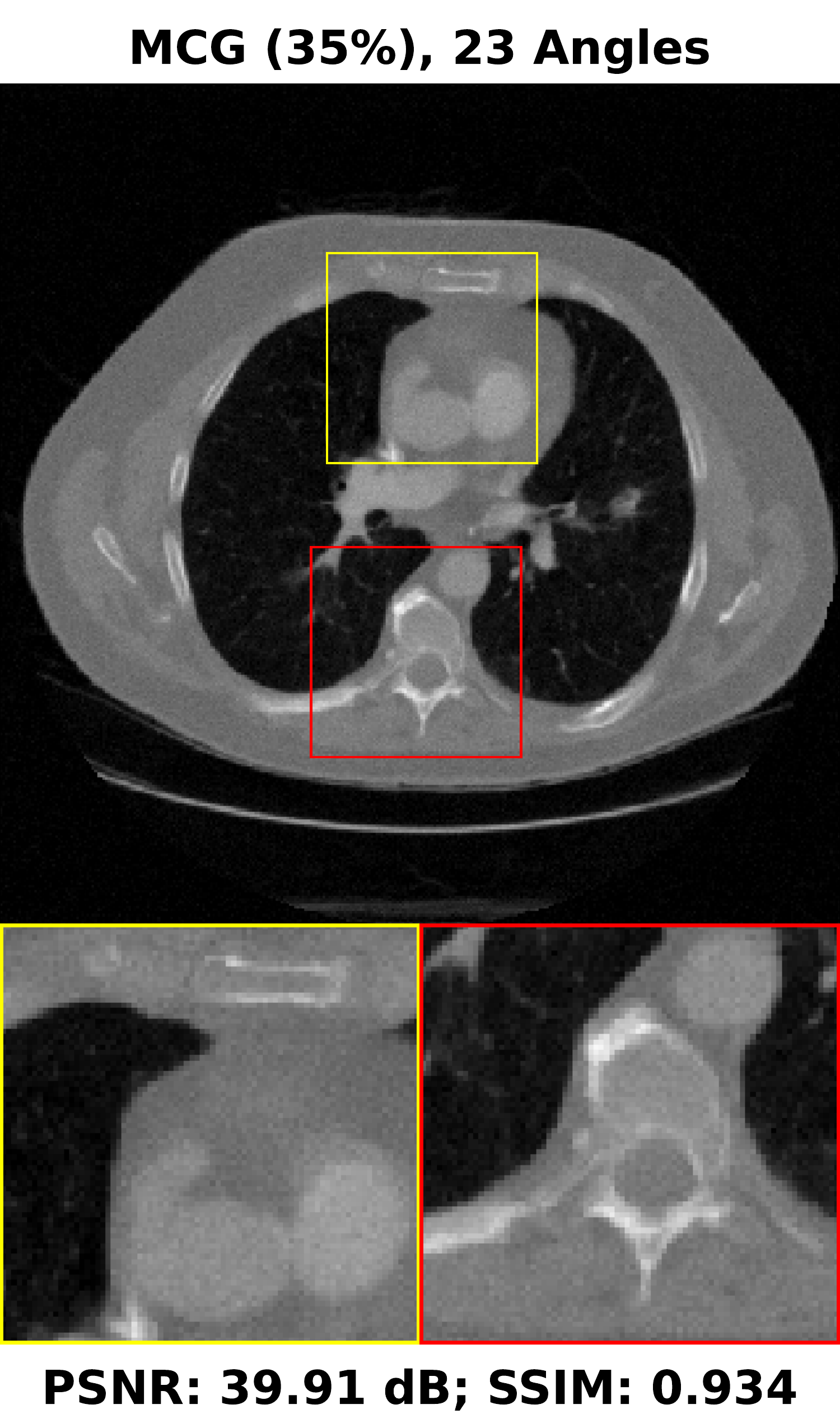}
        \end{subfigure}
        \hfill
        \begin{subfigure}[b]{0.245\textwidth}
            \includegraphics[width=\textwidth]{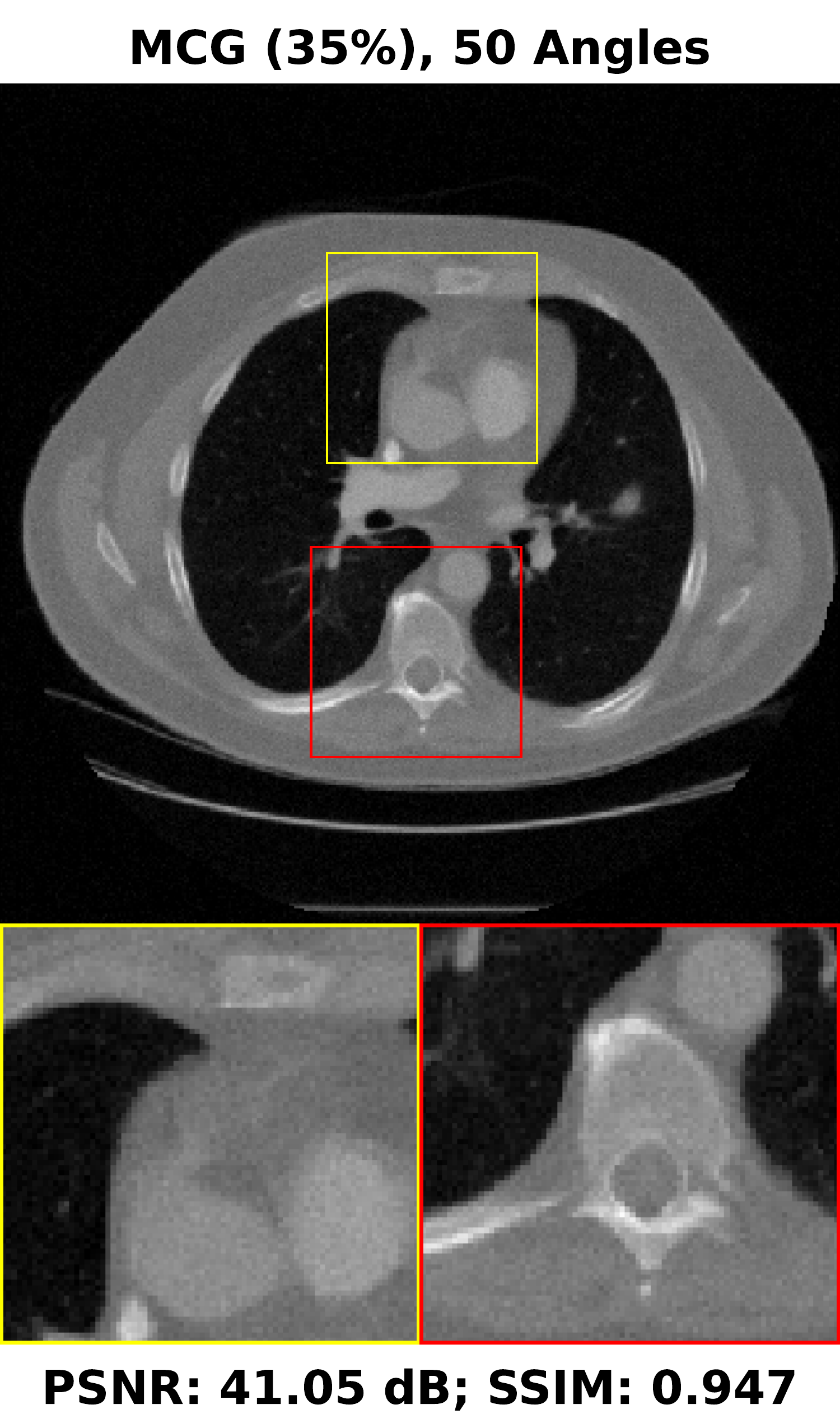}
        \end{subfigure}
    \caption{
        Examples of reconstructions of slices from the LIDC test set given : 9,
        23 and 50 experimental viewing angles. They are obtained with MCG
        trained on the 2\%, 10\% and 35\% LIDC sub-datasets.
    }
    \label{fig:mcg_lidc_ex}
    \end{figure}

    \begin{figure}[h!]
    \centering
        \begin{subfigure}[b]{0.245\textwidth}
            \includegraphics[width=\textwidth]{images/GTX.pdf}
        \end{subfigure}
        \hfill
        \begin{subfigure}[b]{0.245\textwidth}
            \includegraphics[width=\textwidth]{images/img_cglo_lidc/cglo_x2_a9.pdf}
        \end{subfigure}
        \hfill
        \begin{subfigure}[b]{0.245\textwidth}
            \includegraphics[width=\textwidth]{images/img_cglo_lidc/cglo_x2_a23.pdf}
        \end{subfigure}
        \hfill
        \begin{subfigure}[b]{0.245\textwidth}
            \includegraphics[width=\textwidth]{images/img_cglo_lidc/cglo_x2_a50.pdf}
        \end{subfigure}

        \begin{subfigure}[b]{0.245\textwidth}
            \includegraphics[width=\textwidth]{images/GTX.pdf}
        \end{subfigure}
        \hfill
        \begin{subfigure}[b]{0.245\textwidth}
            \includegraphics[width=\textwidth]{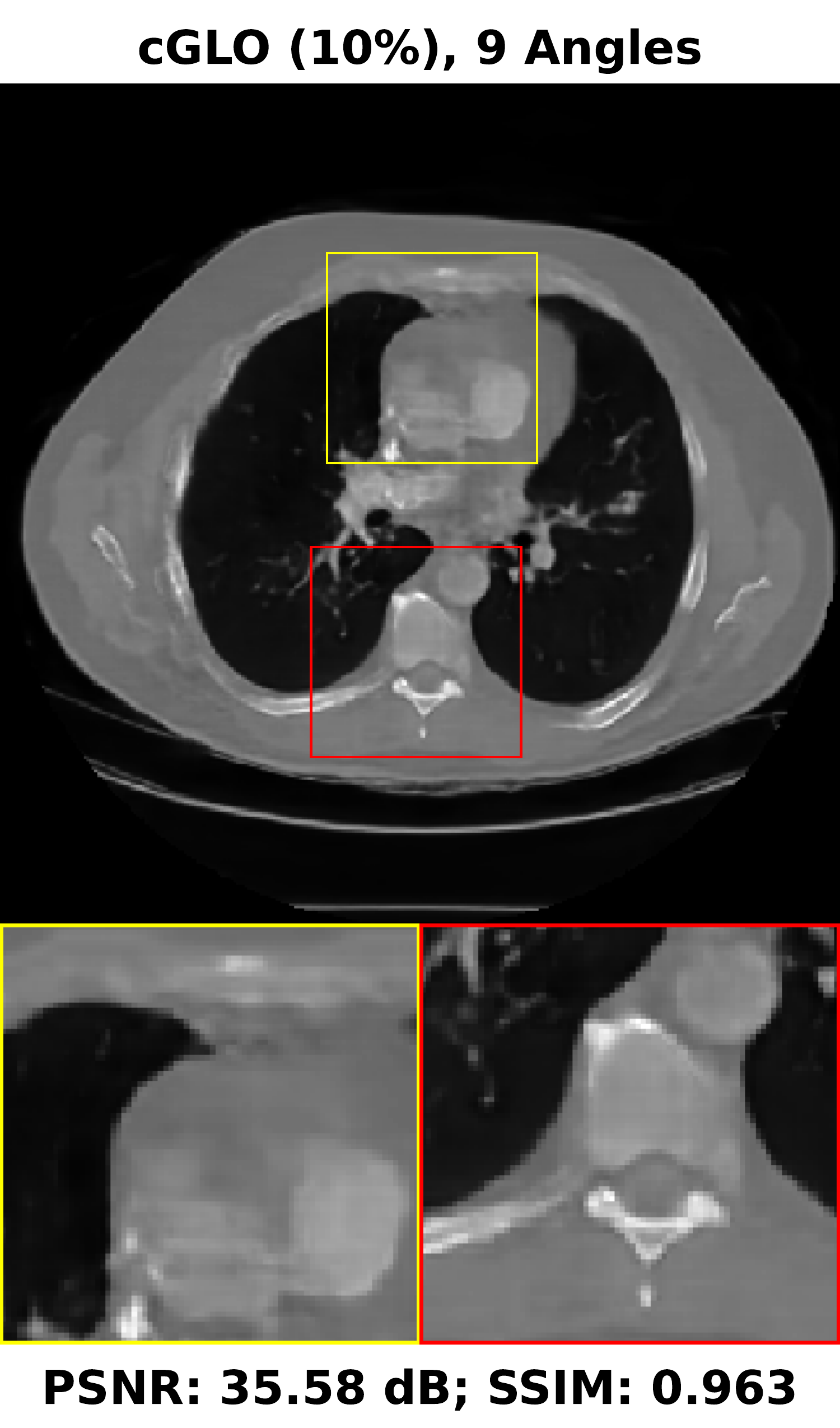}
        \end{subfigure}
        \hfill
        \begin{subfigure}[b]{0.245\textwidth}
            \includegraphics[width=\textwidth]{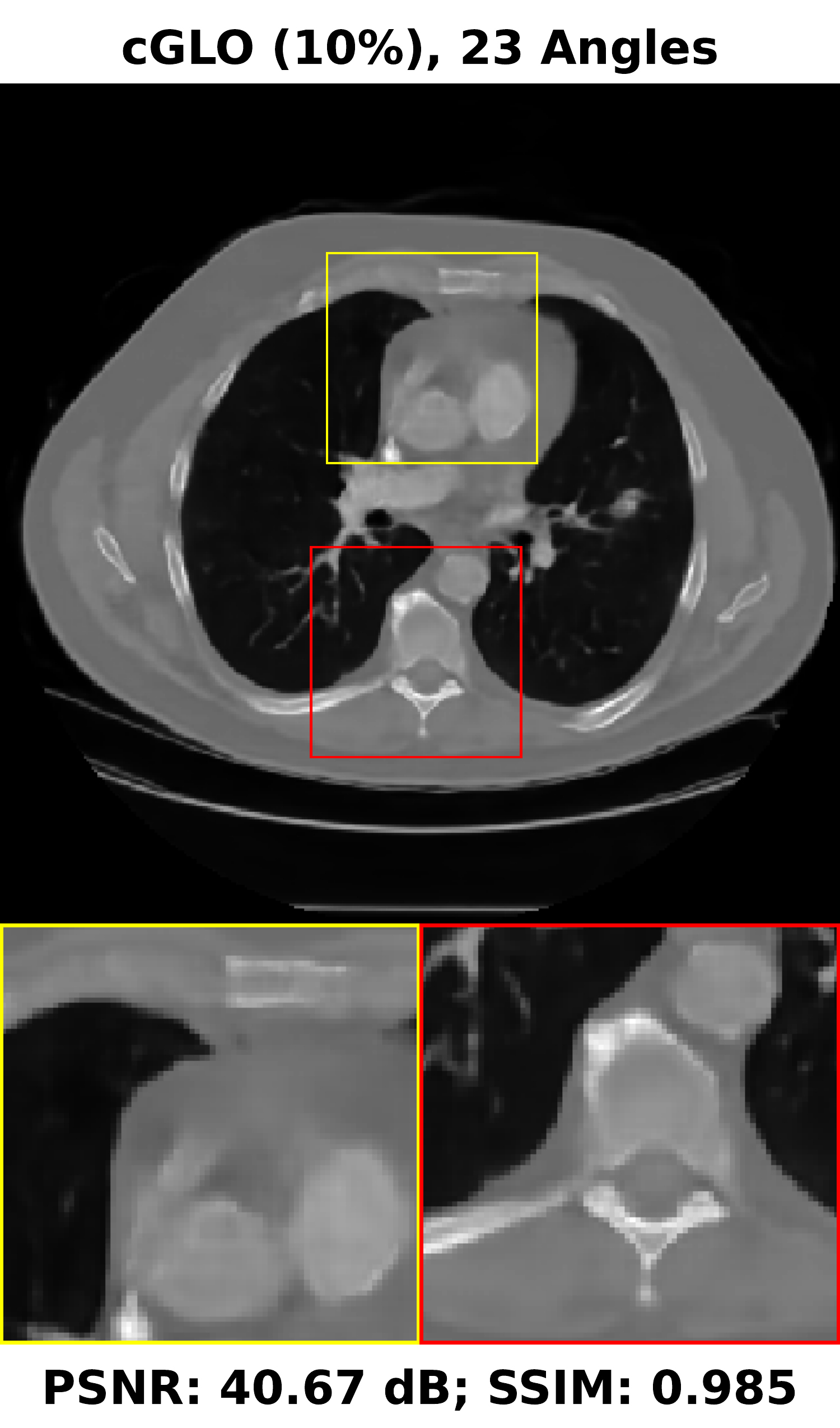}
        \end{subfigure}
        \hfill
        \begin{subfigure}[b]{0.245\textwidth}
            \includegraphics[width=\textwidth]{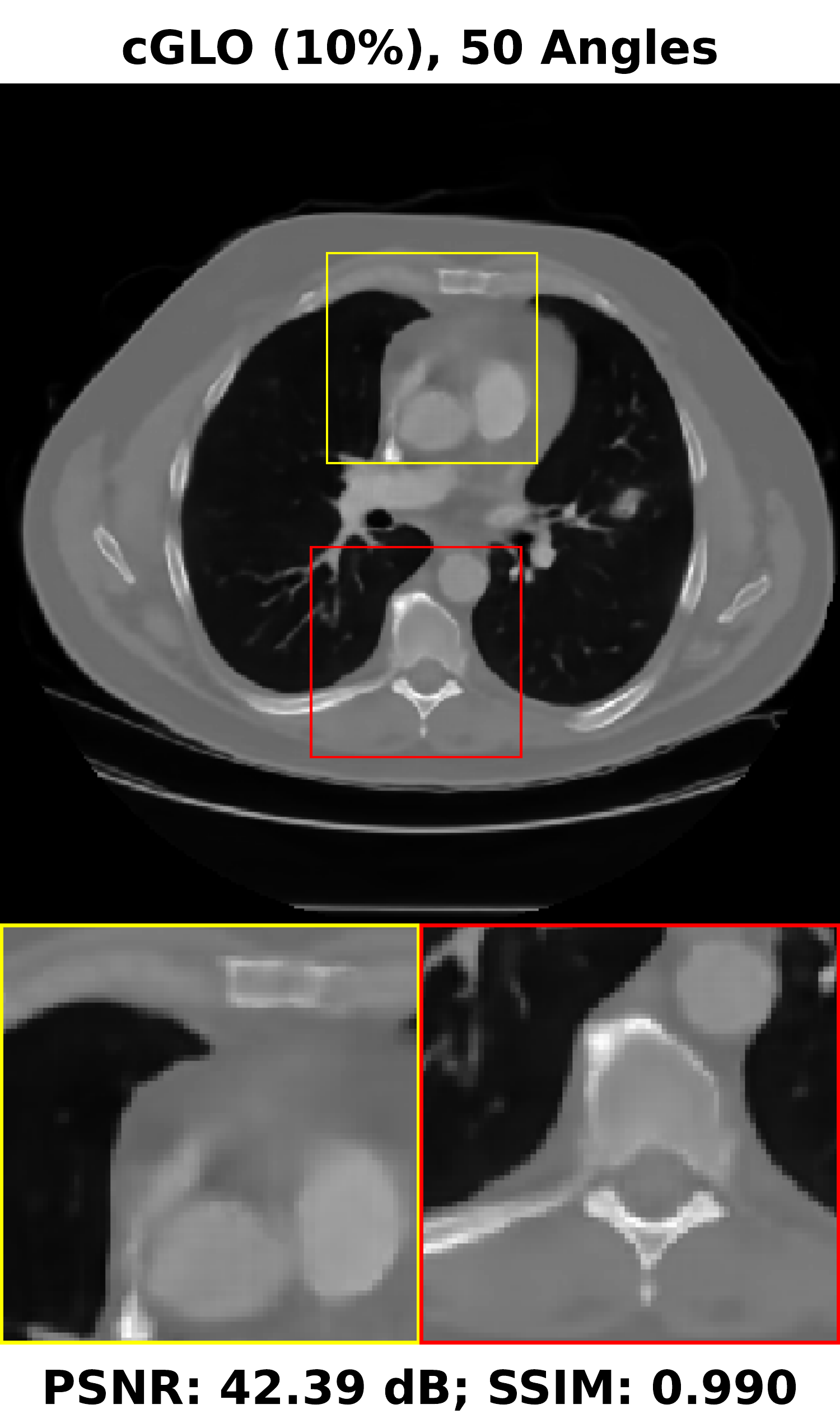}
        \end{subfigure}

        \begin{subfigure}[b]{0.245\textwidth}
            \includegraphics[width=\textwidth]{images/GTX.pdf}
        \end{subfigure}
        \hfill
        \begin{subfigure}[b]{0.245\textwidth}
            \includegraphics[width=\textwidth]{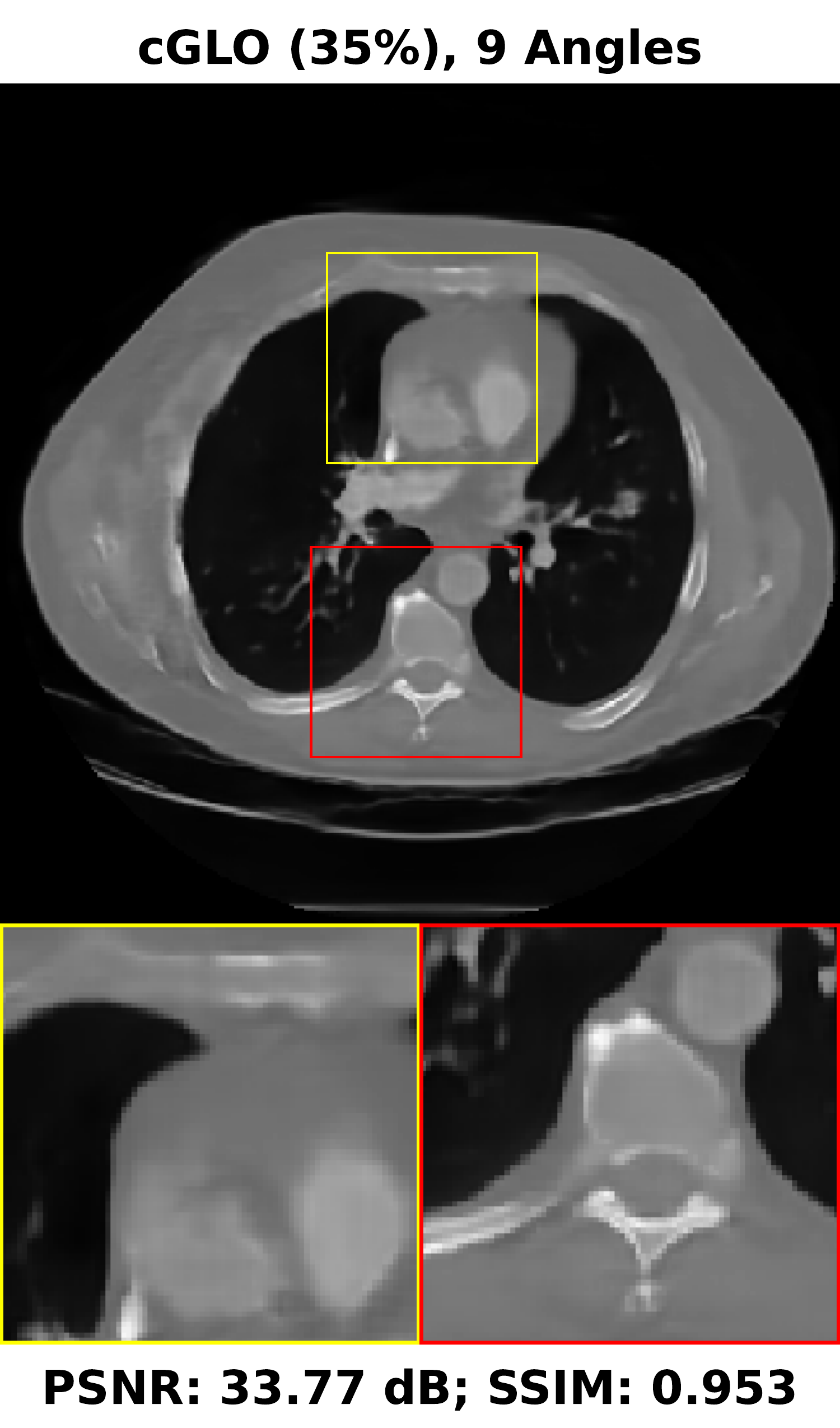}
        \end{subfigure}
        \hfill
        \begin{subfigure}[b]{0.245\textwidth}
            \includegraphics[width=\textwidth]{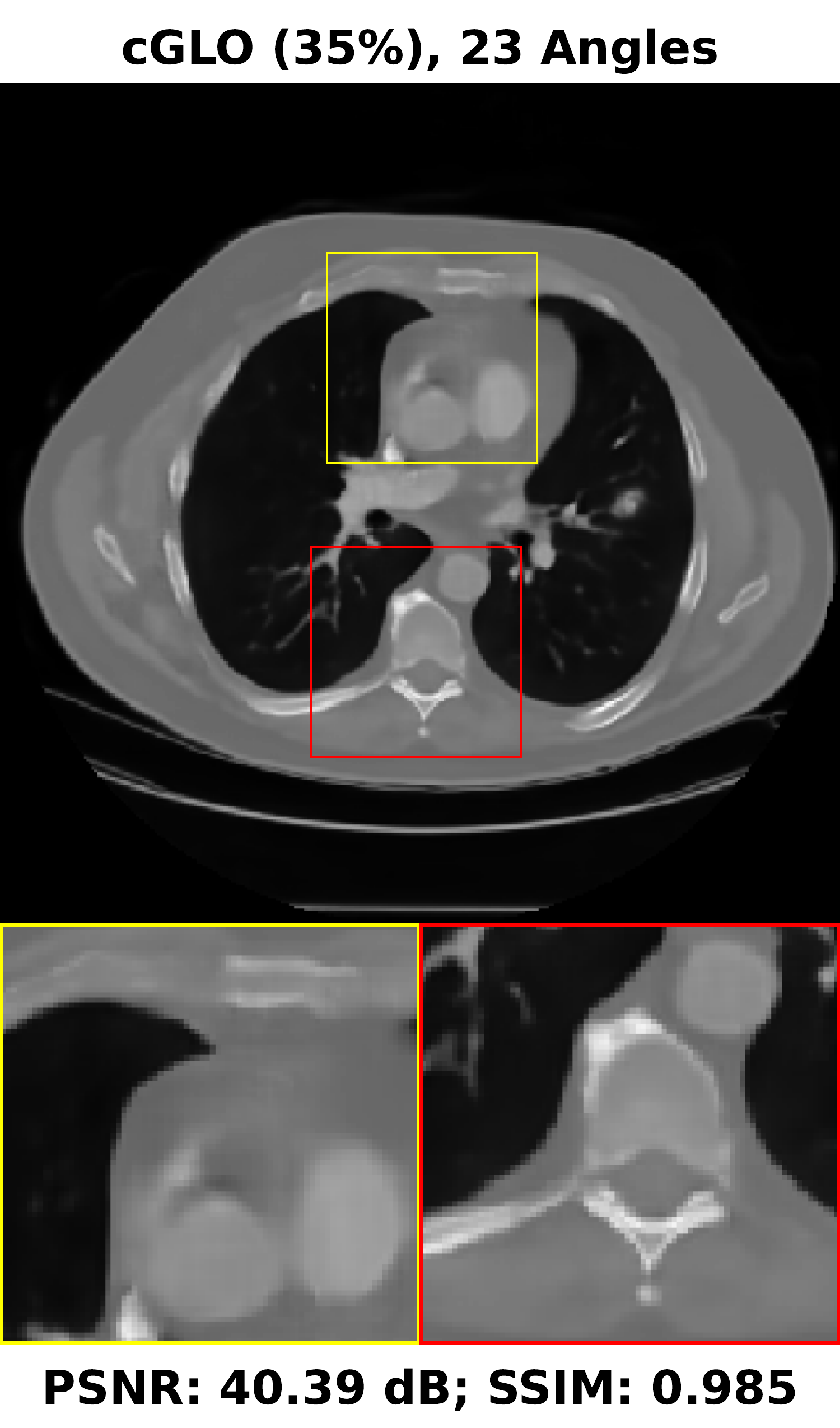}
        \end{subfigure}
        \hfill
        \begin{subfigure}[b]{0.245\textwidth}
            \includegraphics[width=\textwidth]{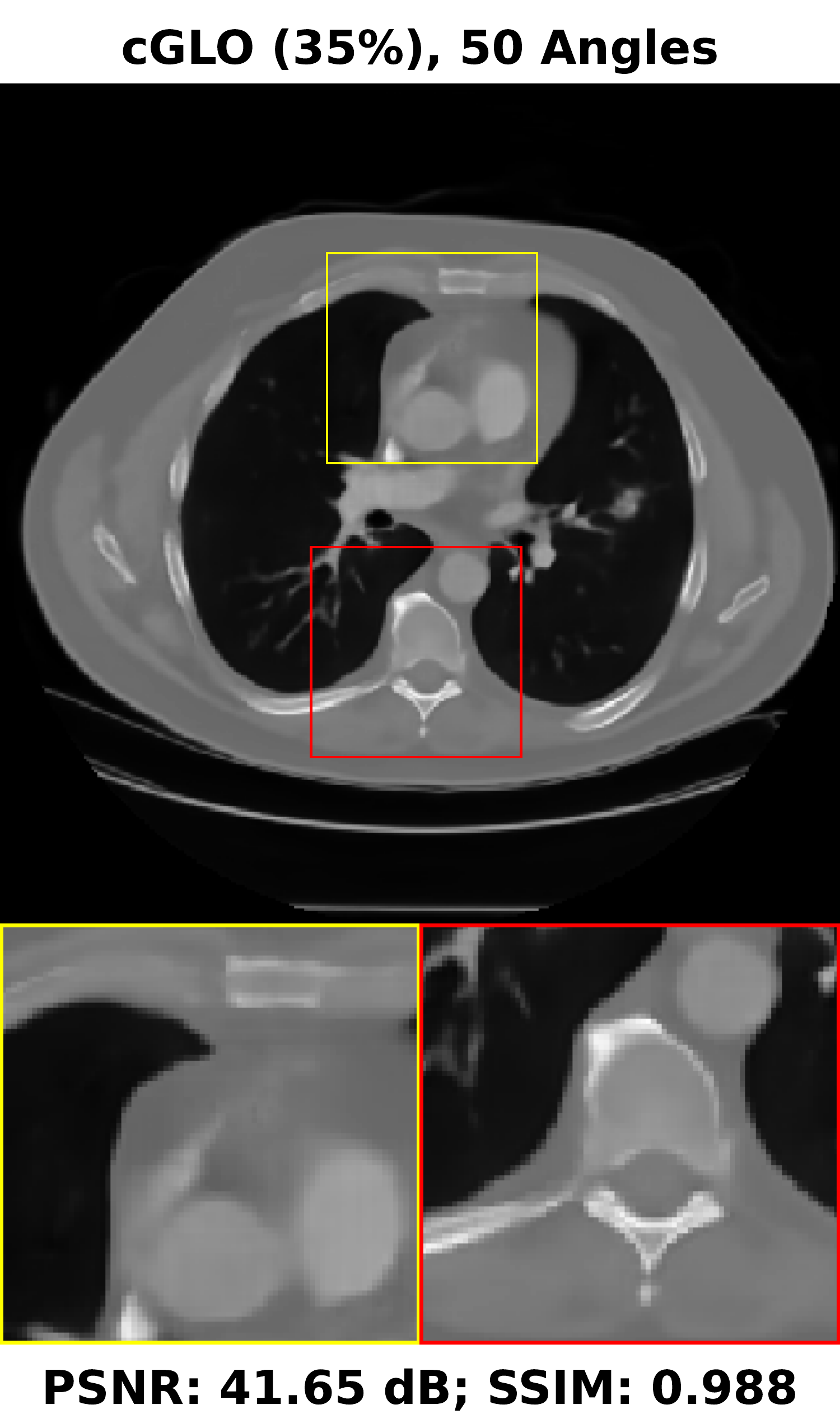}
        \end{subfigure}
    \caption{
        Examples of reconstructions of slices from the LIDC test set given : 9,
        23 and 50 experimental viewing angles. They are obtained with cGLO
        trained on the 2\%, 10\% and 35\% LIDC sub-datasets.
    }
    \label{fig:cglo_lidc_ex}
    \end{figure}

\end{document}